\documentclass[10pt,twocolumn,letterpaper]{article}

\usepackage{cvpr}              %

\usepackage[dvipsnames]{xcolor}

\usepackage{tikz}
\usepackage{graphicx}
\setkeys{Gin}{keepaspectratio}
\usepackage{amsmath}
\usepackage{amssymb}
\usepackage{booktabs}
\usepackage{float}
\usepackage{pgfplots}
\usepackage{tikz}
\usetikzlibrary{patterns}
\usepackage[precision=2, unit=mm]{lengthconvert}
\usepackage{diagbox}
\usepackage{enumitem}
\usepackage{tabularx}
\usepackage{multirow}
\usepackage{wrapfig}
\usepackage{cuted}
\usepackage[rightcaption]{sidecap}
\usepackage{calc}
\usepackage{bbm}
\usepackage{adjustbox}
\usepackage{mathtools}
\pgfplotsset{compat=1.18}

\newcolumntype{?}{!{\vrule width 1pt}}
\definecolor{cvprblue}{rgb}{0.21,0.49,0.74}
\usepackage[pagebackref,breaklinks,colorlinks,citecolor=cvprblue]{hyperref}

\def\paperID{203} %
\def\confName{3DV\xspace}
\def\confYear{2026\xspace}

\title{Symmetry Informative and Agnostic Feature Disentanglement for 3D Shapes}

\newcommand{\authorspace}{\hspace{0.5cm}}
\newcommand{\affiliationspace}{\hspace{0.12cm}}
\author{
Tobias Weißberg
\authorspace Weikang Wang
\authorspace Paul Roetzer
\authorspace Nafie El Amrani
\authorspace Florian Bernard \\
University of Bonn
\affiliationspace \& \affiliationspace
Lamarr Institute
}

\usepackage[utf8]{inputenc} %
\usepackage[T1]{fontenc}    %
\usepackage{hyperref}       %
\usepackage{url}            %
\usepackage{booktabs}       %
\usepackage{tabularx}
\usepackage{amsfonts}       %
\usepackage{nicefrac}       %
\usepackage{microtype}      %
\usepackage{xcolor}         %
\usepackage{graphicx}
\usepackage{adjustbox}
\usepackage{comment}
\usepackage{amsmath}
\usepackage{rotating}
\usepackage{multirow}
\usepackage{array}
\usepackage{caption}
\usepackage{subcaption}
\usepackage{pgfplots}
\pgfplotsset{compat=1.18} 
\usepackage[11pt]{moresize}
\usepackage[percentage]{overpic}
\usepackage{bbm}
\usepackage{enumitem}
\setitemize{noitemsep,topsep=0pt,parsep=0pt,partopsep=0pt,leftmargin=*}

\definecolor{ForestGreen}{rgb}{0.13, 0.55, 0.13}
\usepackage{pifont}
\usepackage{optidef}

\definecolor{mycolor1}{RGB}{90,130,213}%
\definecolor{mycolor3}{RGB}{231,92,46}%
\definecolor{mycolor11}{RGB}{134,168,235}%
\definecolor{mycolor33}{RGB}{231,156,130}%
\definecolor{mycolor2}{RGB}{230,130,46}%

\begin{document}
\twocolumn[{%
\maketitle
\begin{center}
    \captionsetup{type=figure}
    \newcommand{\imageheightat}{0.10\textheight}
\newcommand{\imageheightbt}{0.10\textheight}
\newcommand{\imagespacingt}{\hspace{0.1cm}}

\begin{tabular}{ccc?ccccc} 
\multicolumn{3}{c?}{\textbf{Feature Disentanglement}}
& \textbf{Intrinsic Symmetry}
& \multicolumn{2}{c}{\textbf{Left/Right Classification}} 
& \multicolumn{2}{c}{\textbf{Shape Matching}} \\

Input & Sym-info & Sym-agno & & & &  &  \\

\adjustbox{valign=m}{\includegraphics[height=\imageheightat]{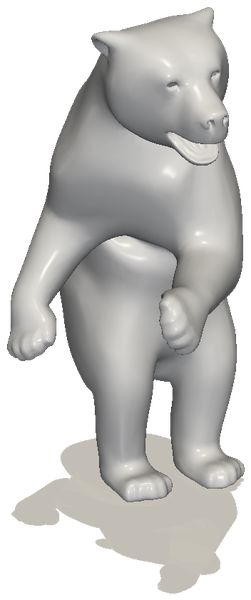}} &
\imagespacingt
\adjustbox{valign=m}{\includegraphics[height=\imageheightat]{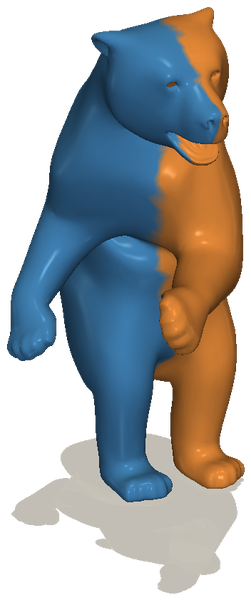}} &
\imagespacingt
\adjustbox{valign=m}{\includegraphics[height=\imageheightat]{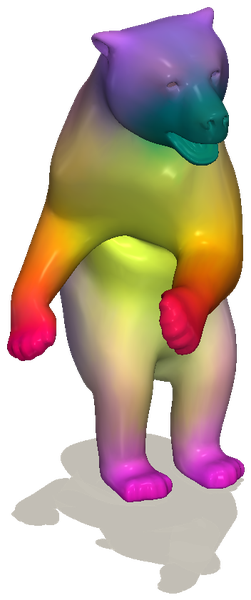}} & 
\imagespacingt
\adjustbox{valign=m}{\includegraphics[height=\imageheightat]{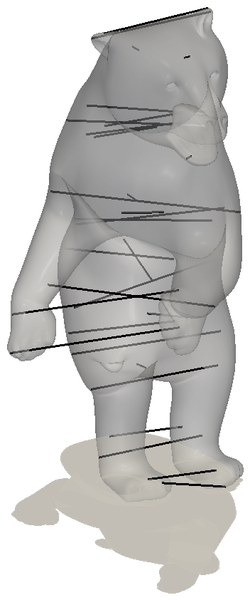}} & 
\imagespacingt
\adjustbox{valign=m}{\includegraphics[height=\imageheightat]{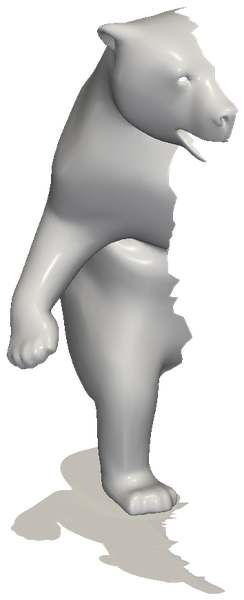}} &
\hspace{-1.5cm}
\adjustbox{valign=m}{\includegraphics[height=\imageheightbt]{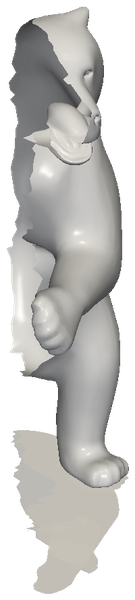}} &
\imagespacingt
\adjustbox{valign=m}{\includegraphics[height=\imageheightbt]{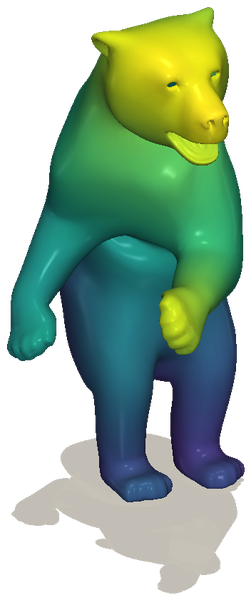}} &
\imagespacingt
\adjustbox{valign=m}{\includegraphics[height=\imageheightbt]{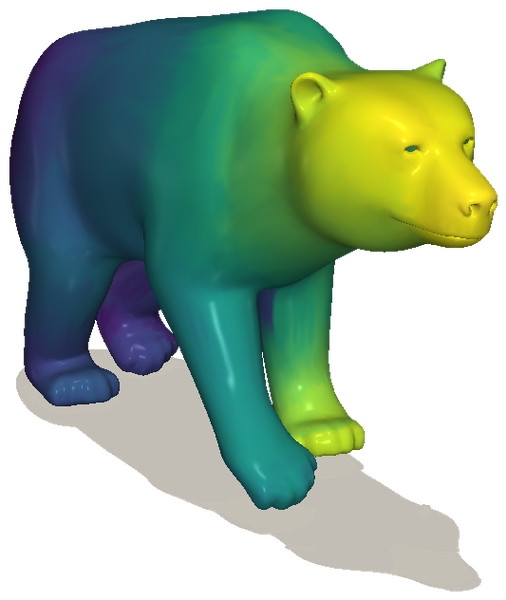}} \\

\adjustbox{valign=m}{\includegraphics[height=\imageheightat]{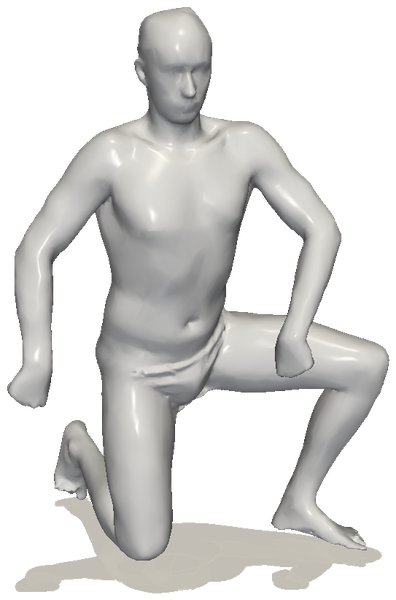}} &
\imagespacingt
\adjustbox{valign=m}{\includegraphics[height=\imageheightat]{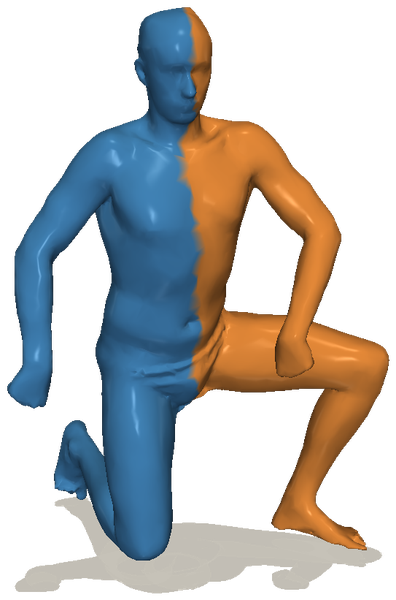}} &
\imagespacingt
\adjustbox{valign=m}{\includegraphics[height=\imageheightat]{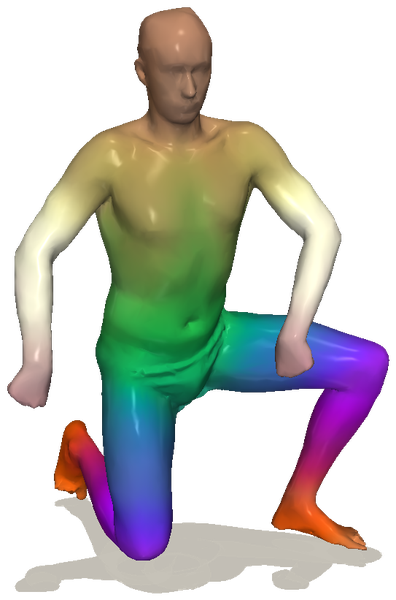}} & 
\imagespacingt
\adjustbox{valign=m}{\includegraphics[height=\imageheightat]{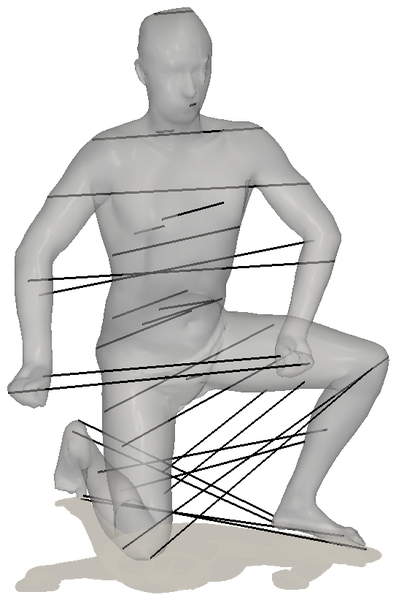}} & 
\imagespacingt
\adjustbox{valign=m}{\includegraphics[height=\imageheightat]{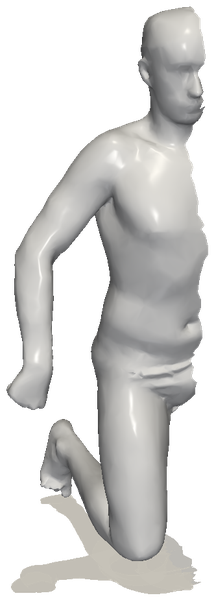}} &
\hspace{-1.5cm}
\adjustbox{valign=m}{\includegraphics[height=\imageheightbt]{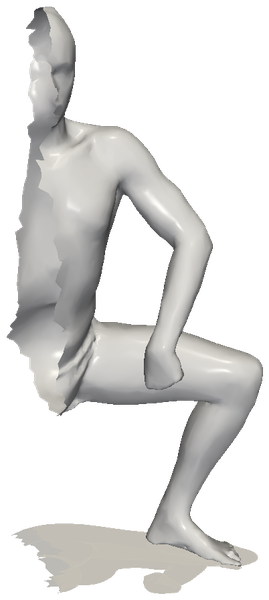}} &
\imagespacingt
\adjustbox{valign=m}{\includegraphics[height=\imageheightbt]{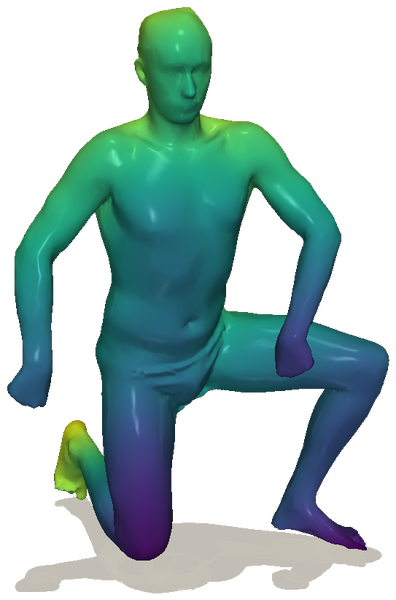}} &
\imagespacingt
\adjustbox{valign=m}{\includegraphics[height=\imageheightbt]{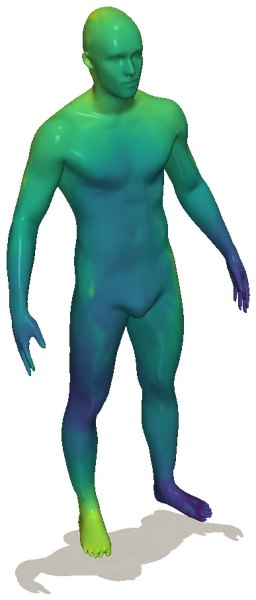}} \\

\end{tabular}

    \caption{Left: Visualization of disentangled intrinsic symmetry-informative (abbreviated as sym-info) and intrinsic symmetry-agnostic (abbreviated as sym-agno) descriptors. Right: Results of various shape analysis tasks (intrinsic symmetry detection, left/right classification and shape matching) using our disentangled intrinsic symmetry-aware descriptor pairs. %
    }
    \label{fig:teaser}
\end{center}
}]

\newcommand{\imageheighta}{0.1\textheight}
\newcommand{\imageheightb}{0.1\textheight}
\newcommand{\imagespacing}{\hspace{1cm}}

\begin{abstract}
\label{sec:abs}

Shape descriptors, i.e.,~per-vertex features of 3D meshes or point clouds, are fundamental to shape analysis. Historically, various handcrafted geometry-aware descriptors and feature refinement techniques have been proposed. Recently, several studies have initiated a new research direction by leveraging features from image foundation models to create semantics-aware descriptors, demonstrating advantages across tasks like shape matching, editing, and segmentation. Symmetry, another key concept in shape analysis, has also attracted increasing attention. Consequently, constructing symmetry-aware shape descriptors is a natural progression. Although the recent method $\chi$~\cite{wang2025symmetry} successfully extracted symmetry-informative features from semantic-aware descriptors, its features are only one-dimensional, neglecting other valuable semantic information. Furthermore, the extracted symmetry-informative feature is usually noisy and yields small misclassified patches. To address these gaps, we propose a feature disentanglement approach which is simultaneously symmetry informative and symmetry agnostic. Further, we propose a feature refinement technique to improve the robustness of predicted symmetry informative features. Extensive experiments, including intrinsic symmetry detection, left/right classification, and shape matching, demonstrate the effectiveness of our proposed framework compared to various state-of-the-art methods, both qualitatively and quantitatively. Project page: \href{https://tweissberg.github.io/chirality/}{https://tweissberg.github.io/chirality/} %
\end{abstract}
    
\section{Introduction}
\label{sec:intro}

Unlike image analysis, where features from foundational models (e.g.~DINO-V2~\cite{oquab2023dinov2}, CLIP~\cite{radford2021learning}, StableDiffusion~\cite{rombach2022high}) have already surpassed handcrafted features across a broad range of tasks~\cite{hedlin2024unsupervised, zhang2024telling, zhang2023tale, chen2023diffusiondet, tosi2024diffusion}, per-vertex features in shape analysis have still mostly been dominated by handcrafted descriptors 
such as WKS \cite{aubry2011wave}, HKS \cite{sun2009concise}, SHOT \cite{salti2014shot}, with different feature refinement methods \cite{sharp2022diffusionnet, cao2023unsupervised}. %

To address this gap, a recent work, Diff3F~\cite{dutt2024diffusion}, has introduced a new semantic-aware shape descriptor construction pipeline, which aggregated pixel-level 2D foundation model features from surrounding multi-view images rendered from a shape. Works that directly use Diff3F features~\cite{yuan2024cadtalk, fu2024sync4d, xie2025echomatch, roetzer2025fast, antic2024sdfit} or implement similar ideas together with other modules \cite{zhu2025densematcher, uzolas2025surface, chen2024unsupervised}, have demonstrated advantages over handcrafted features across various shape analysis tasks, including matching~\cite{shtedritski2024shic, xie2025echomatch, roetzer2025fast, zhu2025densematcher, uzolas2025surface, chen2024unsupervised}, editing~\cite{yuan2024cadtalk, fu2024sync4d} and shape-image correspondence~\cite{antic2024sdfit, shtedritski2024shic}.

However, as demonstrated in~\cite{zhang2023tale, zhang2024telling}, features from 2D foundation models can be ambiguous between intrinsic symmetry structures, such as left and right eyes of a cat. Since Diff3F~\cite{dutt2024diffusion} aggregates features from these 2D foundation models, it inherits left/right ambiguity which can lead to incorrect predictions in downstream tasks as shown in~\cite{wang2025symmetry}.

In order to address this intrinsic symmetry ambiguity in Diff3F~\cite{dutt2024diffusion} features, the recent work~\cite{wang2025symmetry} has proposed an intrinsic symmetry feature extraction pipeline to distill a one-dimensional symmetry-informative feature out of input semantic-aware shape descriptors. 
However, despite its impressive accuracy on left/right classification and the ability to boost matching performance, certain drawbacks exist. First, \cite{wang2025symmetry} only extracts a single one-dimensional symmetry-informative feature, rather than disentangling input feature into symmetry informative and agnostic parts. Secondly, it has limited ability to perform tasks such as intrinsic symmetry detection due to lack of sophisticated losses guiding the training process.
In addition, as noted in~\cite{wang2025symmetry}, resulting symmetry-informative features are noisy, i.e.~contain small mispredicted patches, see Fig.~\ref{fig:motivation} for an illustration.

In order to improve upon these drawbacks, we propose
a feature disentanglement approach which simultaneously extracts symmetry informative and symmetry agnostic features.
The former encodes only the left/right information, while the latter contains the remaining information of the input features. Ideally, the symmetry-informative descriptor tells whether a vertex belongs to the left or right part of the shape, 
and the symmetry-agnostic descriptor should be similar for intrinsically symmetric pairs of points, e.g. points on the left/right thumb of a human shape.
We adopt a combination of various unsupervised losses to guide the training of our intrinsic symmetry disentanglement network. In addition, to obtain a robust symmetry-informative descriptor, we also propose a symmetry-informative feature refinement technique based on a Markov Random Field (MRF) energy minimization formalism.
Various experiments including intrinsic symmetry detection, left/right classification and matching have been conducted to show the effectiveness of our disentanglement framework compared to other state-of-the-art methods. 
We summarize our main contributions as follows:
\begin{itemize}
    \item We propose an intrinsic symmetry-aware shape feature disentanglement framework, capable of decomposing per-vertex shape features into symmetry-informative and symmetry-agnostic pairs.
    \item We propose a generally applicable symmetry-aware feature refinement technique to obtain a robust symmetry-informative descriptor. 
    \item We show the broad applicability of our method by applying it to various shape analysis tasks including symmetry detection, left/right classification, and shape matching.
    \item Through extensive empiric evaluation we demonstrate the superiority of our disentangled symmetry-aware descriptors over state-of-the-art methods.
\end{itemize}
\input{./figures/motivation}

\section{Related work}
\label{sec:related}

In the following we discuss works that are most relevant to our framework. We discuss symmetry in shape analysis and 2D foundation models used in 3D shape descriptors.

\subsection{Symmetry in shape analysis}

As a fundamental pattern observable in our world, symmetry attracts significant interest in shape analysis, because of its potential to enhance a wide range of tasks including matching~\cite{yoshiyasu2014symmetry, cosmo2017consistent, zhang2013symmetry}, segmentation~\cite{riklin2009symmetry, riklin2006segmentation}, completion~\cite{sung2015data, ma2023symmetric, rumezhak2021towards} and editing~\cite{mitra2007symmetrization, zheng2015skeleton}. Given the vast literature on symmetry detection, we introduce only the most relevant works and refer readers to surveys~\cite{mitra2013symmetry, liu2010computational}.

Following a common convention~\cite{mitra2013symmetry, raviv2007symmetries, raviv2010full, kim2010mobius}, symmetry can be categorized into extrinsic and intrinsic ones. Extrinsic symmetry can be characterized via rigid transformations, while intrinsic symmetry is shape-inherent and is invariant to isometric transformations (e.g.,~symmetry independently of poses of a human body). We focus on intrinsic symmetry in this work, since it is more general in the sense that it contains extrinsic symmetry as a subset.

\citet{ovsjanikov2008global} show that the intrinsic symmetries of a shape could be transformed into the extrinsic symmetries in the signature space defined by the eigenfunctions of the Laplace-Beltrami operator, and devise an algorithm to detect and compute them.  Its follow-up work~\citet{nagar2018fast} improves efficiency based on the hypothesis that if a shape is intrinsically symmetric, then so is the geodesic distance between two intrinsic symmetric vertices. \citet{xu2009partial} use a generalized voting scheme to find the partial intrinsic symmetry curve without explicitly finding the intrinsic symmetric counterpart for each vertex. \citet{liu2012finding} propose a method for detecting intrinsic symmetry on genus-zero mesh surfaces by extracting closed curves through the construction of a weighted blend of conformal maps derived from triplets of extremal points identified via the average geodesic distance (AGD) function. Similarly, \citet{kim2010mobius} also detect symmetry invariant point sets using critical points of the AGD function, but get intrinsic symmetry from M\"{o}bius transformations computed by those selected points. 
\citet{xu2012multi} efficiently find pairs of intrinsically symmetric points using a voting based approach to detect intrinsic symmetries of shapes in different scales. 
\citet{wang2017group} describe symmetries as linear transformations of the eigenfunctions of the Laplace-Beltrami operator on shapes, and they propose an efficient global intrinsic symmetry detection method based on this new representation. \citet{qiao2022learning} parametrize intrinsic symmetry using a functional map matrix, which could be easily computed given the signs of Laplacian eigenfunctions under the symmetric mapping.

\begin{figure*}[ht!]
    \centering
    \includegraphics[width=\textwidth]{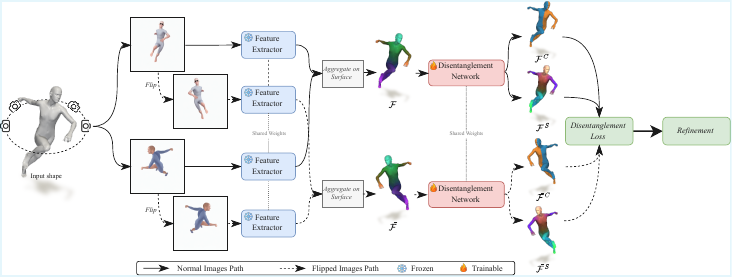}
    \caption{Overview of our method. As in Diff3F~\cite{dutt2024diffusion} and $\chi$~\cite{wang2025symmetry}, we render different views from a mesh. For each view, 2D foundation models are used to extract features from it and its horizontally flipped counterpart. Shape descriptors are obtained by back-projecting those image features. The disentanglement network then learns to disentangle symmetry from symmetry-agnostic information using a combination of disentanglement losses. A refinement technique is applied on $\mathcal{F}^{\text{C}}$ to increase robustness.
    }
    \label{fig:method_overview}
\end{figure*}

\subsection{2D foundation models assisted 3D shape descriptors}
Recently, with the emergence of various 2D foundation models \cite{oquab2023dinov2, caron2021emerging, radford2021learning, rombach2022high} and their demonstrated superiority over traditional features in various image-related tasks~\cite{hedlin2024unsupervised, zhang2024telling, zhang2023tale, chen2023diffusiondet, tosi2024diffusion}, an increasing number of works in 3D areas are considering using 2D foundation models to obtain 3D shape descriptors.
3D Highlighter~\cite{decatur20233d} uses a pre-trained CLIP~\cite{radford2021learning} encoder to localize semantic regions on a mesh using text as input. As a popular 3D scene representation, NeRF~\cite{mildenhall2021nerf} has various following works that leverage 2D foundation models for 3D representations. \citet{kobayashi2022decomposing} distill the knowledge of 2D foundation models~\cite{radford2021learning, caron2021emerging} to a 3D feature field optimized in parallel to the radiance field for semantic scene decomposition task. FeatureNeRF~\cite{ye2023featurenerf} leverages 2D foundation models~\cite{caron2021emerging, rombach2022high} to 3D space via neural rendering, and then extract deep features for querying 3D points from NeRF layers.
NeRF Analogies~\cite{fischer2024nerf} performs transfer along semantic affinity driven by semantic features from some 2D foundation model~\cite{caron2021emerging} to achieve multi-view appearance consistency. Different from implicit NeRF representations, there is also a line of research aiming at building explicit per-vertex 3D shape descriptors from 2D foundation models. With a shape as input, Diff3F~\cite{dutt2024diffusion} constructs per-vertex shape descriptor by averaging correspondent pixel features of 2D foundation models~\cite{oquab2023dinov2, rombach2022high} from surrounding rendered multi-view images. \citet{uzolas2025surface} refines Diff3F~\cite{dutt2024diffusion} features on the shape matching task by introducing a constraint that the geodesic distance between refined features of each vertex pair should be consistent with geodesic distance of this vertex pair. DenseMatcher~\cite{zhu2025densematcher} uses a similar pipeline as Diff3F~\cite{dutt2024diffusion} to first extract per-vertex features, and further refine shape descriptors using a trainable DiffusionNet~\cite{sharp2022diffusionnet}.

A recent work, $\chi$~\cite{wang2025symmetry}, is most relevant to this work. It extracts intrinsic symmetry-informative features from input semantic-aware shape descriptors aggregated from 2D foundation models similarly to Diff3F~\cite{dutt2024diffusion}. In this work, our proposed framework disentangles input semantic-aware shape descriptors into symmetry-informative and symmetry-agnostic pairs, rather than only extracting one-dimensional feature as $\chi$~\cite{wang2025symmetry} does. Better performances on intrinsic symmetry detection and shape matching show the effectiveness and validate the necessity of this newly introduced symmetry-agnostic descriptor, thus our proposed framework.

\section{Symmetry informative and agnostic feature disentanglement}
\label{sec: method}

In this section, we introduce our intrinsic symmetry-aware feature disentanglement framework (Fig.~\ref{fig:method_overview}) in detail. We disentangle input shape descriptors into symmetry-informative and symmetry-agnostic pairs. We first explain the process of obtaining per-vertex semantic-aware feature (Sec.~\ref{sec:feature}), which our lightweight disentanglement network (Sec.~\ref{sec:disentanglement}) uses as input. Next, the losses guiding the training of our network are elaborated in detail (Sec.~\ref{sec:losses}). And finally, we propose a general symmetry-informative descriptor refinement technique to improve the robustness of descriptors (Sec.~\ref{sec:refinement}). %

\subsection{Shape descriptors extraction}
\label{sec:feature}

We follow Diff3F~\cite{dutt2024diffusion} and $\chi$~\cite{wang2025symmetry} to extract per-vertex semantic-aware shape descriptors. 
We consider an untextured 3D surface shape in form of a mesh $\mathcal{X} = (\mathcal{V}, \mathcal{E})$ as input.
Here, $\mathcal{V}$ are the vertices of the shape and $\mathcal{E}$ the edges.
For that shape $\mathcal{X}$, we first generate $N$ surrounding textured views $\lbrace I_n \in \mathbb{R}^{H \times H \times 3} \rbrace_{n=1}^N$ from $N$ fixed camera poses using ControlNet~\cite{zhang2023adding}.
Then features from 2D foundation models are extracted, concatenated and upsampled to form the feature map $ F_n \in \mathbb{R}^{H \times W \times d}$ of each view $I_n$ with feature dimension $d$.

Using the known camera poses, we can recover correspondences between each vertex $v$ and pixels in each view.
Then we can compute the per-vertex feature $\mathcal{F}_v \in \mathbb{R}^d$ as the average of all correspondent pixel features over all $N$ views.
We flip each view from $\lbrace I_n \rbrace_{n=1}^N$ horizontally (i.e.~along x-axis) to obtain $\lbrace \bar{I}_n \rbrace_{n=1}^N$. Then we compute the feature map $\hat{F}_n \in \mathbb{R}^{H \times W \times d}$ of $\bar{I}_n$ similarly to $F_n$. Finally we flip $\hat{F}_n$ horizontally to obtain $\bar{F}_n$. The per-vertex semantic flipped feature $\mathcal{\bar{F}}_v$ is computed from $\lbrace \bar{F}_n \rbrace_{n=1}^N$ similarly to $\mathcal{F}_v$.

Finally, we apply vertex-wise normalization to each feature in the feature set $\lbrace \mathcal{F}_v \rbrace_{v=1}^{\lvert V \rvert}$, i.e.~$\mathcal{F}_v = \mathcal{F}_v / \lVert \mathcal{F}_v \rVert_2$.
We denote the stack of all features along the vertex dimension as $\mathcal{F} \in \mathbb{R}^{\lvert V \rvert \times d}$. $\bar{\mathcal{F}}$ can be obtained analogously.

\subsection{Symmetry-aware feature disentanglement}
\label{sec:disentanglement}

We use a lightweight auto-encoder as our disentanglement network. 
Specifically, we use an encoder $\text{E}$ to obtain an intermediate feature $\mathcal{F}^{\text{mid}} \in \mathbb{R}^{\vert V \vert \times d}$ from the input $\mathcal{F}$. 
Further, we feed $\mathcal{F}^{\text{mid}}$ into a decoder $\text{D}$ to reconstruct the input.

With intermediate features $\mathcal{F}^{\text{mid}}$ as input, we use the disentanglement steps depicted below:
\begin{itemize}
    \item First, we perform per-vertex normalization on $\mathcal{F}^{\text{mid}}$, i.e.~$\mathcal{F}^{\text{mid}}_v = \mathcal{F}^{\text{mid}}_v / \lVert \mathcal{F}^{\text{mid}}_v \rVert_2$, where $v \in \mathcal{V}$.
    \item To facilitate feature disentanglement, we project each normalized vertex descriptor $\mathcal{F}^{\text{mid}}_v \in \mathbb{R}^d$ using a global trainable orthonormal square matrix $A$ as: $\mathcal{F}^{\text{proj}}_v = \mathcal{F}^{\text{mid}}_v  A$.
    \item After that, we separate the projected descriptor $\mathcal{F}^{\text{proj}}_v$ into two: $\mathcal{F}^{\text{proj}}_v :=[\mathcal{F}^{\text{C}}_v, \mathcal{F}_v^{\text{S}}]$, where $\mathcal{F}_v^{\text{C}} \in \mathbb{R}$ represents the symmetry-informative (chirality \cite{wang2025symmetry}) descriptor, and $\mathcal{F}_v^{\text{S}} \in \mathbb{R}^{d-1}$ represents the symmetry-agnostic descriptor.
    \item Additionally, we apply normalization on the symmetry-agnostic descriptor $\mathcal{F}_v^{S}$ as: $\mathcal{F}_v^{\text{S}} = \mathcal{F}_v^{\text{S}} / \lVert \mathcal{F}_v^{\text{S}} \rVert_2$.
\end{itemize}
Similarly, we follow the same steps to arrive at respective flipped symmetry informative and agnostic descriptors $\bar{\mathcal{F}}_v^{\text{C}}, \bar{\mathcal{F}}_v^{\text{S}}$ using flipped $\mathcal{\bar{F}}$ as input. 

In order to make the disentanglement framework intrinsic symmetry aware, we use a combination of semantic and geometric losses to guide the training of the network (the trainable components are encoder $\text{E}$, decoder $\text{D}$, and the orthonormal matrix $\text{A}$), which we discuss next.

\subsection{Unsupervised losses}
\label{sec:losses}

With stacked per-vertex feature $\mathcal{F}$, we collect the disentangled symmetry-informative and symmetry-agnostic features from the network, and stack them along the vertex dimension to obtain $\mathcal{F}^{\text{C}} \in [-1, 1]^{\lvert V \rvert}$ and $\mathcal{F}^{\text{S}} \in [-1, 1]^{\lvert V \rvert \times (d-1)}$, respectively. Similarly, we compute $\mathcal{\bar{F}}^{\text{C}}$ and $\mathcal{\bar{F}}^{\text{S}}$ using $\mathcal{\bar{F}} $ as input.

\paragraph{Dissimilarity loss $\mathcal{L}_{\text{dis}}$~\cite{wang2025symmetry}.} We apply a dissimilarity loss $\mathcal{L}_{\text{dis}}$ to push symmetry-informative descriptors of each vertex and its intrinsic symmetric counterpart away from each other: %
\begin{equation}
    \mathcal{L}_{\text{dis}} = -\frac{1}{\sqrt{\lvert V \rvert}}\lVert \mathcal{F}^{\text{C}} - \mathcal{\bar{F}}^{\text{C}} \rVert_2.
\end{equation}

\paragraph{Similarity loss $\mathcal{L}_{\text{sim}}$.} Further, we apply a similarity loss $\mathcal{L}_{\text{sim}}$ to make sure the symmetry-agnostic descriptors of each vertex and its intrinsic symmetric counterpart are close:
\begin{equation}
    \mathcal{L}_{\text{sim}} = \frac{1}{\sqrt{\lvert V \rvert}}\lVert \mathcal{F}^{\text{S}} - \mathcal{\bar{F}}^{\text{S}} \rVert_F.
\end{equation}

\paragraph{Reconstruction loss $\mathcal{L}_{\text{rec}}$~\cite{wang2025symmetry}.} We want to avoid information loss of the intermediate feature $\mathcal{F}^{\text{mid}}$ given by the encoder $\text{E}$.
Thus, we apply a reconstruction loss
\begin{equation}
    \mathcal{L}_{\text{rec}} = \frac{1}{\sqrt{|V|}}\lVert [\mathcal{F},  \bar{\mathcal{F}}] - \text{D}(\text{E}([\mathcal{F}, \bar{\mathcal{F}}])) \rVert_F.
\end{equation}
Here, $[\mathcal{F}, \bar{\mathcal{F}}]$ corresponds to the column stack of $\mathcal{F}$ and $\bar{\mathcal{F}}$.

\paragraph{Boundary loss $\mathcal{L}_{\text{bou}}$.} To regularize the boundary to follow an approximately straight path along the surface manifold, we introduce a boundary loss $\mathcal{L}_{\text{bou}}$. This loss is built upon two components. First, we define the tangential cosine similarity $C_v(u, w)$ between the incoming edge $(u, v)$ and outgoing edge $(v, w)$ for given vertices $u, v, w \in V$. %
Second, we define $L(u, v, w)$ as the sum of squared differences of symmetry-informative features across these edges:
\begin{equation}
    L(u,v,w) = \lVert \mathcal{F}^{\text{C}}_u - \mathcal{F}^{\text{C}}_v \rVert_2^2 + \lVert \mathcal{F}^{\text{C}}_v - \mathcal{F}^{\text{C}}_w \rVert_2^2,
    \label{eq:boundary_loss}
\end{equation}
where $\bar{L}(u,v,w)$ is defined analogously using $\bar{\mathcal{F}}$.
To encourage colinear neighboring pairs $u, w \in V$ around each vertex $v \in V$ with similar symmetry-informative features $\mathcal{F}^{\text{C}}_v$, we formulate the loss as
\begin{equation}
    \begin{aligned}
    \mathcal{L}_{\text{bou}} =
    \frac{1}{\vert V \vert} \sum_{v \in V} \bigl( & \min_{(u,v,w) \in \mathcal{S}_v} \lbrace L(u,v,w) - C_v(u, w) \rbrace \\
    + &\min_{(u,v,w) \in \mathcal{S}_v} \lbrace \bar{L}(u,v,w) - C_v(u, w) \rbrace \bigr).
    \end{aligned}
\end{equation}
Here, $\mathcal{S}_v$ denotes the set of vertex tuples $(u,v,w)$ such that $(u,v), (v,w) \in \bar{\mathcal{E}}$,
where $\bar{\mathcal{E}}$ is the set of directed edges which is obtained from $\mathcal{E}$ by interpreting every undirected edge as two opposing directed edges.
A more thorough explanation and visualization can be found in Sec.~\ref{sec: boundary_loss} in the supplementary material.

\paragraph{Consistency loss $\mathcal{L}_{\text{con}}$.} To regularize the features, we enforce self-consistency by matching a shape to itself through its mirrored version. %
Let $W \in (\mathbb{R}^+)^{\vert V \vert \times \vert V \vert}$ with
$
    W_{i, j} = \left( \mathcal{F}^{\text{C}}_{v_i} -\mathcal{F}^{\text{C}}_{v_j} \right)^2
$ be the squared difference %
between $\mathcal{F}^{\text{C}}$ for different vertices $v_i, v_j \in V$ %
, and let $
    C_c = \mathcal{F}^{\text{S}}  (\mathcal{F}^{\text{S}})^{\top} \in [-1, 1]^{\vert V \vert \times \vert V \vert} 
$ be the similarity matrix between the $\mathcal{F}^{\text{S}}$ features.  %
We then compute $\Pi \in [0, 1]^{\vert V \vert \times \vert V \vert}$ as the element-wise product after the min-max normalization to the $[0,1]$ interval %
, i.e.
\begin{equation}
    \Pi = \text{norm}(W) \odot \text{norm}(C_c).
\end{equation}
Intuitively, $\Pi$ represents a soft assignment of each vertex to its intrinsically symmetric counterpart. Consequently, $\Pi^2$ should ideally approximate the identity matrix, representing a mapping of the shape to itself via the symmetric reflection. $\bar{\Pi}$ is defined analogously with $\bar{\mathcal{F}}$.

This allows us to introduce the loss $\mathcal{L}_{\text{con}}$ as 
\begin{equation}
    \mathcal{L}_{\text{con}} = \frac{1}{\vert V \vert} \lVert [\mathbb{I}_{\vert V \vert},  \mathbb{I}_{\vert V \vert}] - %
    [\Pi^2, \bar{\Pi}^2] \rVert_F,
\end{equation}
where $\mathbb{I}_{\vert V \vert}$ is an identity matrix of size $\vert V \vert \times \lvert V \rvert$.

\paragraph{Overall loss. } The overall loss is a weighted combination of those unsupervised losses:
\begin{equation}
    \mathcal{L} = \mathcal{L}_{\text{dis}} + \lambda_1 \mathcal{L}_{\text{sim}} + \lambda_2 \mathcal{L}_{\text{rec}} + \lambda_3 \mathcal{L}_{\text{bou}} + \lambda_4 \mathcal{L}_{\text{con}}.
\end{equation}
For choices of $\lambda_1, \lambda_2, \lambda_3$ and $\lambda_4$, we refer to the implementation details in Sec.~\ref{sec:implementation} in the supplementary materials.

\def\lrfeature{\mathcal{F}^C}
\def\lrfeatureRefined{\Tilde{\mathcal{F}}^C}
\subsection{Symmetry aware feature refinement}
\label{sec:refinement}

As shown in Fig.~\ref{fig:motivation}, $\chi~\cite{wang2025symmetry}$ produces incorrect small patches in symmetry-informative descriptors.
To circumvent such undesirable solutions, we propose a simple yet effective refinement technique based upon a Markov Random Field (MRF) energy minimization formalism.
In particular, we compute refined symmetry-informative features $\lrfeatureRefined$ by solving the following MRF problem
\begin{equation}
\label{eq:map-mrf}
    \underset{\lrfeatureRefined\in\{0,1\}^{|\mathcal{V}|}}{\min}\sum_{v\in \mathcal{V}} \theta_v(\lrfeatureRefined_v) + \sum_{(v,\bar{v})\in \mathcal{E}}  \theta_{v\bar{v}}(\lrfeatureRefined_v, \lrfeatureRefined_{\bar{v}}).
\end{equation}
We first do min-max normalization of $\lrfeature$ to transform it into $\lrfeature \in [0,1]^{|\mathcal{V}|}$. Then we choose for the unary potentials $\theta_v(0) = \lrfeature_v$ and $\theta_v(1) = 1 - \lrfeature_v$ for all $v\in\mathcal{V}$.
Furthermore, for the pairwise potentials, we use $\theta_{v\bar{v}}(0, 0) = \theta_{v\bar{v}}(1, 1) = 0$ and $\theta_{v\bar{v}}(0, 1) = \theta_{v\bar{v}}(1, 0) = 1$ for all edges $(v,\bar{v})\in \mathcal{E}$.
As an intuition, using these potentials essentially favors a refined symmetry-informative feature $\lrfeatureRefined$ which is as close as possible to the predicted feature $\lrfeature$ while having as little boundary as possible (which effectively reduces the patches).
We efficiently solve \eqref{eq:map-mrf} using graph cuts~\cite{delong2012fast,boykov2004experimental,kolmogorov2004energy,boykov2002fast}.

Since this post processing technique does not rely on the design of our disentanglement framework, it could be applied on any coarse symmetry-informative features. We validate the general usage of this technique also on 
$\chi$~\cite{wang2025symmetry}.

\section{Experiments}
\label{sec:experiments}

We conduct different shape analysis experiments on our disentangled symmetry-informative and symmetry-agnostic descriptors pair to show the effectiveness of our proposed framework. In Sec.~\ref{sec:dataset}, we first introduce all datasets we use. In Sec.~\ref{sec:intrinsic_symmetry}, intrinsic symmetry detection shows the superiority of our disentangled descriptors pair compared to several other shape descriptors. Then, in Sec.~\ref{sec:left_right}, the left/right classification experiment validates our symmetry-informative descriptor alone. We also introduce a new metric to quantify the robustness of symmetry-informative descriptors and the results prove the effectiveness of our refinement technique. The application of our features in shape matching is explored in Sec.~\ref{sec:matching}.
Lastly, to verify the necessity of our losses, an ablation study is done in Sec.~\ref{sec:ablation}.

\begin{table*}[tbh]
\setlength{\tabcolsep}{3.5pt}
  \centering
  \begin{tabular}{@{}lccccccccc@{}}
    \toprule
    Train & \textbf{BeCoS} & \multicolumn{2}{c}{\textbf{BeCoS\textsubscript{-h}}} & \multicolumn{2}{c}{\textbf{BeCoS\textsubscript{-a}}} & \multicolumn{2}{c}{\textbf{FAUST}} & \multicolumn{2}{c}{\textbf{SMAL}} \\
    \cmidrule(r){2-2}
    \cmidrule(r){3-4}
    \cmidrule(r){5-6}
    \cmidrule(r){7-8}
    \cmidrule(r){9-10}
    Test & \textbf{BeCoS} & \textbf{BeCoS\textsubscript{-h}} & \textbf{BeCoS\textsubscript{-a}} & \textbf{BeCoS\textsubscript{-h}} & \textbf{BeCoS\textsubscript{-a}} & \textbf{FAUST} & \textbf{SCAPE} & \textbf{SMAL} & \textbf{TOSCA} \\
    \midrule
    DINO+SD + GT & 0.098 & 0.091 & 0.100 & 0.091 & 0.100 & 0.037 & 0.061 & 0.125 & 0.059 \\
    Ours + GT & \textbf{0.059} & \textbf{0.058} & \textbf{0.093} & \textbf{0.073} & \textbf{0.061} & \textbf{0.025} & \textbf{0.032} & \textbf{0.070} & \textbf{0.050} \\
    \midrule
    DINO+SD + $\chi$ & 0.145 & 0.164 & 0.182 & 0.257 & 0.132 & 0.054 & 0.128 & 0.129 & 0.154 \\
     Ours & \underline{0.101} & \underline{0.114} & 0.162 & 0.197 & \underline{0.087} & \textbf{0.042} & \underline{0.073} & \textbf{0.075} & \underline{0.115} \\
    DINO+SD + ($\chi$ + refine) & 0.120 & 0.122 & \underline{0.161} & \underline{0.144} & 0.120 & 0.049 & 0.103 & 0.135 & 0.124 \\
    Ours + refine & \textbf{0.078} & \textbf{0.077} & \textbf{0.146} & \textbf{0.108} & \textbf{0.078} & \textbf{0.042} & \textbf{0.049} & \underline{0.076} & \textbf{0.084} \\
    \bottomrule
  \end{tabular}
  \caption{Geodesic error of intrinsic symmetry detection using different features across datasets. The best and second-best results for each case are written in bold and underlined, respectively. Both our sym-agno feature and the refinement step lead to consistently lower errors.}
  \label{tab:intrinsic_symmetry}
\end{table*}

\subsection{Datasets}
\label{sec:dataset}

We use a recently proposed benchmark BeCoS~\cite{ehm2024beyond} to generate a large scale dataset with rich annotations including inter-shape correspondences, per-shape self-correspondences (i.e.~intrinsic symmetric pairs annotations), and left/right annotations. Additionally, we also experiment on several popular datasets: FAUST~\cite{bogo2014faust}, SCAPE~\cite{anguelov2005scape}, SMAL~\cite{Zuffi_CVPR_2017} and TOSCA~\cite{bronstein2008numerical}. More details can be found below.

\begin{itemize}
    \item The BeCoS~\cite{ehm2024beyond} dataset consists of humanoid and four-legged animals with 20370/284/274 train/test/validation split, generated from 7 remeshed shape datasets, namely TOSCA~\cite{bronstein2008numerical}, FAUST \cite{bogo2014faust}, SCAPE~\cite{anguelov2005scape}, KIDS~\cite{rodola2014dense}, DT4D~\cite{magnet2022smooth}, SMAL~\cite{Zuffi_CVPR_2017} and SHREC’20~\cite{dyke2020shrec}.
    \item BeCoS\textsubscript{-h}~\cite{ehm2024beyond} consists of only humanoid shapes of BeCoS~\cite{ehm2024beyond} with 9697/64/58 train/test/validation split. 
    \item BeCoS\textsubscript{-a}~\cite{ehm2024beyond} consists of only four-legged animals of BeCoS~\cite{ehm2024beyond} with 10263/220/216 train/test/validation split.
    \item FAUST~\cite{bogo2014faust}, SCAPE~\cite{anguelov2005scape}, SMAL~\cite{Zuffi_CVPR_2017} and TOSCA~\cite{bronstein2008numerical} datasets. We follow the original train/validation/test splits where available, and otherwise use the \textsc{BeCoS}~\cite{ehm2024beyond} splits. Since these datasets lack left/right and self-correspondence annotations, we obtain them via \textsc{BeCoS}.
\end{itemize}

\subsection{Intrinsic symmetry detection}
\label{sec:intrinsic_symmetry}

\paragraph{Metrics.}

For a given shape $\mathcal{X}$ represented with a vertex set $V \in \mathbb{R}^{|V| \times 3}$, intrinsic symmetry detection aims to match each vertex $v \in V$ to its intrinsic symmetric correspondent vertex $\bar{v} \in V$. Thus we use the average matching error as our evaluation metrics: 
\begin{equation}
    \text{err}_{\text{int}} = \frac{1}{\vert V \vert} \sum_{v \in V} \ \text{dist}(f(v),  v_{\text{gt}}^{\text{sym}}), 
    \label{eq:matching_error}
\end{equation}
where $f(v)$ is the predicted intrinsic symmetric vertex of $v$, $v_{\text{gt}}^{\text{sym}}$ is the ground truth intrinsic symmetry vertex of $v$, and $\text{dist}(\cdot, \cdot)$ is the geodesic distance normalized by the square root of the area of a shape following~\cite{kim2011blended}.

\newcommand{\imageheightaa}{0.05\textheight}
\newcommand{\imageheightab}{0.06\textheight}
\newcommand{\imagespacingm}{\hspace{-0.4cm}}
\newcommand{\imagespacingma}{\hspace{-0.3cm}}

\begin{figure}
    \centering
    \begin{tabular}{cccccc}
          \footnotesize{Source} & \imagespacingma & \imagespacingma \footnotesize{$\chi$} & \imagespacingm \footnotesize{$\chi$} + Refine & \imagespacingm \footnotesize{Ours}  & \imagespacingm \footnotesize{Ours + Refine} \\

          \multirow{2}{*}{\adjustbox{valign=m}{\includegraphics[height=0.05\textheight]{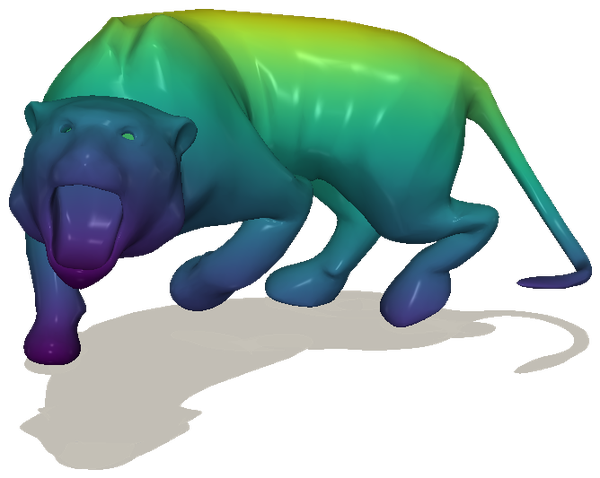}}} & \imagespacingma
          \rotatebox[]{90}{\footnotesize{Result}} & \imagespacingma
          
          \adjustbox{valign=m}{\includegraphics[height=\imageheightaa]{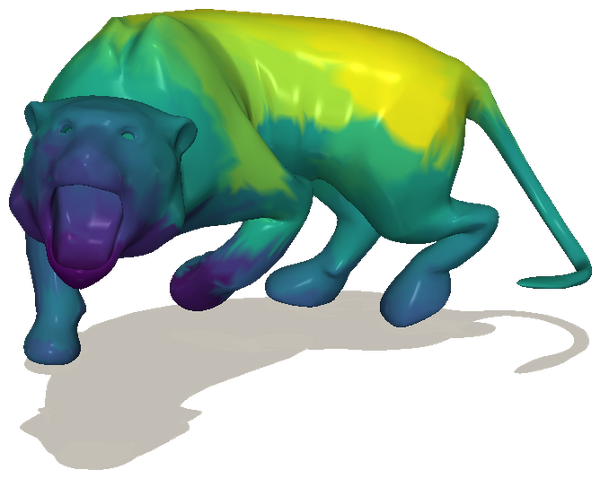}} & \imagespacingm
          \adjustbox{valign=m}{\includegraphics[height=\imageheightaa]{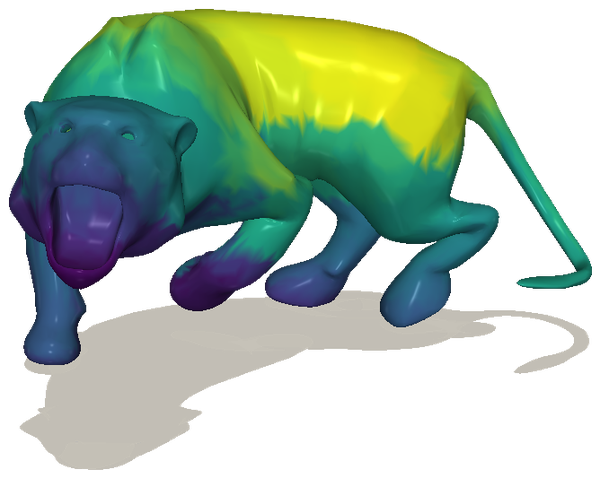}} & \imagespacingm
          \adjustbox{valign=m}{\includegraphics[height=\imageheightaa]{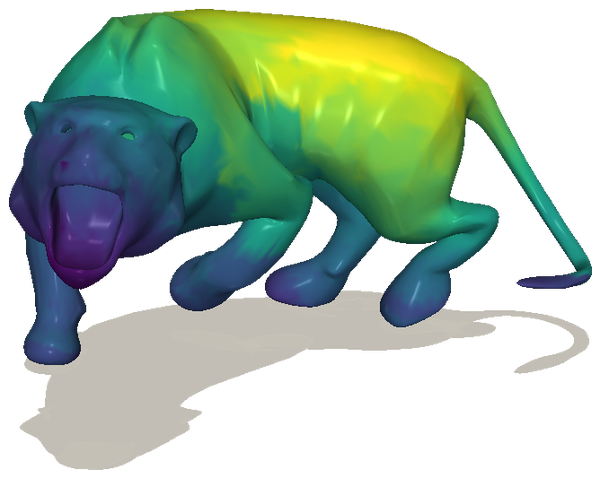}} & \imagespacingm
          \adjustbox{valign=m}{\includegraphics[height=\imageheightaa]{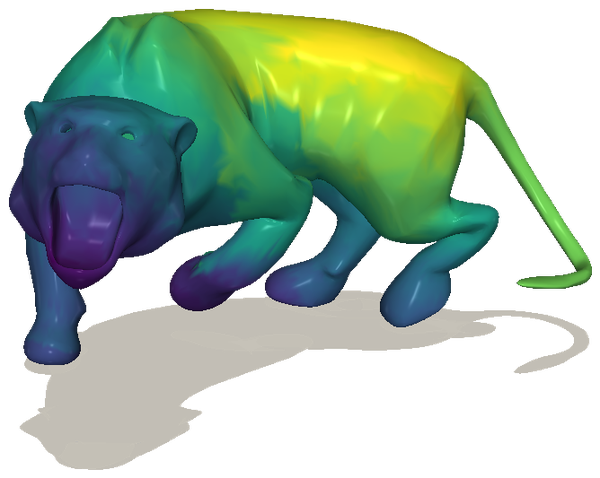}} \\

           & \imagespacingma \rotatebox[]{90}{\footnotesize{Error}} & \imagespacingma
          \adjustbox{valign=m}{\includegraphics[height=\imageheightaa]{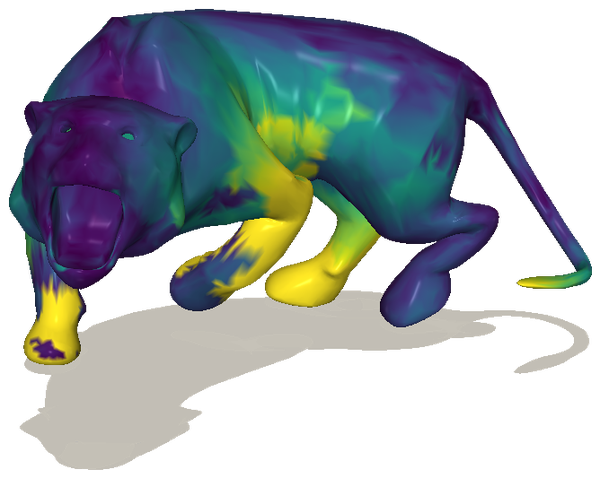}} & \imagespacingm
          \adjustbox{valign=m}{\includegraphics[height=\imageheightaa]{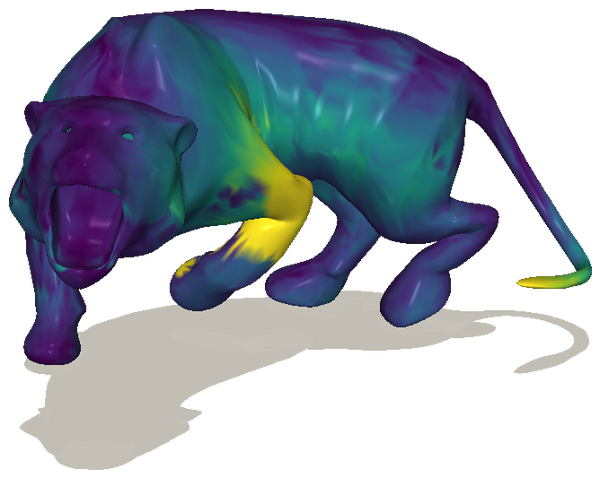}} & \imagespacingm
          \adjustbox{valign=m}{\includegraphics[height=\imageheightaa]{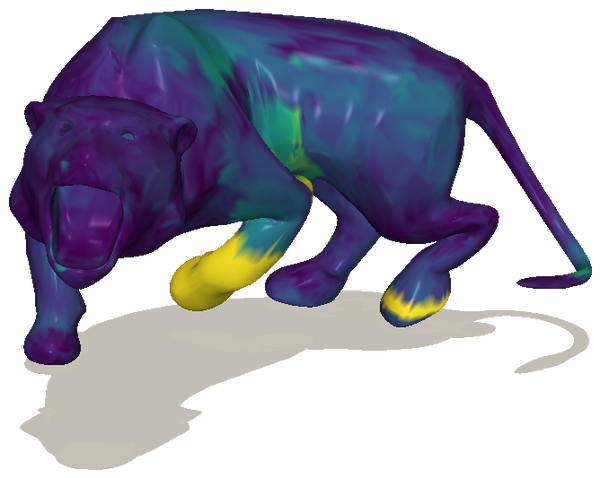}} & \imagespacingm
          \adjustbox{valign=m}{\includegraphics[height=\imageheightaa]{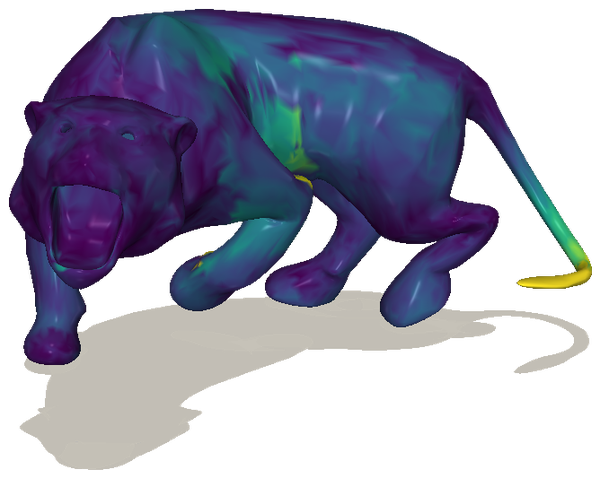}} \\

          \multirow{2}{*}{\adjustbox{valign=m}{\includegraphics[height=0.06\textheight]{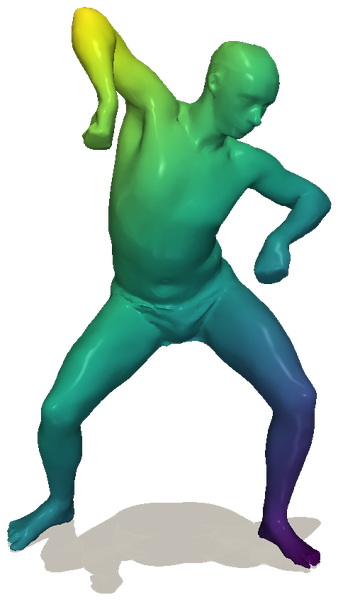}}} & \imagespacingma
          \rotatebox[]{90}{\footnotesize{Result}} & \imagespacingma
          \adjustbox{valign=m}{\includegraphics[height=\imageheightab]{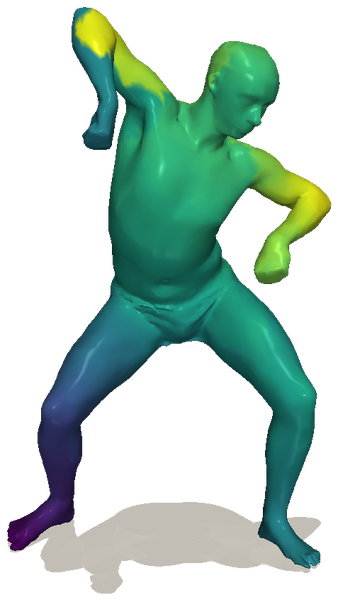}} & \imagespacingm
          \adjustbox{valign=m}{\includegraphics[height=\imageheightab]{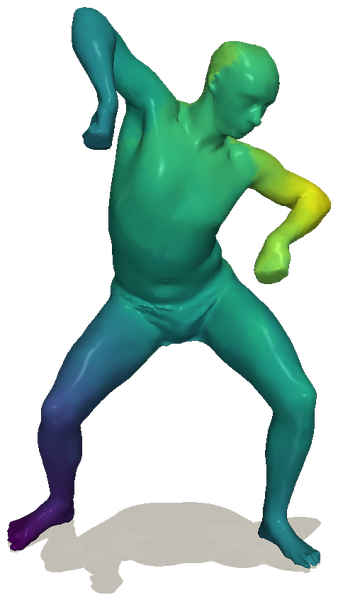}} & \imagespacingm
          \adjustbox{valign=m}{\includegraphics[height=\imageheightab]{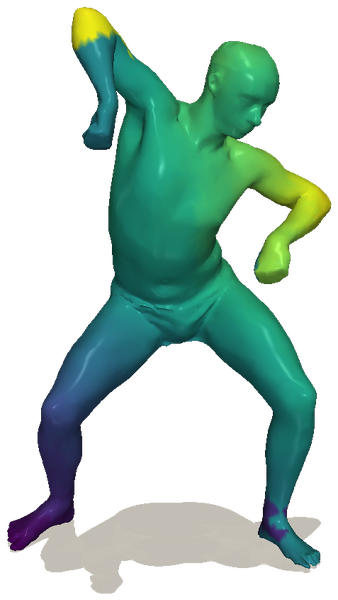}} & \imagespacingm
          \adjustbox{valign=m}{\includegraphics[height=\imageheightab]{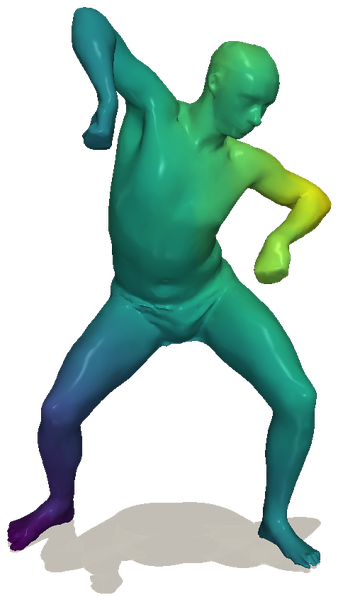}} \\

           & \imagespacingma \rotatebox[]{90}{\footnotesize{Error}} & \imagespacingma
          \adjustbox{valign=m}{\includegraphics[height=\imageheightab]{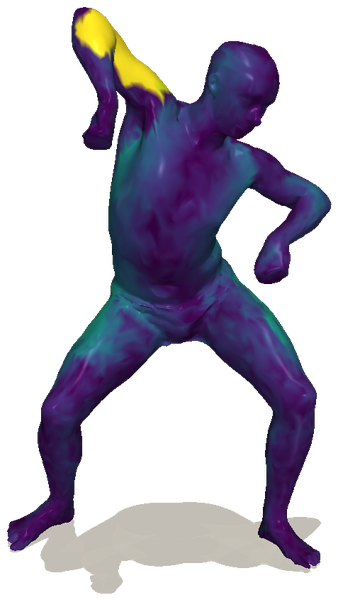}} & \imagespacingm
          \adjustbox{valign=m}{\includegraphics[height=\imageheightab]{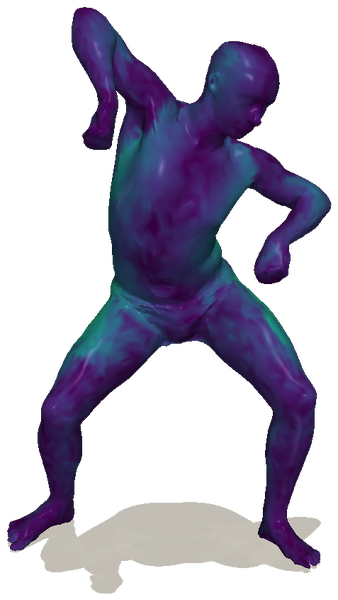}} & \imagespacingm
          \adjustbox{valign=m}{\includegraphics[height=\imageheightab]{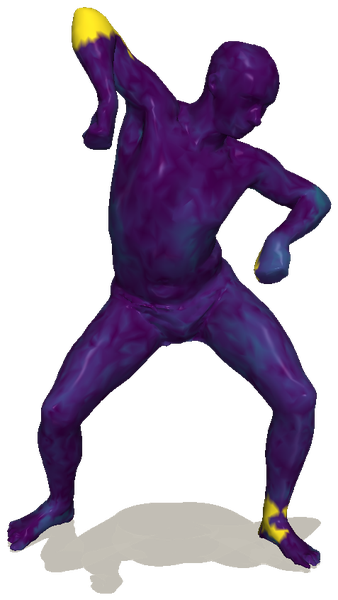}} & \imagespacingm
          \adjustbox{valign=m}{\includegraphics[height=\imageheightab]{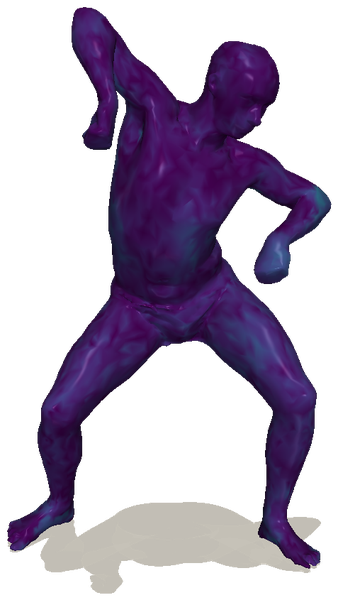}} \\

    \end{tabular}
    \caption{Results of intrinsic symmetry detection visualized as a mirrored matching of a shape to itself. Our method produces noticeably better quality than $\chi$ \cite{wang2025symmetry} both with and without refinement. Examples are chosen to reflect $\chi$'s failure modes.}
    \label{fig:intrinsic_symmetry}
\end{figure}

\paragraph{Baselines.}

We compare performance of our disentangled descriptors with other shape descriptors. For a detailed evaluation of our framework, we include three different baselines. All methods first cluster the vertices into two sets; the clustering strategy differs across methods as follows:
\begin{itemize}
    \item DINO+SD+GT: We cluster vertices using ground-truth left/right annotations.
    \item DINO+SD+$\chi$: We use $\chi$~\cite{wang2025symmetry} features for clustering.
    \item DINO+SD+($\chi$+refine): We use $\chi$~\cite{wang2025symmetry} features refined with Eq.~\ref{eq:map-mrf} for clustering.
\end{itemize}
Then we use descriptors constructed following Diff3F~\cite{dutt2024diffusion} pipeline with concatenated DINO-V2~\cite{oquab2023dinov2} and StableDiffusion~\cite{rombach2022high} features to match between these two sets using cosine similarity.

\paragraph{Results.}

From Tab.~\ref{tab:intrinsic_symmetry}, superiority of our disentangled descriptors is three-fold. First, comparison of DINO+SD+GT with Ours+GT %
(We use ground-truth for clustering and then $\mathcal{F}^{\text{S}}$ for matching)
shows that our disentangled symmetry-agnostic descriptor $\mathcal{F}^{\text{S}}$ filters out intrinsic symmetry information and leads to better detection results, which validates the necessity of $\mathcal{F}^{\text{S}}$. 
Secondly, comparison of DINO+SD+$\chi$ with Ours %
(We use $\mathcal{F}^{\text{C}}$ for clustering and then $\mathcal{F}^{\text{S}}$ for matching)
shows that without ground truth left/right annotations, our disentangled descriptors surpasses previous work $\chi$~\cite{wang2025symmetry}, validating the usefulness of our framework in an unsupervised setting. Finally, the comparison of DINO+SD+$\chi$ with DINO+SD+($\chi$+refine) or the comparison of Ours with Ours + refine validates the effectiveness of our refinement technique onto symmetry-informative descriptors. Fig.~\ref{fig:intrinsic_symmetry} shows qualitative results consistent with our conclusions.

\subsection{Left/right classification}
\label{sec:left_right} 

\begin{table*}[tbh]
  \centering
  \begin{tabular}{@{}lccccccccc@{}}
    \toprule
    Train & \textbf{BeCoS} & \multicolumn{2}{c}{\textbf{BeCoS\textsubscript{-h}}} & \multicolumn{2}{c}{\textbf{BeCoS\textsubscript{-a}}} & \multicolumn{2}{c}{\textbf{FAUST}} & \multicolumn{2}{c}{\textbf{SMAL}} \\
    \cmidrule(r){2-2}
    \cmidrule(r){3-4}
    \cmidrule(r){5-6}
    \cmidrule(r){7-8}
    \cmidrule(r){9-10}
    Test & \textbf{BeCoS} & \textbf{BeCoS\textsubscript{-h}} & \textbf{BeCoS\textsubscript{-a}} & \textbf{BeCoS\textsubscript{-h}} & \textbf{BeCoS\textsubscript{-a}} & \textbf{FAUST} & \textbf{SCAPE} & \textbf{SMAL} & \textbf{TOSCA} \\
    \midrule
    \citet{liu2012finding} & 79.98 & 79.83 & 80.46 & 79.83 & 80.46 & 90.45 & 80.84 & 75.71 & 72.88 \\
    \midrule
    $\chi$~\cite{wang2025symmetry} & \underline{92.51} & 94.25 & 84.97 & 92.16 & \underline{91.63} & 95.56 & 96.14 & 96.49 & 88.14 \\
    Ours & 91.65 & 94.13 & 85.09 & 92.18 & 90.96 & \textbf{96.28} & \underline{96.90} & \underline{96.72} & \underline{91.12} \\
    \midrule
    $\chi$~\cite{wang2025symmetry} + refine & \textbf{93.04} & \textbf{94.69} & \underline{86.24} & \underline{93.27} & \textbf{92.05} & 95.73 & 96.53 & 96.58 & 89.91 \\
    Ours + refine & 92.03 & \underline{94.49} & \textbf{87.27} & \textbf{93.64} & 91.17 & \underline{96.26} & \textbf{97.29} & \textbf{96.83} & \textbf{91.74} \\
    \bottomrule
  \end{tabular}
  \caption{Left/right classification accuracy ($acc_{\text{L/R}}$) results. The best and second-best results for each case are written in bold and underlined, respectively. The refinement step consistently leads to higher accuracies, while our features result in similar accuracies to $\chi$~\cite{wang2025symmetry}.}
  \label{tab:left_right_classification}
\end{table*}

\paragraph{Metrics.}

Given a set of shapes $X = \lbrace \mathcal{X}_1, ...,\mathcal{X}_N \rbrace$, where each shape $\mathcal{X}_n$ has a vertex set $V_{\mathcal{X}_n}$, we use the same metrics defined in $\chi$~\cite{wang2025symmetry} to compute the left/right classification accuracy:
\begin{equation}
    \text{acc}_{\text{L/R}} = \max \lbrace \text{hit}, 1-\text{hit} \rbrace,
    \label{eq:acc_lr}
\end{equation}
where 
\begin{equation}
    \text{hit} = \frac{1}{N} \sum_{n=1}^N \frac{1}{\vert V_{\mathcal{X}_n} \vert} \sum_{v \in V_{\mathcal{X}_n}} \mathbbm{1}(\text{sign}(\mathcal{F}_v^{\text{C}}) = \mathcal{F}_v^{gt}).
    \label{eq:hit}
\end{equation}

\begin{table}[tbh]
 \centering
\resizebox{\columnwidth}{!}{%
 
  \begin{tabular}{@{}ll|cccc@{}}
    \toprule
    Train & Test & $\chi$ \cite{wang2025symmetry} & Ours & \shortstack{$\chi$ \cite{wang2025symmetry} \\ + refine} & \shortstack{Ours \\ + refine} \\
    \cline{1-6}
    \textbf{BeCoS} & \textbf{BeCoS}              & 10.96 & 11.07 & 2.58 & 2.22 \\
    \cline{1-6}
    \multirow{2}{*}{\textbf{BeCoS\textsubscript{-h}}} & \textbf{BeCoS\textsubscript{-h}} & 9.19 & 9.80 & 2.53 & 2.34 \\
    & \textbf{BeCoS\textsubscript{-a}} & 21.51 & 21.94 & 3.86 & 2.90 \\
    \cline{1-6}
    \multirow{2}{*}{\textbf{BeCoS\textsubscript{-a}}} & \textbf{BeCoS\textsubscript{-h}} & 15.25 & 15.33 & 3.06 & 2.73 \\
    & \textbf{BeCoS\textsubscript{-a}} & 10.04 & 9.42 & 2.48 & 2.23 \\
    \cline{1-6}
    \multirow{2}{*}{\textbf{FAUST}} & \textbf{FAUST}              & 4.30 & 3.60 & 2.60 & 2.20 \\
    & \textbf{SCAPE} & 6.88 & 6.88 & 2.63 & 2.38 \\
    \cline{1-6}
    \multirow{2}{*}{\textbf{SMAL}} & \textbf{SMAL}               & 4.63 & 4.50 & 2.00 & 2.00 \\
     & \textbf{TOSCA}              & 51.02 & 42.28 & 3.91 & 2.40 \\
    \bottomrule
  \end{tabular}
  }
  \caption{Average number of connected components for different datasets and methods. The refinement step is able to effectively reduce the number of connected components of our solutions.}
  \label{tab:connected_components}
\end{table}

\noindent $\mathcal{F}_v^{\text{C}}$ is symmetry-informative descriptor for vertex $v \in V_{\mathcal{X}_n}$ of each shape $\mathcal{X}_n$, $\mathbbm{1}$ is the indicator function, and $\mathcal{F}_v^{gt}$ is the ground truth left/right annotation of vertex $v \in V_{\mathcal{X}_n}$.

\paragraph{Baselines.}

We use disentangled descriptor $\mathcal{F}^{\text{C}}$ to perform left/right classification, and compare with descriptors given by $\chi$~\cite{wang2025symmetry}, and refined descriptors from $\chi$~\cite{wang2025symmetry} using our proposed refinement technique. 
An axiomatic method \citet{liu2012finding} that extracts closed symmetric curves based on self-correspondence map prediction is also included.

\paragraph{Results.}

Tab.~\ref{tab:left_right_classification} summarizes left/right classification results on different datasets. Although our disentangled symmetry-informative feature $\mathcal{F}^{\text{C}}$ achieves comparable results with $\chi$~\cite{wang2025symmetry}, %
we can conclude that our proposed refinement technique indeed improves symmetry-informative descriptors. In addition, for each shape in every dataset, we first do a 2-center clustering using symmetry-informative descriptor, and then compute the number of connected components of the shape, and finally average this number over the whole dataset. Results shown in Tab.~\ref{tab:connected_components} indeed validate the effectiveness of our proposed refinement technique again. Fig.~\ref{fig:left_right_classification} also indicates consistent conclusions qualitatively.

\newcommand{\imageheightae}{0.07\textheight}
\newcommand{\imageheightah}{0.09\textheight}
\newcommand{\imageheightba}{0.06\textheight}
\newcommand{\imageheightbb}{0.08\textheight}

\begin{figure}
    \centering
    \begin{tabular}{ccc}
        Ground truth & \citet{liu2012finding} & $\chi$~\cite{wang2025symmetry} \\

          \adjustbox{valign=m}{\includegraphics[height=\imageheightbb]{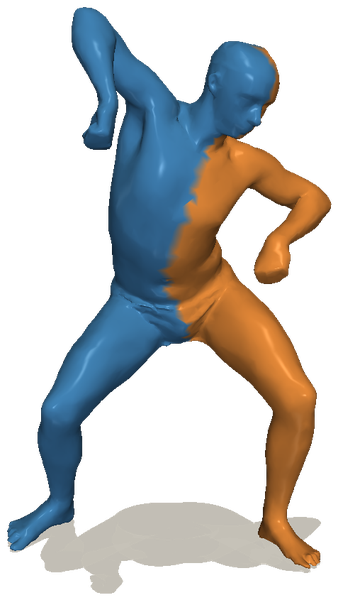}} &
          
        \adjustbox{valign=m}{
        \begin{tikzpicture}
            \node[anchor=south west,inner sep=0] (img1) 
                {\includegraphics[height=\imageheightbb]{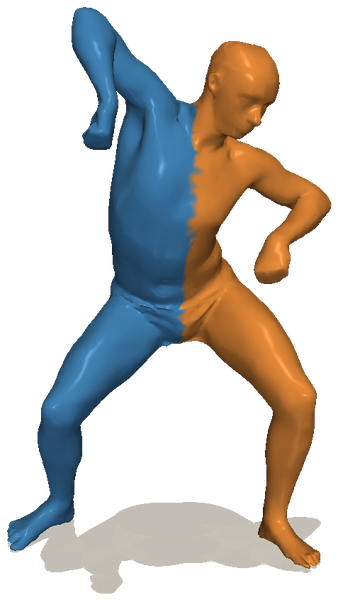}};
            \draw[red, thick] (0.65,1.5) circle (0.2); 
        \end{tikzpicture}
        } &
          
          \adjustbox{valign=m}{\includegraphics[height=\imageheightbb]{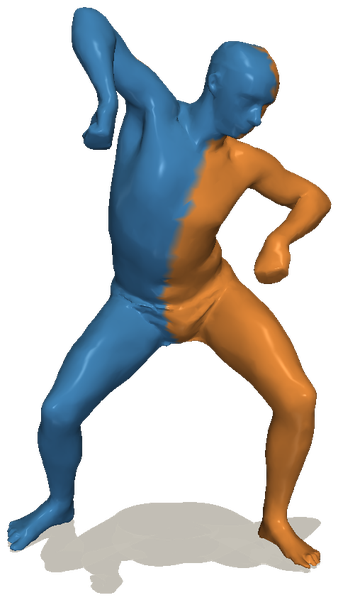}} \\
          
          \adjustbox{valign=m}{\includegraphics[height=\imageheightba]{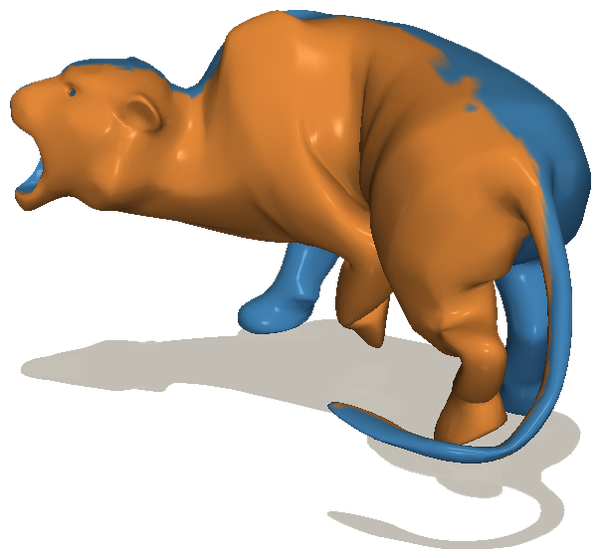}} &
          \adjustbox{valign=m}{
            \begin{tikzpicture}
                \node[anchor=south west,inner sep=0] (img1) 
                    {\includegraphics[height=\imageheightba]{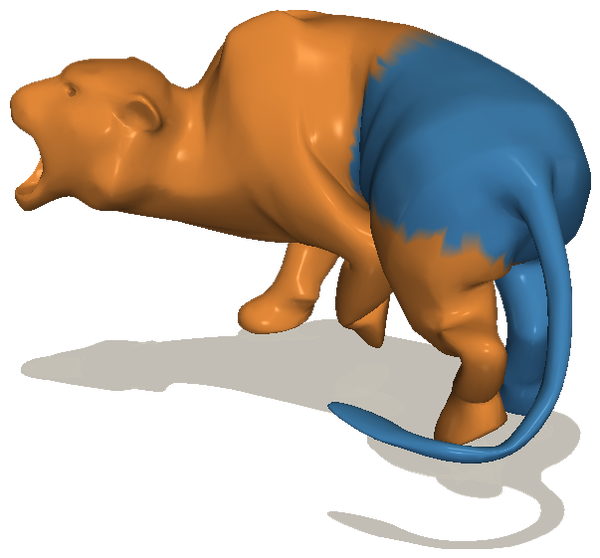}};
                \draw[red, thick] (1.0,1.0) circle (0.3); 
            \end{tikzpicture}
            } &
          \adjustbox{valign=m}{
            \begin{tikzpicture}
                \node[anchor=south west,inner sep=0] (img1) 
                    {\includegraphics[height=\imageheightba]{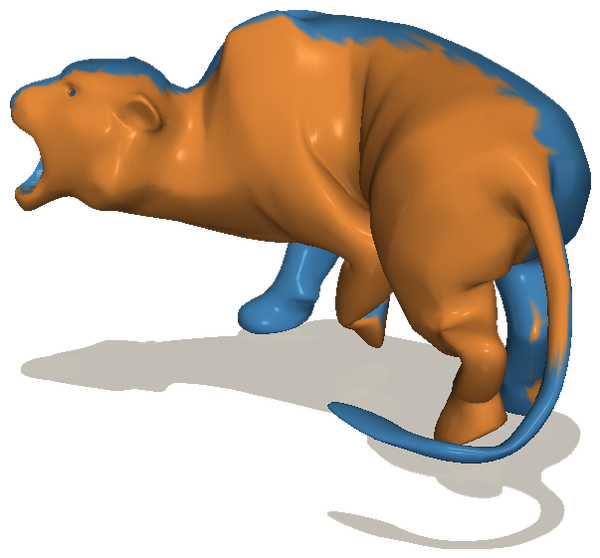}};
                \draw[red, thick] (1.3,0.6) circle (0.1); 
                \draw[red, thick] (1.3,0.40) circle (0.1); 
                \draw[red, thick] (0.90,0.56) circle (0.1); 
            \end{tikzpicture}
            } \\

           $\chi$~\cite{wang2025symmetry} + Refine & Ours & Ours + Refine \\
          \adjustbox{valign=m}{\includegraphics[height=\imageheightbb]{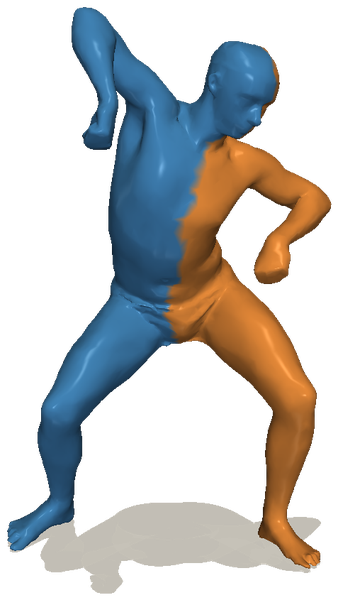}} &
          \adjustbox{valign=m}{\includegraphics[height=\imageheightbb]{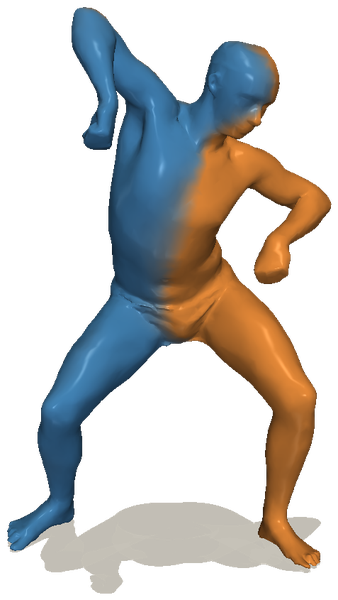}} &
          \adjustbox{valign=m}{\includegraphics[height=\imageheightbb]{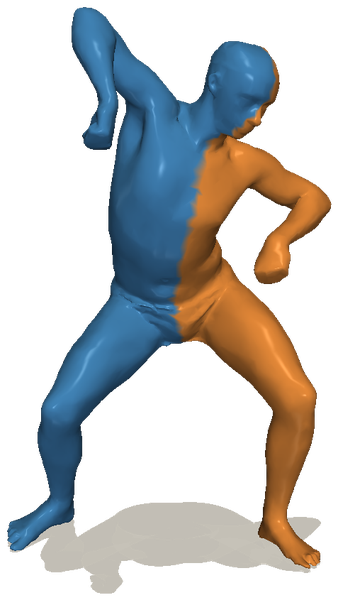}} \\
          \adjustbox{valign=m}{\includegraphics[height=\imageheightba]{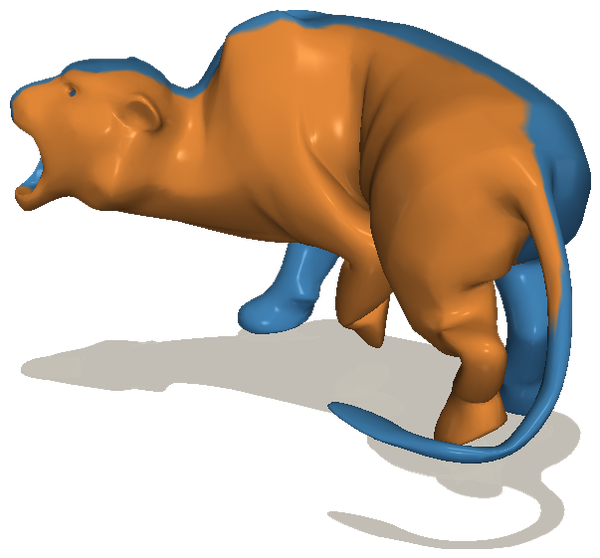}} &
          \adjustbox{valign=m}{
            \begin{tikzpicture}
                \node[anchor=south west,inner sep=0] (img1) 
                    {\includegraphics[height=\imageheightba]{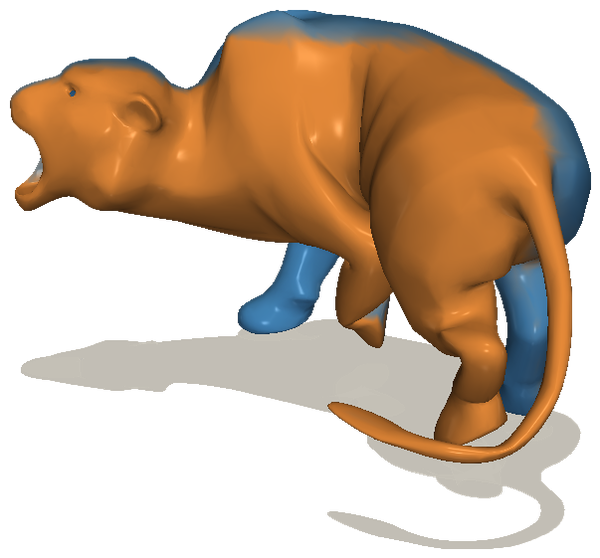}};
                \draw[red, thick] (0.90,0.56) circle (0.1); 
            \end{tikzpicture}
            } &
          \adjustbox{valign=m}{\includegraphics[height=\imageheightba]{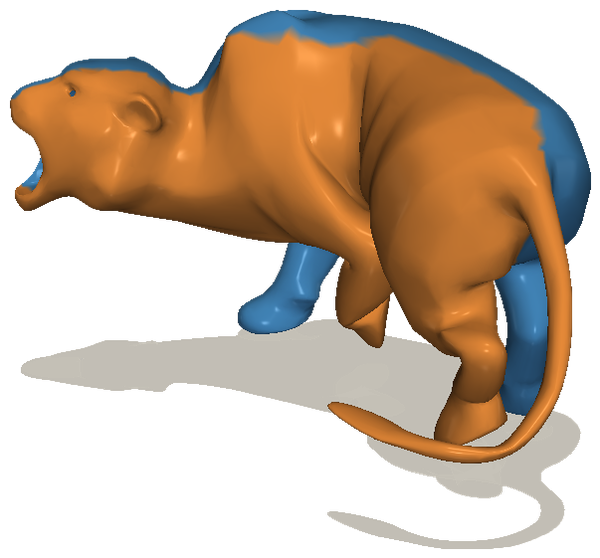}} \\

    \end{tabular}
    \caption{Qualitative results of left/right classification. Both $\chi$ \cite{wang2025symmetry} and our method achieve high accuracy. The refining step is able to reduce the number of incorrectly classified patches significantly.%
    }
    \label{fig:left_right_classification}
\end{figure}

\subsection{Shape matching}
\label{sec:matching}

\begin{table*}[tbh]
\setlength{\tabcolsep}{3.5pt}
  \centering
  \begin{tabular}{@{}lccccccccc@{}}
    \toprule
    Train & \textbf{BeCoS} & \multicolumn{2}{c}{\textbf{BeCoS\textsubscript{-h}}} & \multicolumn{2}{c}{\textbf{BeCoS\textsubscript{-a}}} & \multicolumn{2}{c}{\textbf{FAUST}} & \multicolumn{2}{c}{\textbf{SMAL}} \\
    \cmidrule(r){2-2}
    \cmidrule(r){3-4}
    \cmidrule(r){5-6}
    \cmidrule(r){7-8}
    \cmidrule(r){9-10}
    Test & \textbf{BeCoS} & \textbf{BeCoS\textsubscript{-h}} & \textbf{BeCoS\textsubscript{-a}} & \textbf{BeCoS\textsubscript{-h}} & \textbf{BeCoS\textsubscript{-a}} & \textbf{FAUST} & \textbf{SCAPE} & \textbf{SMAL} & \textbf{TOSCA} \\
    \midrule
    DINO+SD & 0.163 & 0.146 & 0.222 & 0.146 & 0.222 & 0.163 & 0.167 & 0.131 & 0.146 \\
    \midrule
    DINO+SD + $\chi$ \cite{wang2025symmetry} & 0.104 & 0.096 & \underline{0.123} & 0.117 & 0.106 & 0.050 & 0.063 & 0.060 & 0.102 \\
    Ours & \underline{0.079} & 0.091 & 0.159 & 0.134 & \underline{0.074} & \underline{0.056} & 0.054 & \underline{0.046} & \underline{0.095} \\
    \midrule
    DINO+SD + $\chi$ \cite{wang2025symmetry} + refine & 0.100 & \underline{0.087} & \textbf{0.122} & \textbf{0.094} & 0.104 & \textbf{0.048} & \textbf{0.048} & 0.062 & 0.108 \\
    Ours + refine & \textbf{0.075} & \textbf{0.082} & 0.157 & \underline{0.110} & \textbf{0.072} & 0.058 & \underline{0.049} & \textbf{0.045} & \textbf{0.094}\\
    \bottomrule
  \end{tabular}
  \caption{Geodesic error of shape matching using different features across datasets. Refinement generally enhances matching performance, with our complete framework achieving state-of-the-art results on several major datasets.}
  \label{tab:matching}
\end{table*}

\paragraph{Metrics.}

Following Diff3F~\cite{dutt2024diffusion} and $\chi$~\cite{wang2025symmetry}, we use average matching error as metrics. For a source shape $\mathcal{X}$ and a target shape $\mathcal{Y}$, represented with vertex sets $V_{\mathcal{X}} \in \mathbb{R}^{|V_{\mathcal{X}}| \times 3}$ and  $V_{\mathcal{Y}} \in \mathbb{R}^{|V_{\mathcal{Y}}| \times 3}$, respectively, the matching error is 
\begin{equation}
    err_{\text{mat}} = \frac{1}{\vert V_{\mathcal{X}} \vert} \sum_{v \in V_{\mathcal{X}}} \text{dist}(f(v) - y_{gt}), 
\end{equation}
where $f(v)$ is the predicted matching point of $v \in V_{\mathcal{X}}$ in $\mathcal{Y}$, $y_{gt} \in V_{\mathcal{Y}}$ is the ground truth corresponding point of $v$, and $\text{dist}(\cdot, \cdot)$ is the geodesic distance normalized by the square root of the area of $\mathcal{Y}$, following~\cite{kim2011blended}. 

\paragraph{Baselines.}
As a matching baseline, we use pairwise cosine similarity of DINO+SD, DINO+SD concatenated to $\chi$ \cite{wang2025symmetry} and a refined version of $\chi$, respectively. %

\paragraph{Results.} Tab. \ref{tab:matching} summarizes shape matching results across multiple datasets and Fig. \ref{fig:matching_small} visualizes qualitative results across multiple example shape pairs. %
We observe that refining the features improves matching quality for both $\chi$ \cite{wang2025symmetry} and our method across most datasets. 
When combining our new method with our proposed refinement, we are able to achieve significantly better matching performance than the baseline $\chi$ method on most datasets, as visible in Fig.~\ref{fig:matching_small}.

\newcommand{\imageheightac}{0.07\textheight}

\begin{figure}
    \centering
    \begin{tabular}{cccc}
          Source & Target & $\chi$ \cite{wang2025symmetry} &  Ours + refine  \\
          
          \adjustbox{valign=m}{\includegraphics[height=\imageheightac]{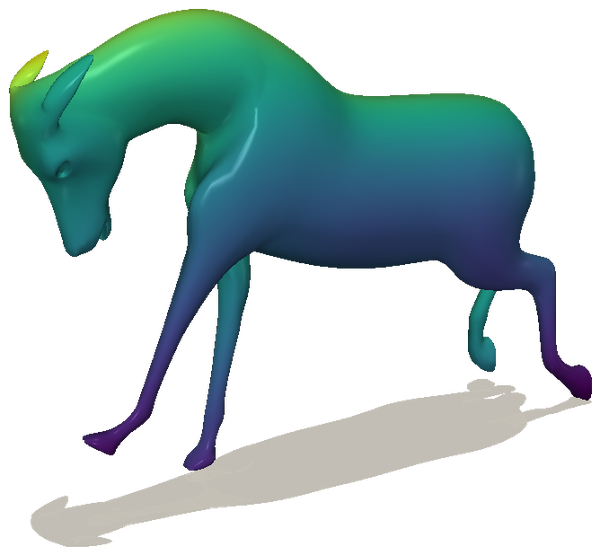}} &
          \adjustbox{valign=m}{\includegraphics[height=\imageheightac]{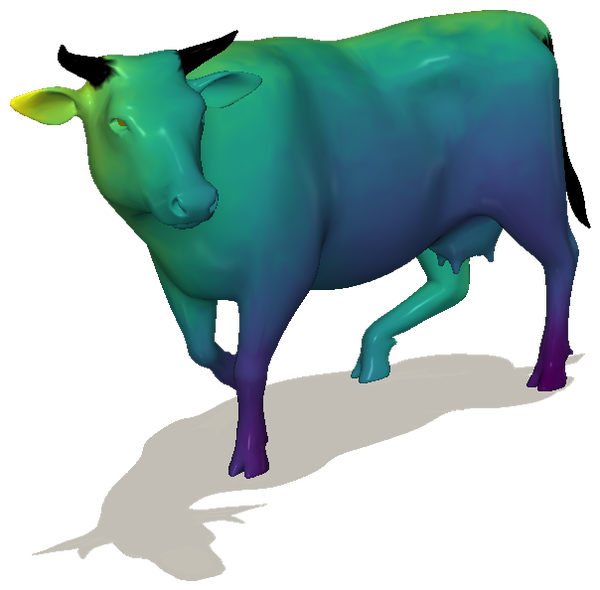}} &

          \adjustbox{valign=m}{
            \begin{tikzpicture}
                \node[anchor=south west,inner sep=0] (img1) 
                    {\includegraphics[height=\imageheightac]{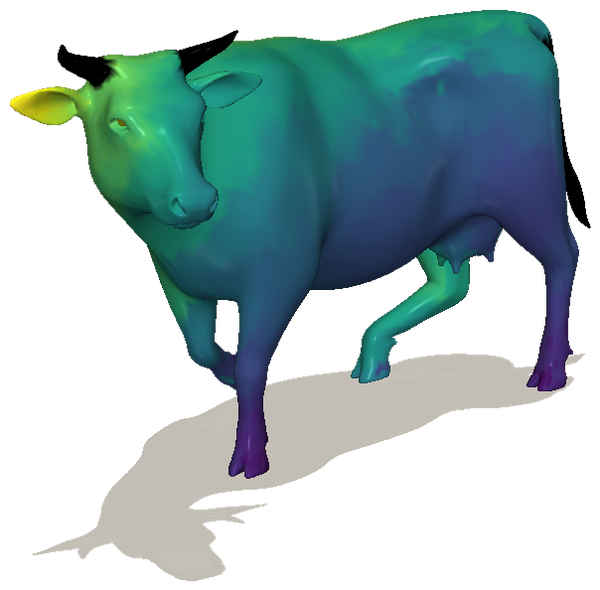}};
                \draw[red, thick] (1.3,1.2) circle (0.4); 
            \end{tikzpicture}
            } &
          
          \adjustbox{valign=m}{\includegraphics[height=\imageheightac]{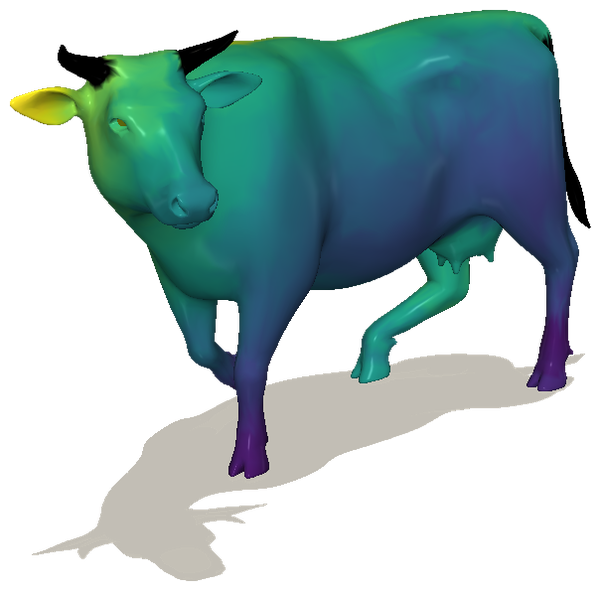}} \\

          \adjustbox{valign=m}{\includegraphics[height=0.05\textheight]{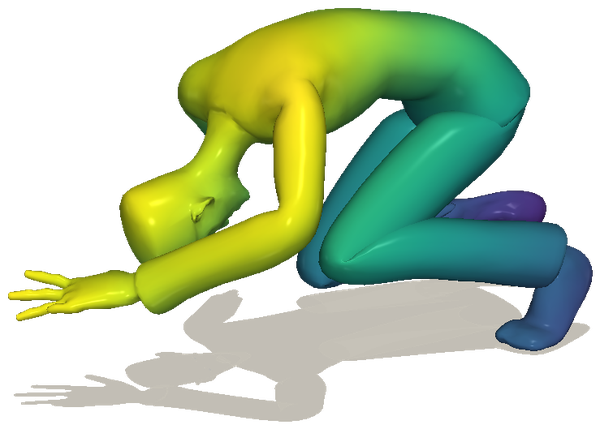}} &
          \adjustbox{valign=m}{\includegraphics[height=\imageheightac]{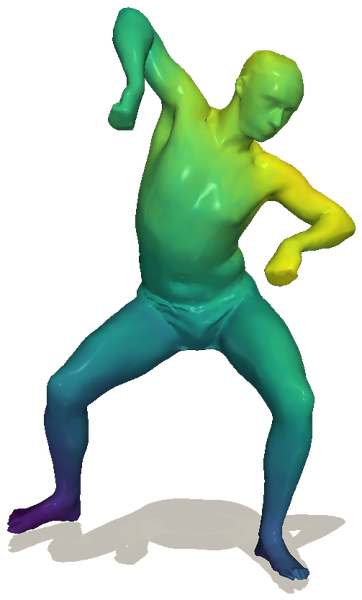}} &

          \adjustbox{valign=m}{
            \begin{tikzpicture}
                \node[anchor=south west,inner sep=0] (img1) 
                    {\includegraphics[height=\imageheightac]{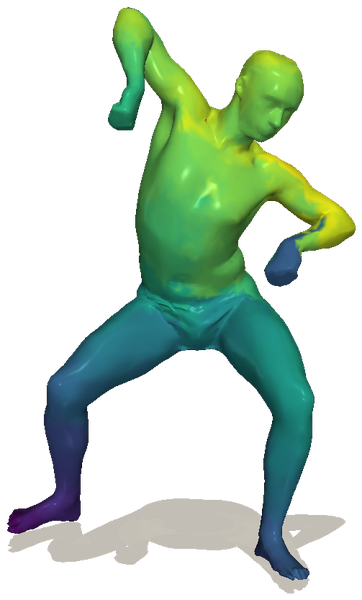}};
                \draw[red, thick] (0.85,0.9) circle (0.2); 
            \end{tikzpicture}
            } &
          
          \adjustbox{valign=m}{\includegraphics[height=\imageheightac]{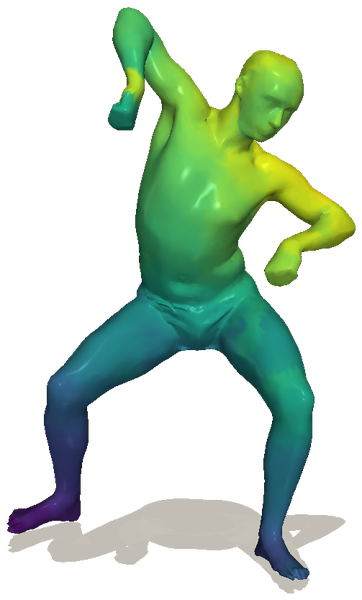}}
          \\

    \end{tabular}
    \caption{Shape matching results of two shape pairs from different categories, using $\chi$~\cite{wang2025symmetry} and our descriptors. Our method is able to create a smoother and more accurate matching.}
    \label{fig:matching_small}
\end{figure}

\subsection{Ablation study}
\label{sec:ablation}

We run ablation studies on FAUST~\cite{bogo2014faust} using left/right accuracy, intrinsic symmetry error, and shape matching error. We also compare with a modified $\chi$~\cite{wang2025symmetry}, denoted $\bar{\chi}$, by adding orthogonality constraints to its MLP as in our method, trained with $\chi$’s losses. Tab.~\ref{tab:ablation} shows that our loss combination yields a good balance of quality, informativeness, and symmetry invariance.

\begin{table}[tbh]
\setlength{\tabcolsep}{3.5pt}
  \centering
  \begin{tabular}{@{}lccccccccc@{}}
    \toprule
    w/o & $\mathcal{L}_{\text{dis}}$ & $\mathcal{L}_{\text{sim}}$ & $\mathcal{L}_{\text{rec}}$ &
    $\mathcal{L}_{\text{bou}}$ & $\mathcal{L}_{\text{con}}$ & $\bar{\chi}$ & Full \\
    \midrule
    acc$_{\text{L/R}}$ & 60.87 & 95.83 & 95.15 & 75.42 & \textbf{97.14} & 90.48 & \underline{96.28} \\
    err$_{\text{sym}}$ & 0.292 & 0.420 & \underline{0.052} & 0.116 & 0.071 & 0.132 & \textbf{0.042} \\
    err$_{\text{mat}}$ & 0.224 & 0.118 & \underline{0.056} & 0.096 & 0.143 & 0.080 & \textbf{0.056} \\
    \bottomrule
  \end{tabular}
  \caption{The ablation study on FAUST validates our overall loss combination across all metrics.}
  \label{tab:ablation}
\end{table}

\section{Limitations}

Although our proposed disentanglement framework shows remarkable performance in various shape analysis tasks on different datasets, it nevertheless has certain drawbacks. 
Similar to \cite{wang2025symmetry}, our method requires mesh connectivity information to regularize the training, restricting its application to certain representations of 3D shapes. In addition, we observe inaccuracies in both symmetry-informative and symmetry-agnostic descriptors for vertices for which two different parts of a shape are in close proximity. Furthermore, our proposed refinement method depends on the predicted, continuous symmetry-informative features and essentially discretizes these into binary values. Consequently, slight deviations of the continuous features could lead to large deviations of the discretized features. Moreover, although our method is theoretically independent of a shape's genus, our current experimental evaluation is restricted to genus-zero surfaces.

\section{Conclusion}
\label{sec:conclusion}

In this work, we propose an intrinsic symmetry aware feature disentanglement network, which takes per vertex semantic shape descriptors as input and outputs robust disentangled symmetry-informative and symmetry-agnostic descriptors. To this end, 
we propose a selection of unsupervised losses, combined with a symmetry-informative feature refinement technique based on Markov Random Field energy minimization, to guide our framework to successfully disentangle symmetry-informative and symmetry-agnostic information. Various experiments, including intrinsic symmetry detection, left/right classification and shape matching, validate the effectiveness of the 3D shape descriptors obtained with our approach.

\section*{Acknowledgments}
This work is supported by the ERC starting grant no.~101160648 (Harmony).

\newpage

{
    \small
    \bibliographystyle{ieeenat_fullname}
    \bibliography{main}
}
\clearpage
\setcounter{page}{1}
\maketitlesupplementary

In the following we provide additional information and details to our paper. 
In particular, we give additional details about the data generation for BeCoS~\cite{ehm2024beyond} dataset, discuss implementation details, provide further insights into our boundary loss and show additional qualitative results.

\section{BeCoS data generation}

$\chi$~\cite{wang2025symmetry} uses a subsampled version of BeCoS~\cite{ehm2024beyond} with a train/validation/test split consisting of 
990/137/142 shape pairs, we use the full benchmark containing
10185/274/284 shape pairs to train all models and to perform all evaluations.

\section{Implementation details}
\label{sec:implementation}
To extract the image features presented in Sec.~\ref{sec:feature}, we use a concatenation of DINO-V2~\cite{oquab2023dinov2} and StableDiffusion~\cite{rombach2022high} foundation models similar to the strategy used in Diff3F~\cite{dutt2024diffusion} and $\chi$~\cite{wang2025symmetry}. This strategy results in image features of dimension $d=3968$. 
The lightweight encoder and decoder described in Sec.~\ref{sec:feature} are implemented as two three-layer MLPs with hidden dimensions 3968. This results in $\mathcal{F}_v^{\text{C}}$ and $\mathcal{F}_v^{\text{S}}$ being of dimension 1 and 3967, respectively. To improve the architecture's autoencoding capabilities, we add skip-connections around the encoder and decoder respectively. %
Finally, we use $\lambda_1 = 1.0, \lambda_2 = 0.2, \lambda_3 = 10.0, \lambda_4 = 2.0$ as the loss coefficients for the overall loss presented in Sec.~\ref{sec:disentanglement}. These coefficients were chosen by hyperparameter search.%

\section{Further insight into our boundary loss \texorpdfstring{$\mathcal{L}_{\text{bou}}$}{L_bou}}
\label{sec: boundary_loss}

\begin{figure}[h!]
    \centering
    \begin{tabular}{c|ccc}
        &
        \adjustbox{valign=m}{\includegraphics[height=0.15\textwidth]{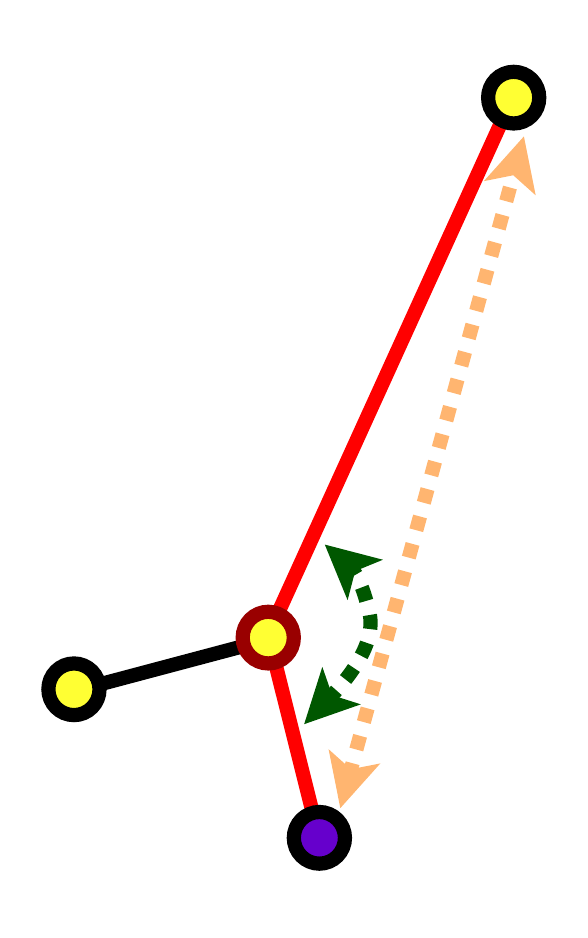}} & 
        \adjustbox{valign=m}{\includegraphics[height=0.15\textwidth]{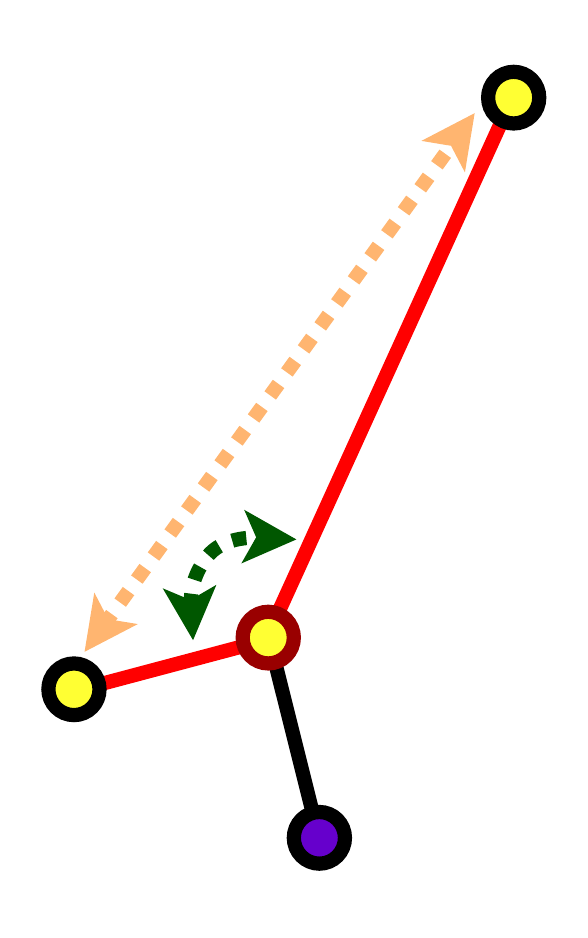}} & 
        \adjustbox{valign=m}{\includegraphics[height=0.15\textwidth]{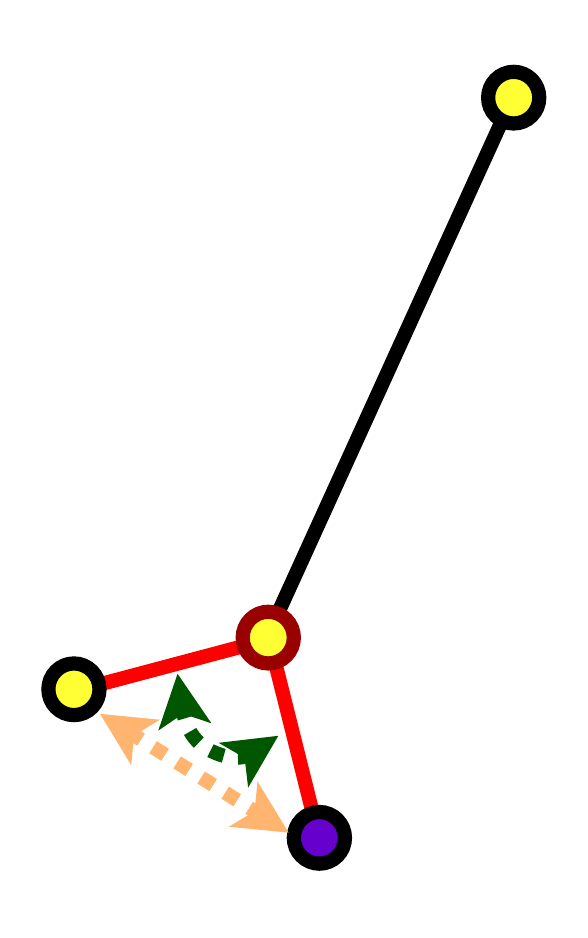}}
        \\
        \toprule

        $-C_v(u, w)$ & Low & Medium & High\\

        $L(u, v, w)$ & High & Low & Low \\
        \midrule

        $L-C_v$ & Medium & Low & Medium \\
        \bottomrule
    \end{tabular}
    \caption{For each vertex in the graph induced by the triangle mesh of a shape, we find a pair of edges which minimize the sum of $-C_v$ (green) and $L$ (orange). $-C_v$ is minimal if the edges are close to parallel, while $L(u, v, w)$ is minimal if $u, v $ and $w$ have the same symmetry-informative feature. A low $\mathcal{L}_{\text{bou}}$ loss implies that there exists a straight direction in which the symmetry-informative feature does not change.}
    \label{fig:boundary_loss}
\end{figure}

In the following, we provide more insights as well as intuitive motivation for our boundary loss $\mathcal{L}_{\text{bou}}$. We assume that the symmetric boundary generally and approximately is a straight line on the surface manifold of a shape. To leverage this knowledge during the training process, we introduce the $\mathcal{L}_{\text{bou}}$ loss. The idea is to regularize the symmetry-informative feature s.t. for each vertex $v \in V$ there should be at least one pair of approximately parallel edges along which the feature does not change.
An example can be seen in Fig.~\ref{fig:boundary_loss}.

\begin{figure}[h!]
    \centering
    \includegraphics[width=0.7\linewidth]{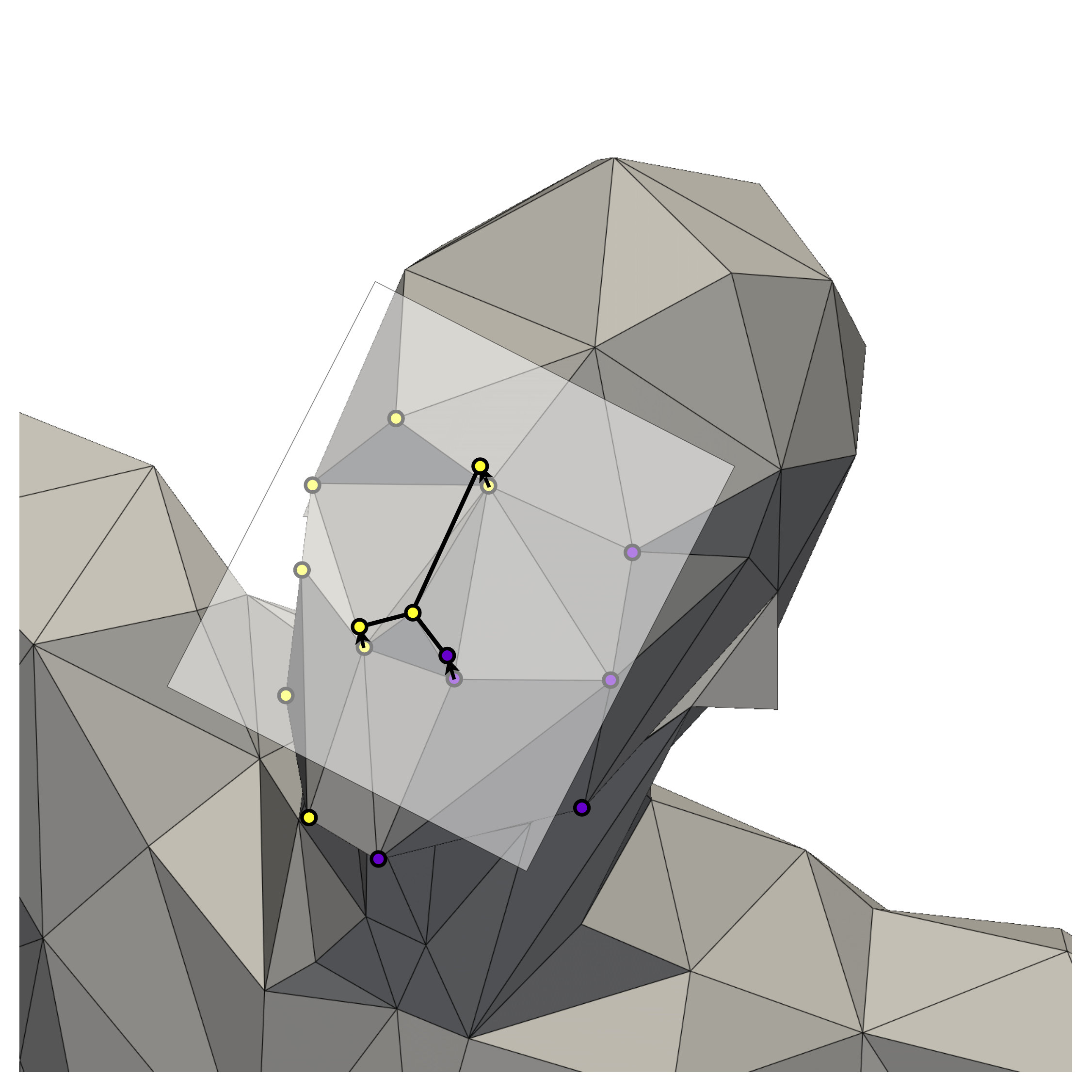}
    \caption{We project all neighbouring vertices onto the tangent space of the center vertex (i.e.~the vertex in the picture which is connected by three edges)  before loss computation such that our loss is independent of the surface curvature.}
    \label{fig:normal_projection}
\end{figure}

Since we want the symmetric boundary to be a straight line in the surface manifold of the mesh, we first project the neighbouring vertices of each vertex onto the tangent space, as shown in Fig.~\ref{fig:normal_projection}.

\section{Additional qualitative results}
\label{sec: additional_figures}

\begin{figure*}
    \centering
    \begin{tabular}{ccccc|cccc}
          & \multicolumn{4}{c}{Without Refinement} & \multicolumn{4}{c}{With Refinement} \\
          \rotatebox{90}{Source} & 
          \adjustbox{valign=m}{\includegraphics[height=\imageheightac]{./figures/images/matching_pred/cow/image_iccv_source.png}} &
          \adjustbox{valign=m}{\includegraphics[height=\imageheightac]{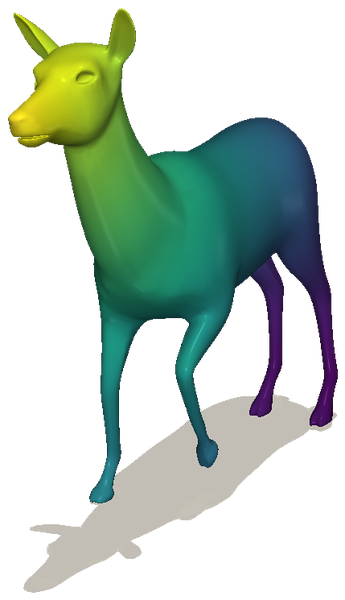}} &
          \adjustbox{valign=m}{\includegraphics[height=0.07\textheight]{./figures/images/matching_pred/human/image_iccv_source.png}} &
          \adjustbox{valign=m}{\includegraphics[height=\imageheightac]{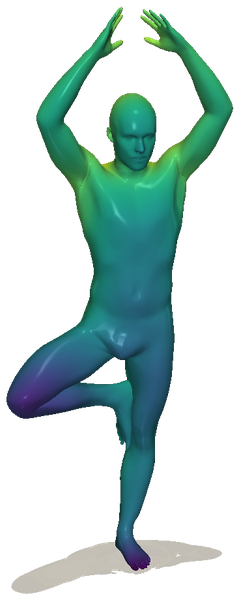}} & 
          \adjustbox{valign=m}{\includegraphics[height=\imageheightac]{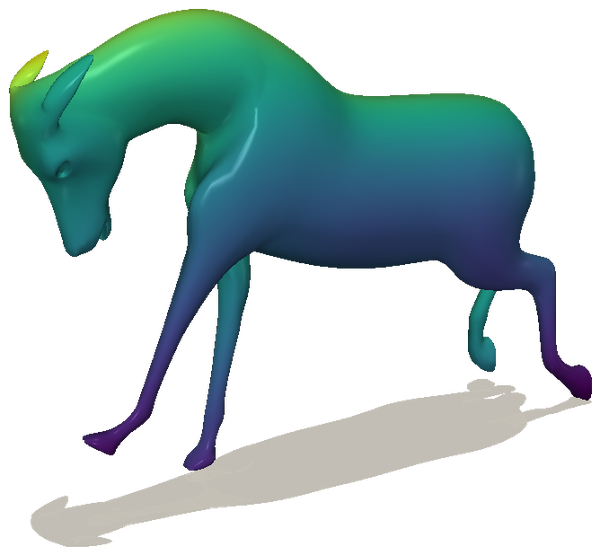}} &
          \adjustbox{valign=m}{\includegraphics[height=\imageheightac]{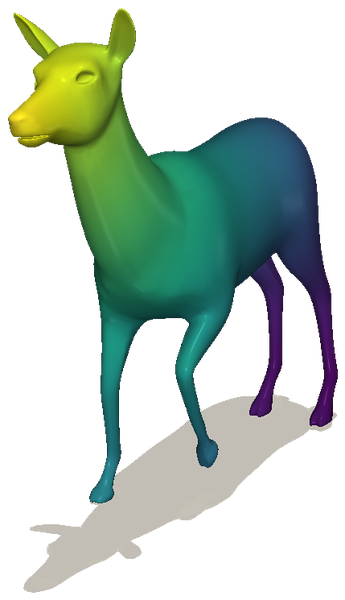}} &
          \adjustbox{valign=m}{\includegraphics[height=0.07\textheight]{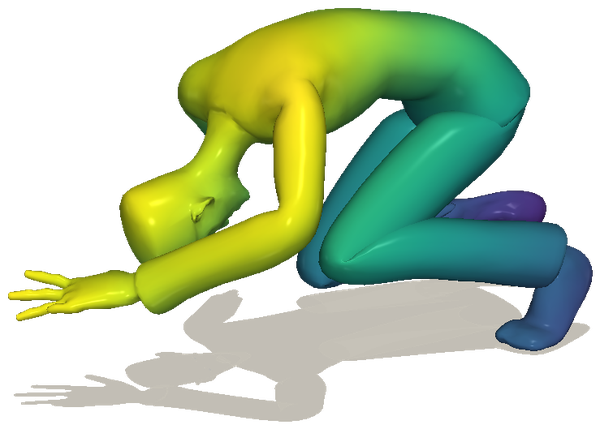}} &
          \adjustbox{valign=m}{\includegraphics[height=\imageheightac]{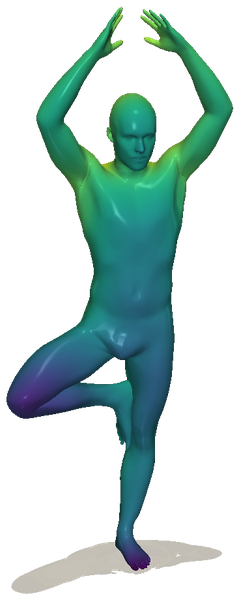}} \\

          \rotatebox{90}{Target} & 
          \adjustbox{valign=m}{\includegraphics[height=\imageheightac]{./figures/images/matching_pred/cow/image_iccv_target.png}} &
          \adjustbox{valign=m}{\includegraphics[height=\imageheightac]{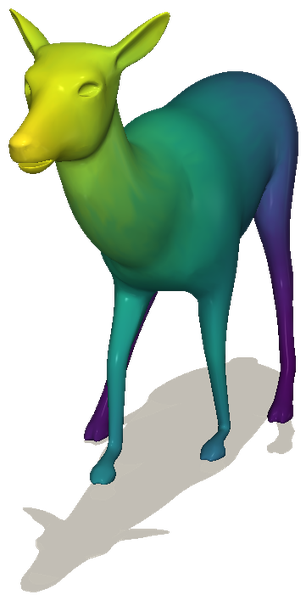}} &
          \adjustbox{valign=m}{\includegraphics[height=\imageheightac]{./figures/images/matching_pred/human/image_iccv_target.png}} &
          \adjustbox{valign=m}{\includegraphics[height=\imageheightac]{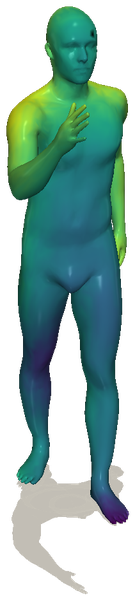}} & 
          \adjustbox{valign=m}{\includegraphics[height=\imageheightac]{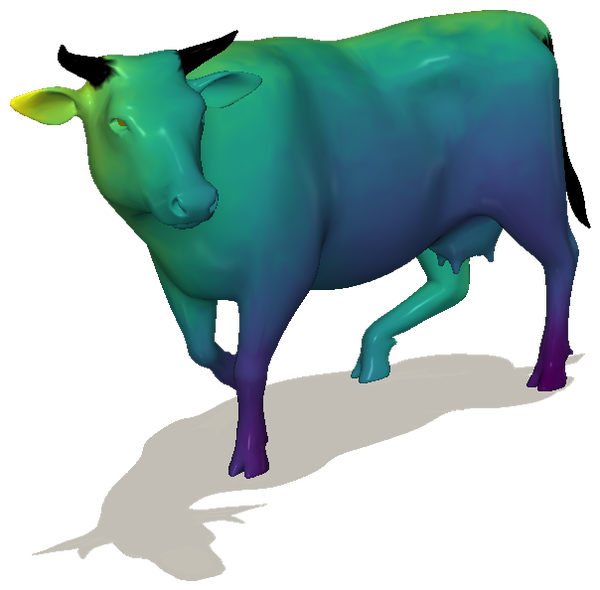}} &
          \adjustbox{valign=m}{\includegraphics[height=\imageheightac]{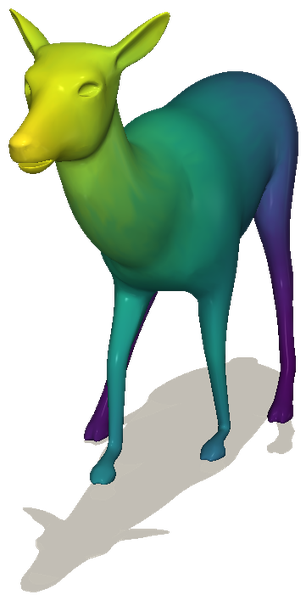}} &
          \adjustbox{valign=m}{\includegraphics[height=\imageheightac]{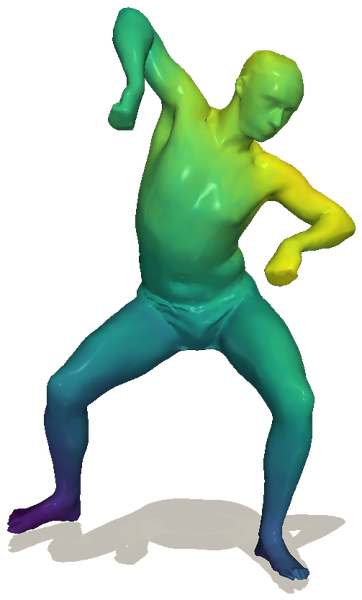}} &
          \adjustbox{valign=m}{\includegraphics[height=\imageheightac]{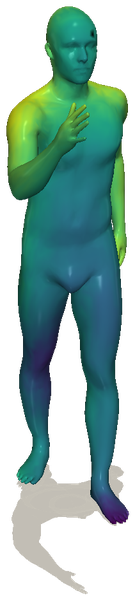}} \\

          \midrule \\

          \rotatebox{90}{$\chi$ \cite{wang2025symmetry}} & 
          \adjustbox{valign=m}{\includegraphics[height=\imageheightac]{./figures/images/matching_pred/cow/image_iccv_result.png}} &
          \adjustbox{valign=m}{\includegraphics[height=\imageheightac]{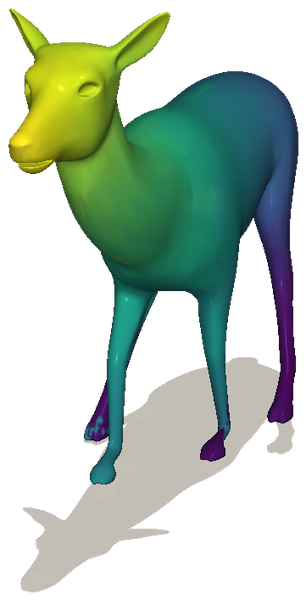}} &
          \adjustbox{valign=m}{\includegraphics[height=\imageheightac]{./figures/images/matching_pred/human/image_iccv_result.png}} &
          \adjustbox{valign=m}{\includegraphics[height=\imageheightac]{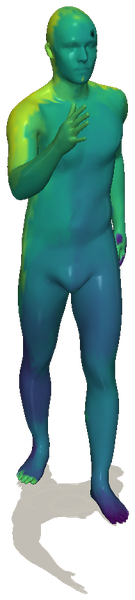}} & 
          \adjustbox{valign=m}{\includegraphics[height=\imageheightac]{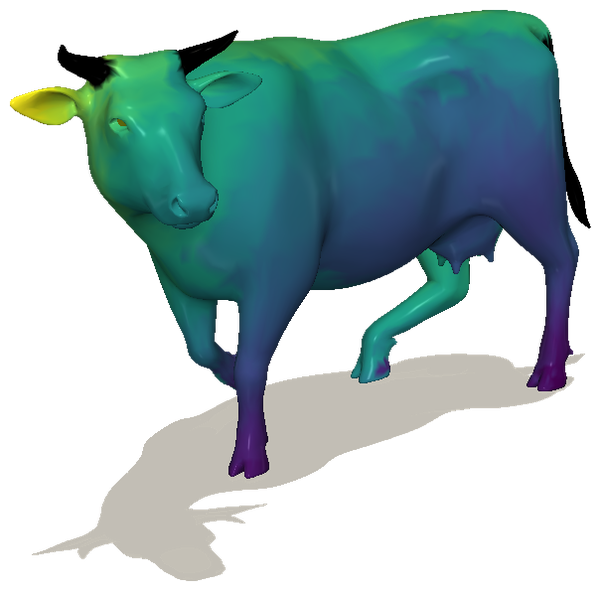}} &
          \adjustbox{valign=m}{\includegraphics[height=\imageheightac]{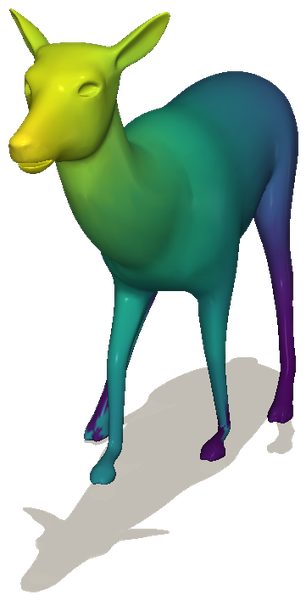}} &
          \adjustbox{valign=m}{\includegraphics[height=\imageheightac]{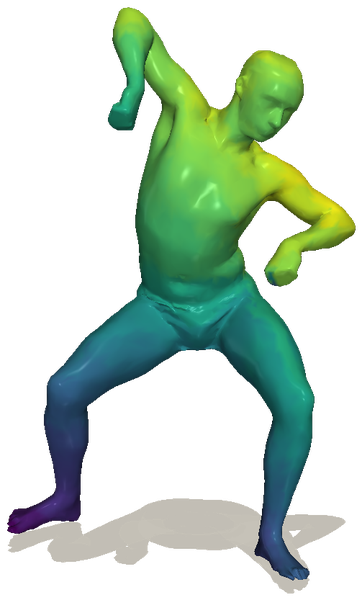}} &
          \adjustbox{valign=m}{\includegraphics[height=\imageheightac]{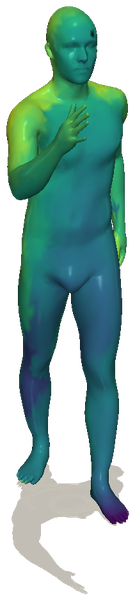}} \\

          \rotatebox{90}{\shortstack{$\chi$ \cite{wang2025symmetry} \\ (Error)}} & 
          \adjustbox{valign=m}{\includegraphics[height=\imageheightac]{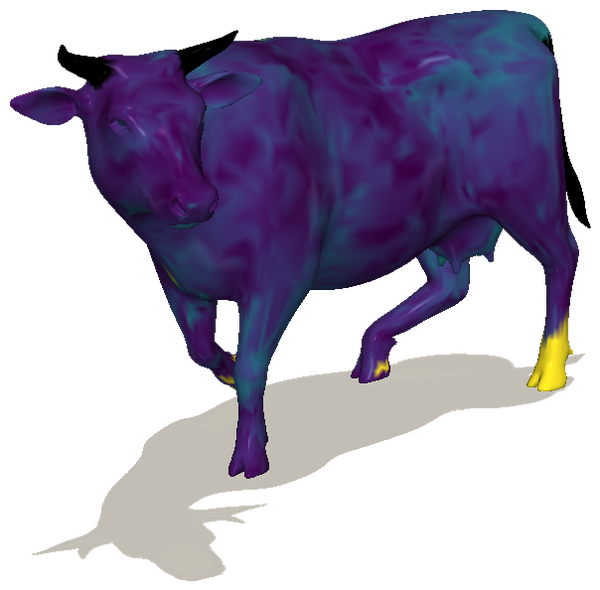}} &
          \adjustbox{valign=m}{\includegraphics[height=\imageheightac]{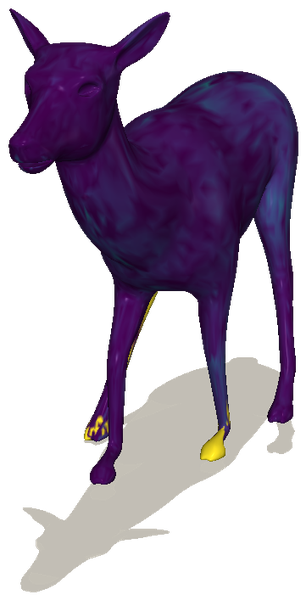}} &
          \adjustbox{valign=m}{\includegraphics[height=\imageheightac]{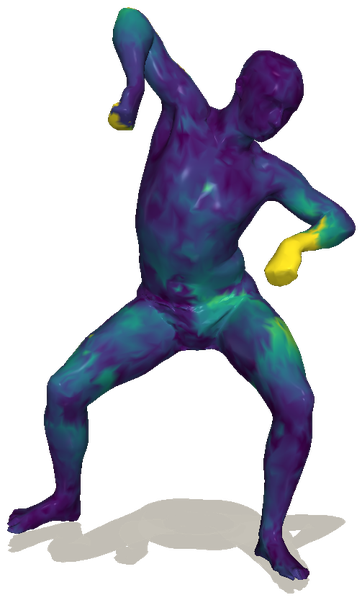}} &
          \adjustbox{valign=m}{\includegraphics[height=\imageheightac]{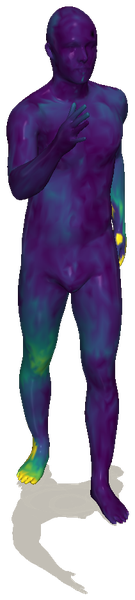}} & 
          \adjustbox{valign=m}{\includegraphics[height=\imageheightac]{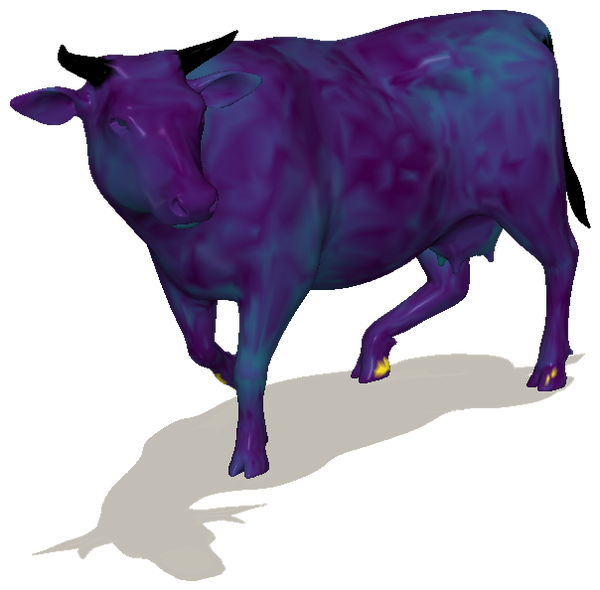}} &
          \adjustbox{valign=m}{\includegraphics[height=\imageheightac]{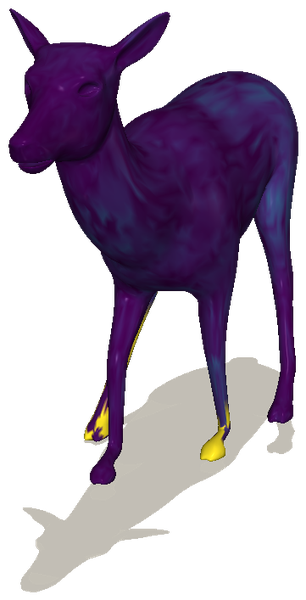}} &
          \adjustbox{valign=m}{\includegraphics[height=\imageheightac]{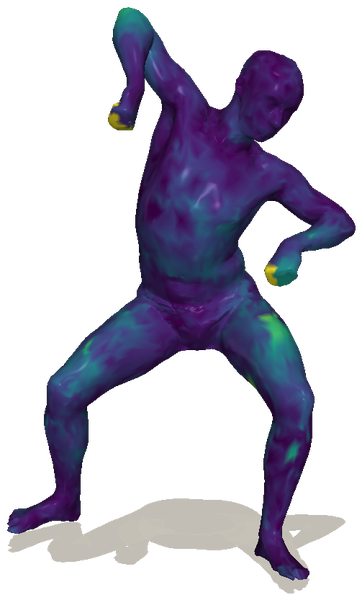}} &
          \adjustbox{valign=m}{\includegraphics[height=\imageheightac]{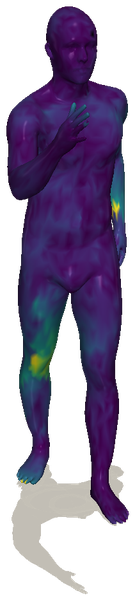}} \\

          \rotatebox{90}{Ours} & 
          \adjustbox{valign=m}{\includegraphics[height=\imageheightac]{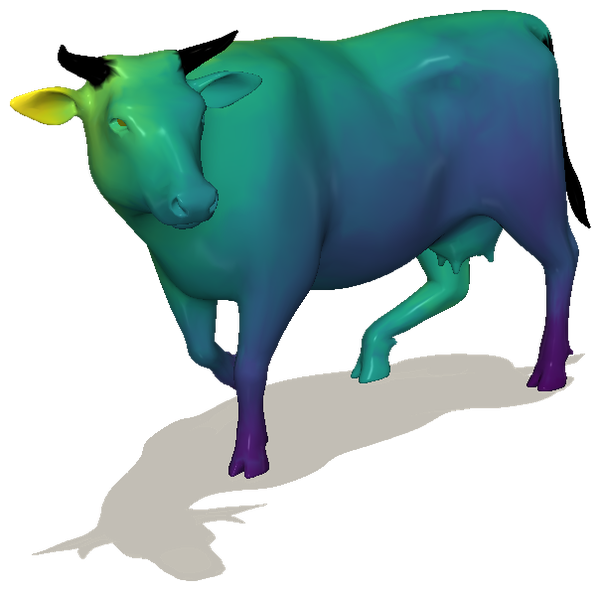}} &
          \adjustbox{valign=m}{\includegraphics[height=\imageheightac]{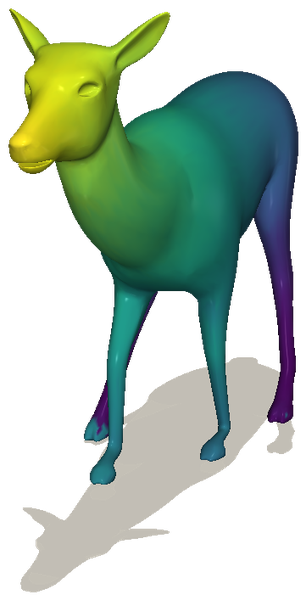}} &
          \adjustbox{valign=m}{\includegraphics[height=\imageheightac]{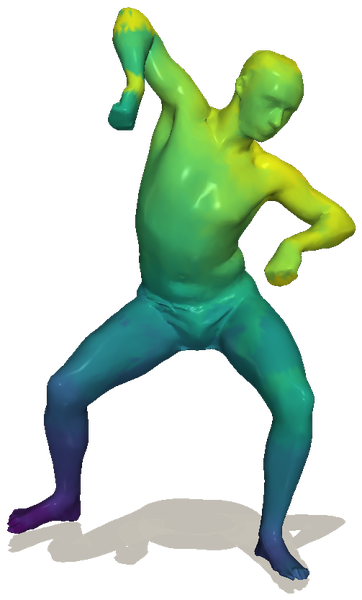}} &
          \adjustbox{valign=m}{\includegraphics[height=\imageheightac]{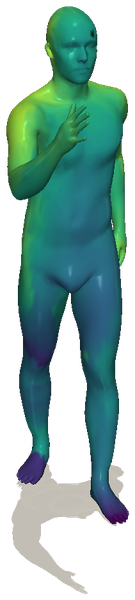}} & 
          \adjustbox{valign=m}{\includegraphics[height=\imageheightac]{./figures/images/matching_post/cow/image_3dv_result.png}} &
          \adjustbox{valign=m}{\includegraphics[height=\imageheightac]{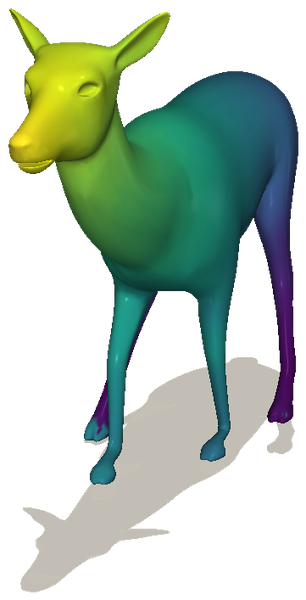}} &
          \adjustbox{valign=m}{\includegraphics[height=\imageheightac]{./figures/images/matching_post/human/image_3dv_result.png}} &
          \adjustbox{valign=m}{\includegraphics[height=\imageheightac]{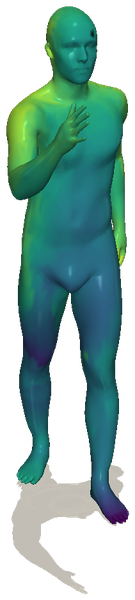}} \\

          \rotatebox{90}{\shortstack{Ours \\ (Error)}} & 
          \adjustbox{valign=m}{\includegraphics[height=\imageheightac]{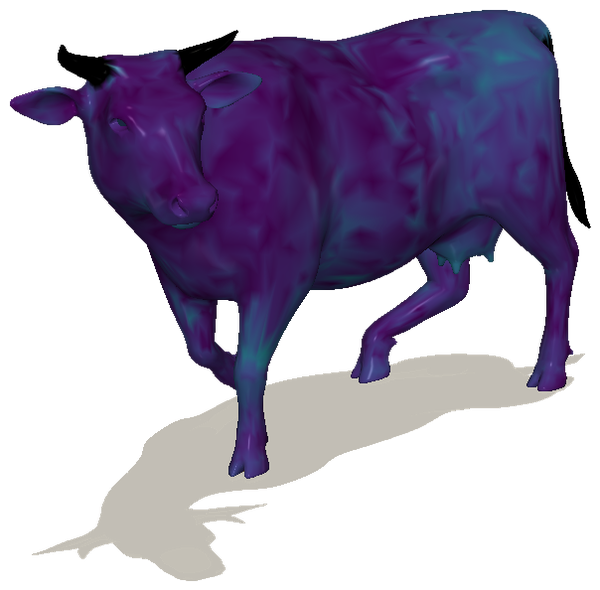}} &
          \adjustbox{valign=m}{\includegraphics[height=\imageheightac]{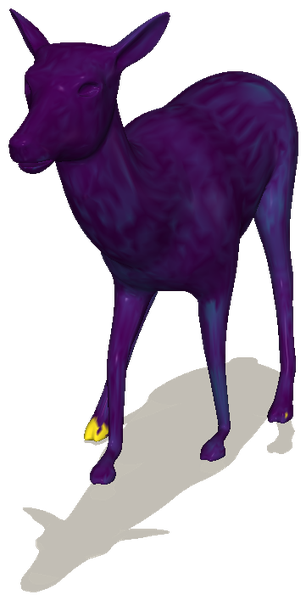}} &
          \adjustbox{valign=m}{\includegraphics[height=\imageheightac]{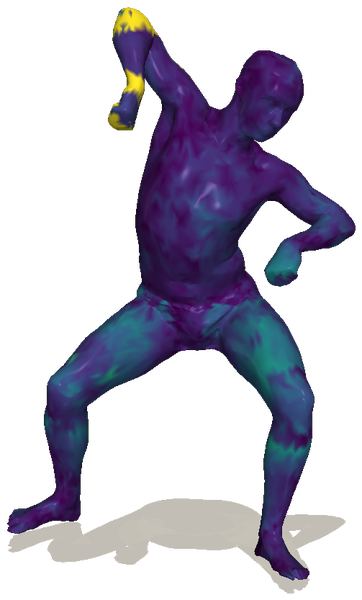}} &
          \adjustbox{valign=m}{\includegraphics[height=\imageheightac]{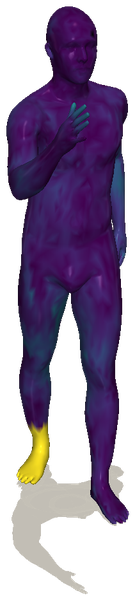}} & 
          \adjustbox{valign=m}{\includegraphics[height=\imageheightac]{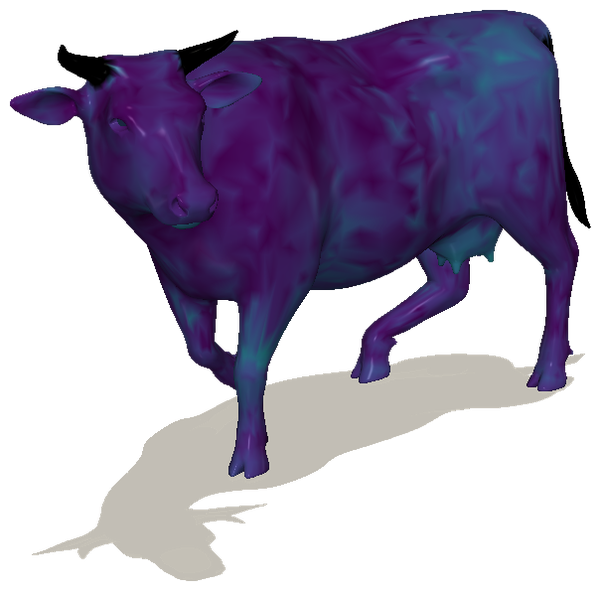}} &
          \adjustbox{valign=m}{\includegraphics[height=\imageheightac]{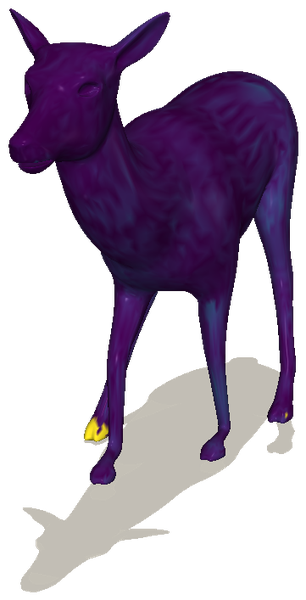}} &
          \adjustbox{valign=m}{\includegraphics[height=\imageheightac]{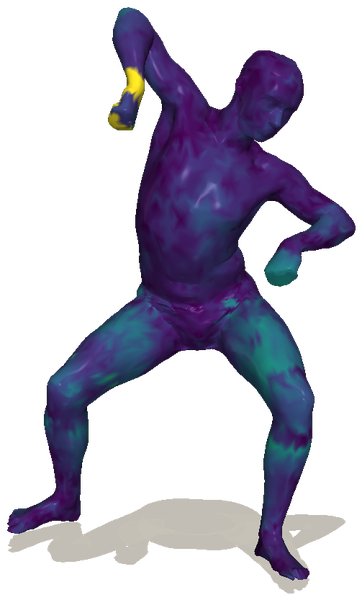}} &
          \adjustbox{valign=m}{\includegraphics[height=\imageheightac]{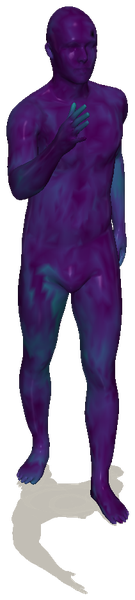}} \\

    \end{tabular}
    \caption{Qualitative results of shape matching.}
    \label{fig: supple_matching}
\end{figure*}

We include more qualitative results here for intrinsic symmetry detection (Fig.~\ref{fig: supple_intrinsic_symmetry}), left/right classification (Fig.~\ref{fig: supple_left/right_classification}) and shape matching (Fig.~\ref{fig: supple_matching}), to better illustrate the superiority of our proposed framework against other methods.

In \cref{fig: supple_intrinsic_symmetry}, we provide, to \cref{fig:intrinsic_symmetry} additional, visualizations of intrinsic symmetries detected with our method and competitors.
In addition, we also include qualitative results of another setting ($\chi$~\cite{wang2025symmetry} (gt) and ours (gt)). We use ground truth left/right annotations to first cluster vertices of a shape into two sets. %
Then for each vertex, we find its intrinsic symmetry counterpart with different cluster label. 
We show results in Fig.~\ref{fig: supple_intrinsic_symmetry} to further confirm the superiority of our disentangled features compared to features given by $\chi$~\cite{wang2025symmetry}.

\newcommand{\imageheightaj}{0.06\textheight}
\newcommand{\imageheightak}{0.1\textheight}

\begin{figure*}
    \centering
    \begin{tabular}{cccccccccc}
          & Source & Result & Error & Lines &  & Source & Result & Error & Lines \\
         \rotatebox{90}{$\chi$ \cite{wang2025symmetry}} & \adjustbox{valign=m}{\includegraphics[height=\imageheightaj]{./figures/images/self_matching_pred/tiger/image_iccv_source.png}} & \adjustbox{valign=m}{\includegraphics[height=\imageheightaj]{./figures/images/self_matching_pred/tiger/image_iccv_target.png}} & \adjustbox{valign=m}{\includegraphics[height=\imageheightaj]{./figures/images/self_matching_pred/tiger/image_iccv_error.png}} & 
         \adjustbox{valign=m}{\includegraphics[height=\imageheightaj]{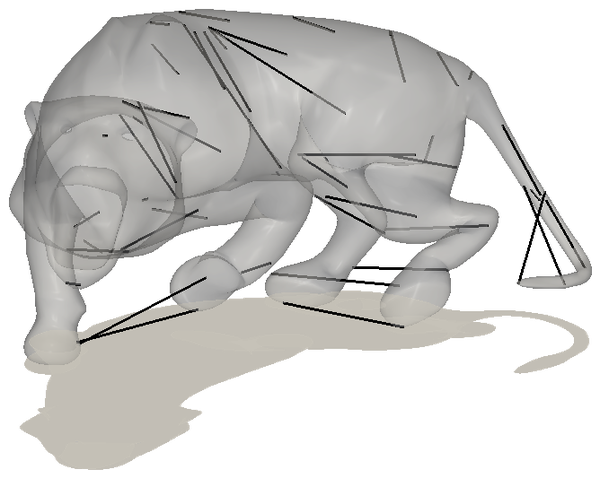}} &
         \imagespacing &
         \adjustbox{valign=m}{\includegraphics[height=\imageheightab]{./figures/images/self_matching_pred/human2/image_iccv_source.png}} & \adjustbox{valign=m}{\includegraphics[height=\imageheightab]{./figures/images/self_matching_pred/human2/image_iccv_target.png}} & \adjustbox{valign=m}{\includegraphics[height=\imageheightab]{./figures/images/self_matching_pred/human2/image_iccv_error.png}} & \adjustbox{valign=m}{\includegraphics[height=\imageheightab]{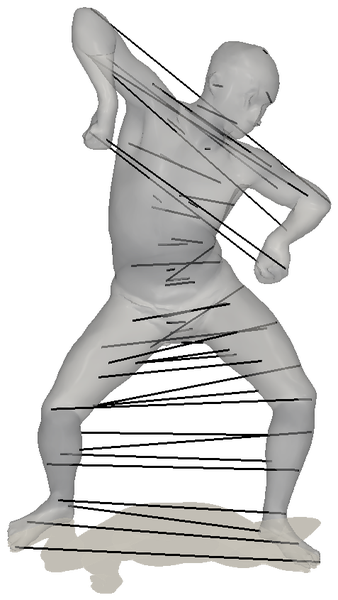}} \\

         \rotatebox{90}{\shortstack{$\chi$  \cite{wang2025symmetry} \\ (refine)}} & \adjustbox{valign=m}{\includegraphics[height=\imageheightaj]{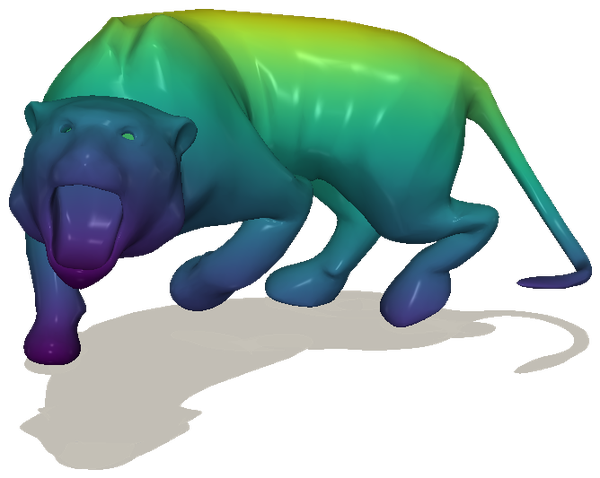}} & \adjustbox{valign=m}{\includegraphics[height=\imageheightaj]{./figures/images/self_matching_post/tiger/image_iccv_target.png}} & \adjustbox{valign=m}{\includegraphics[height=\imageheightaj]{./figures/images/self_matching_post/tiger/image_iccv_error.png}} & 
         \adjustbox{valign=m}{\includegraphics[height=\imageheightaj]{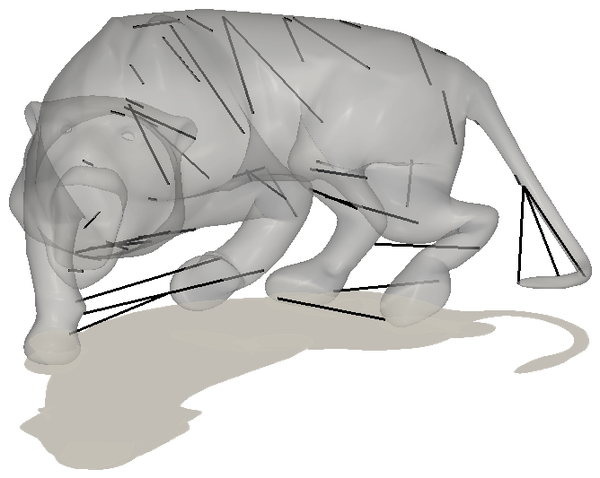}} &
         \imagespacing &
         \adjustbox{valign=m}{\includegraphics[height=\imageheightab]{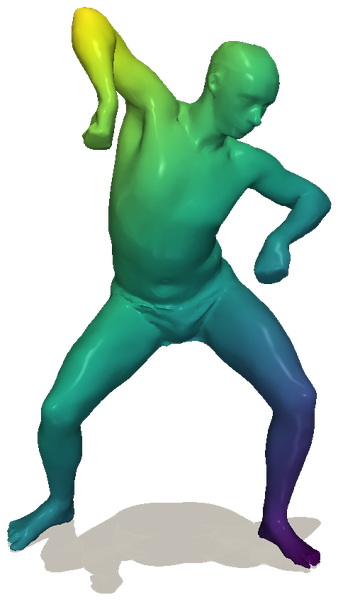}} & \adjustbox{valign=m}{\includegraphics[height=\imageheightab]{./figures/images/self_matching_post/human2/image_iccv_target.png}} & \adjustbox{valign=m}{\includegraphics[height=\imageheightab]{./figures/images/self_matching_post/human2/image_iccv_error.png}} & \adjustbox{valign=m}{\includegraphics[height=\imageheightab]{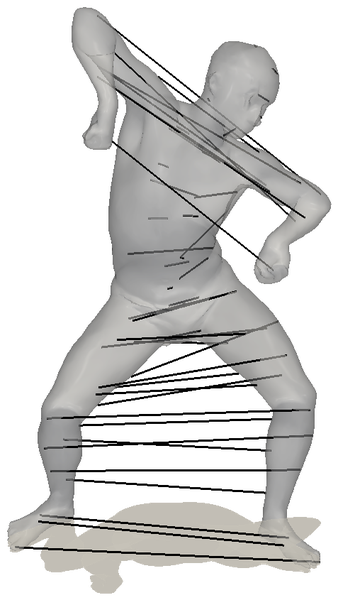}} \\

         \rotatebox{90}{\shortstack{$\chi$  \cite{wang2025symmetry} \\ (gt)}} & \adjustbox{valign=m}{\includegraphics[height=\imageheightaj]{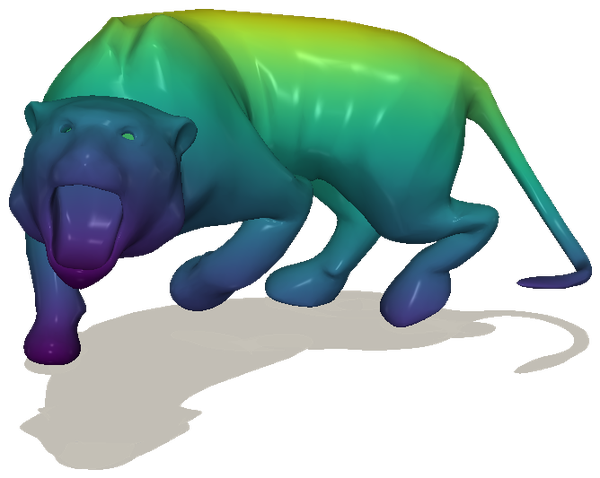}} & \adjustbox{valign=m}{\includegraphics[height=\imageheightaj]{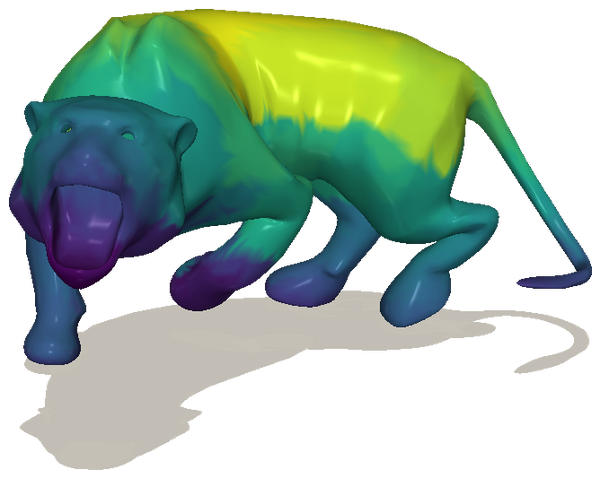}} & \adjustbox{valign=m}{\includegraphics[height=\imageheightaj]{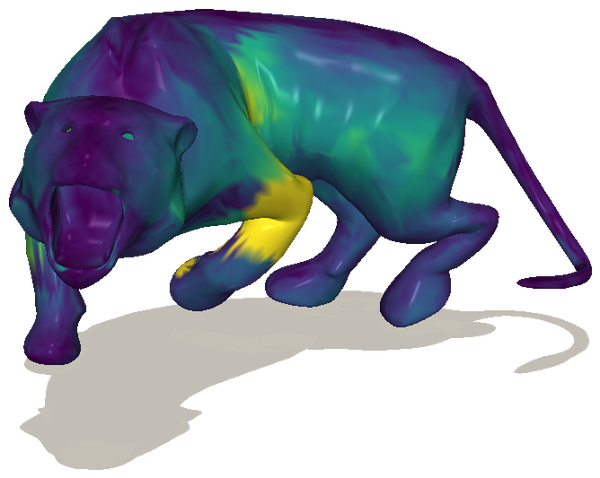}} & 
         \adjustbox{valign=m}{\includegraphics[height=\imageheightaj]{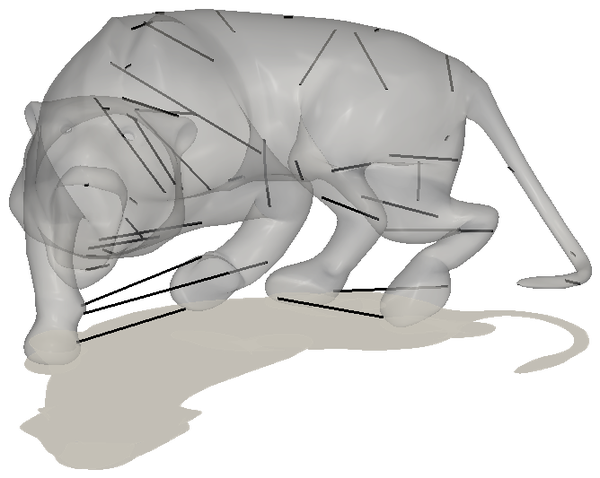}} &
         \imagespacing &
         \adjustbox{valign=m}{\includegraphics[height=\imageheightab]{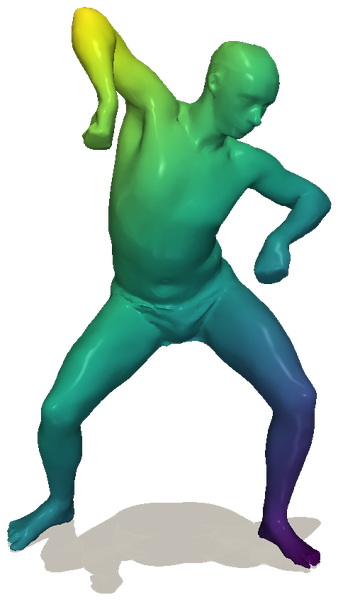}} & \adjustbox{valign=m}{\includegraphics[height=\imageheightab]{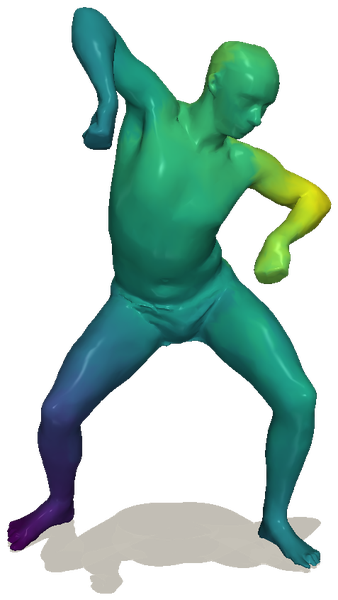}} & \adjustbox{valign=m}{\includegraphics[height=\imageheightab]{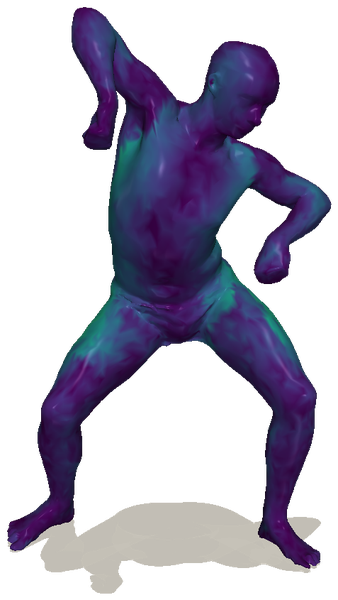}} & \adjustbox{valign=m}{\includegraphics[height=\imageheightab]{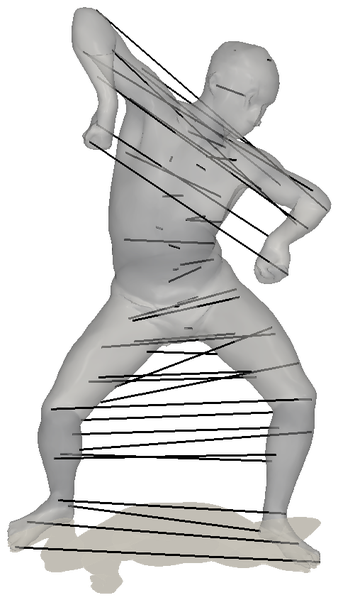}} \\

         \rotatebox{90}{Ours} & \adjustbox{valign=m}{\includegraphics[height=\imageheightaj]{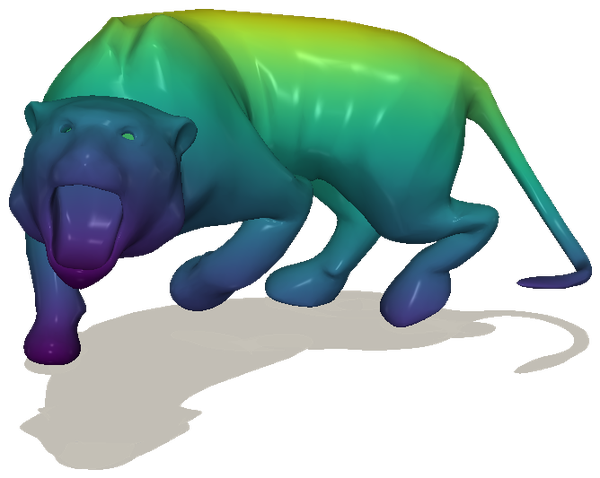}} & \adjustbox{valign=m}{\includegraphics[height=\imageheightaj]{./figures/images/self_matching_pred/tiger/image_3dv_target.png}} & \adjustbox{valign=m}{\includegraphics[height=\imageheightaj]{./figures/images/self_matching_pred/tiger/image_3dv_error.png}} &
         \adjustbox{valign=m}{\includegraphics[height=\imageheightaj]{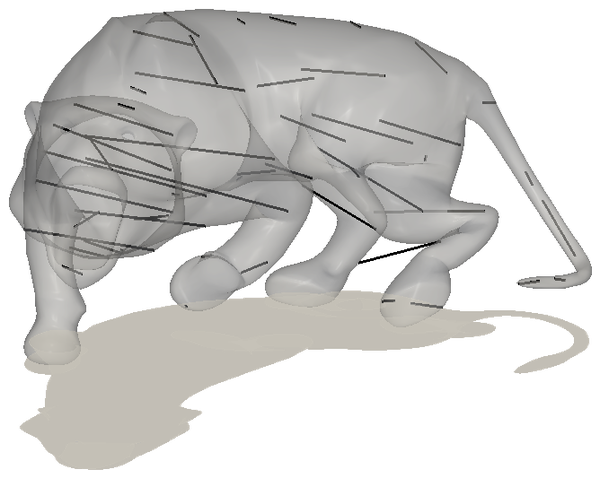}} &
         \imagespacing &
         \adjustbox{valign=m}{\includegraphics[height=\imageheightab]{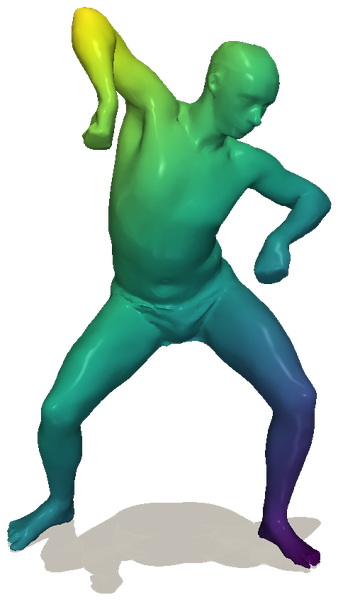}} & \adjustbox{valign=m}{\includegraphics[height=\imageheightab]{./figures/images/self_matching_pred/human2/image_3dv_target.png}} & \adjustbox{valign=m}{\includegraphics[height=\imageheightab]{./figures/images/self_matching_pred/human2/image_3dv_error.png}} &
         \adjustbox{valign=m}{\includegraphics[height=\imageheightab]{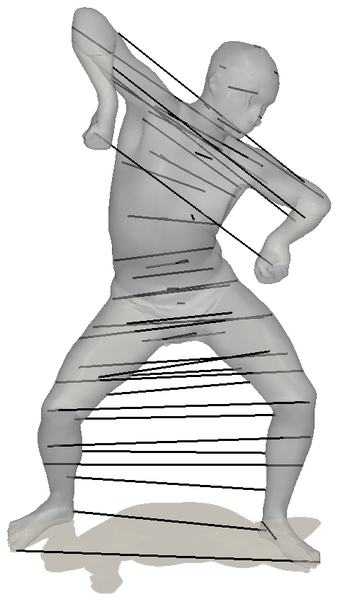}}\\

         \rotatebox{90}{\shortstack{Ours \\( refine)}} & \adjustbox{valign=m}{\includegraphics[height=\imageheightaj]{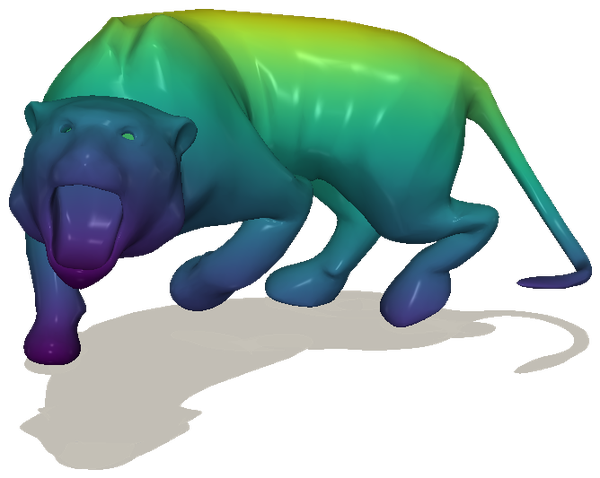}} & \adjustbox{valign=m}{\includegraphics[height=\imageheightaj]{./figures/images/self_matching_post/tiger/image_3dv_target.png}} & \adjustbox{valign=m}{\includegraphics[height=\imageheightaj]{./figures/images/self_matching_post/tiger/image_3dv_error.png}} &
         \adjustbox{valign=m}{\includegraphics[height=\imageheightaj]{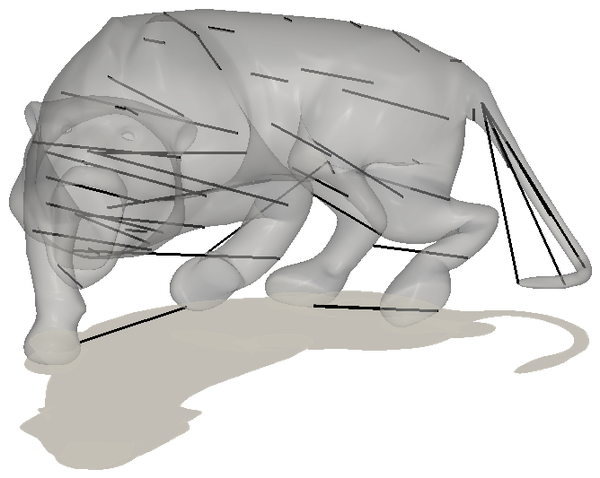}} &
         \imagespacing &
         \adjustbox{valign=m}{\includegraphics[height=\imageheightab]{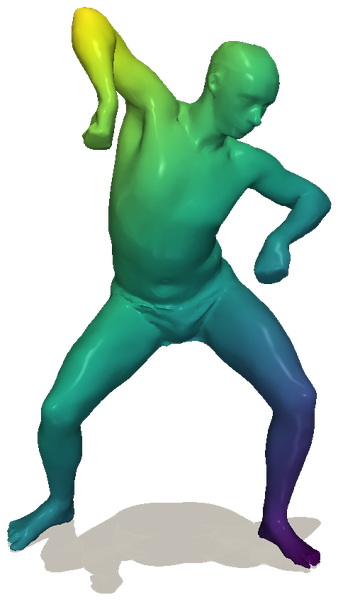}} & \adjustbox{valign=m}{\includegraphics[height=\imageheightab]{./figures/images/self_matching_post/human2/image_3dv_target.png}} & \adjustbox{valign=m}{\includegraphics[height=\imageheightab]{./figures/images/self_matching_post/human2/image_3dv_error.png}} &
         \adjustbox{valign=m}{\includegraphics[height=\imageheightab]{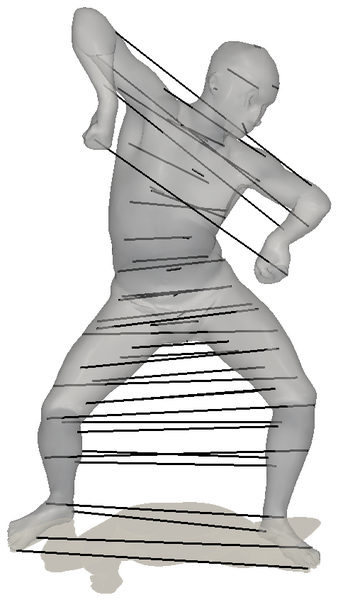}}\\

         \rotatebox{90}{\shortstack{Ours \\ (gt)}} & \adjustbox{valign=m}{\includegraphics[height=\imageheightaj]{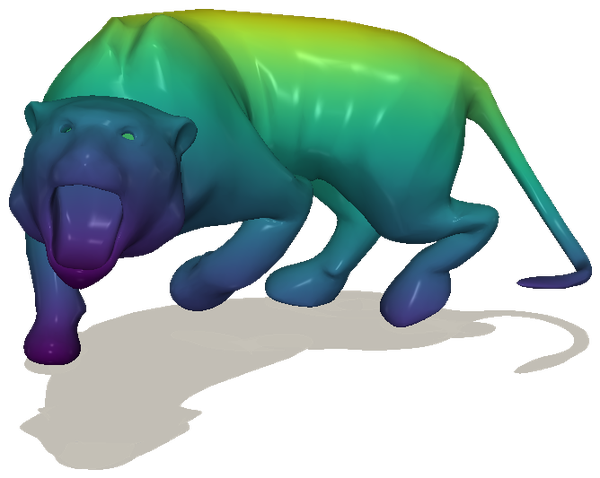}} & \adjustbox{valign=m}{\includegraphics[height=\imageheightaj]{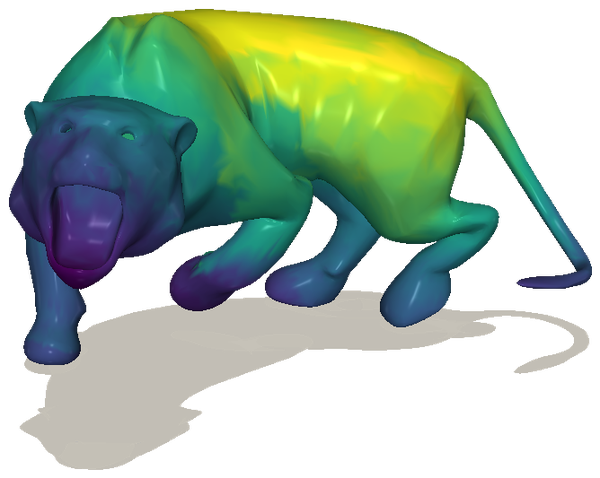}} & \adjustbox{valign=m}{\includegraphics[height=\imageheightaj]{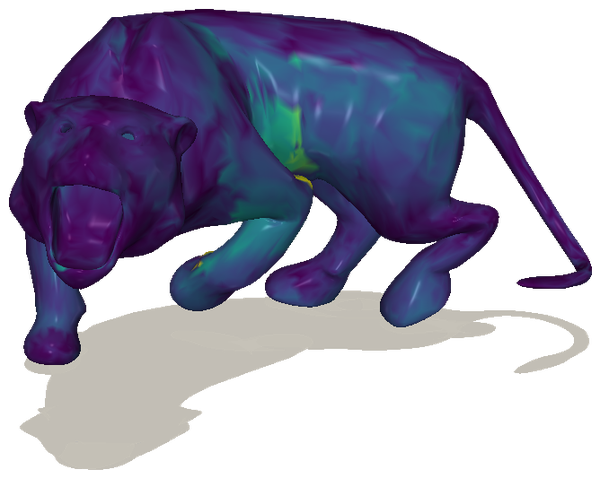}} &
         \adjustbox{valign=m}{\includegraphics[height=\imageheightaj]{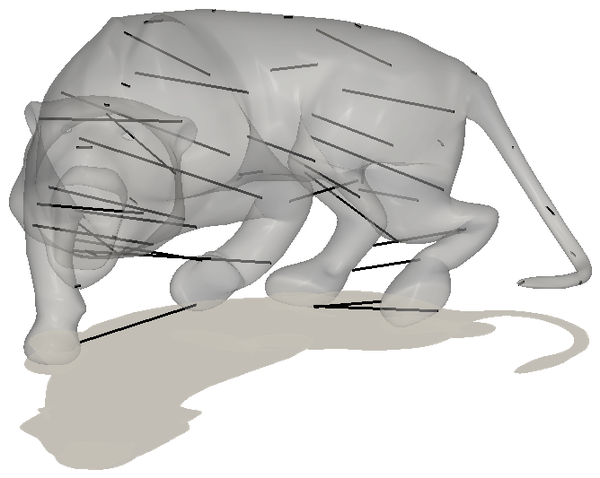}} &
         \imagespacing &
         \adjustbox{valign=m}{\includegraphics[height=\imageheightab]{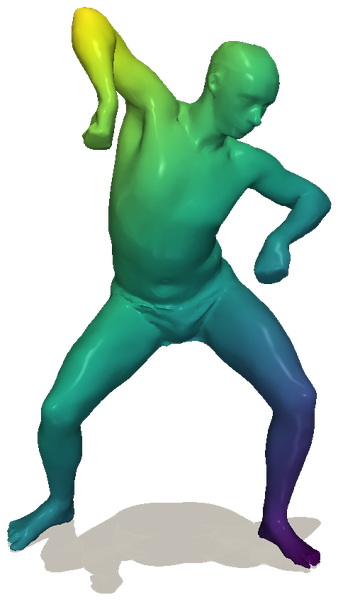}} & \adjustbox{valign=m}{\includegraphics[height=\imageheightab]{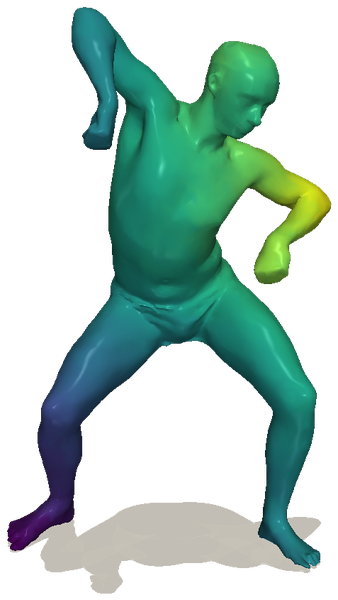}} & \adjustbox{valign=m}{\includegraphics[height=\imageheightab]{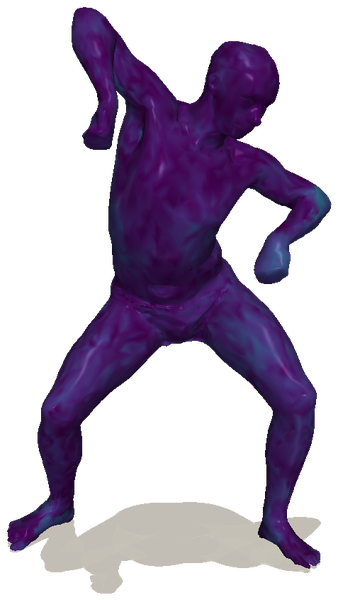}} &
         \adjustbox{valign=m}{\includegraphics[height=\imageheightab]{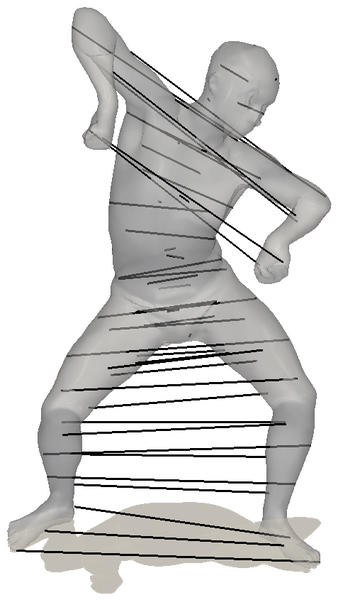}}\\

    \end{tabular}
    \caption{Qualitative results of intrinsic symmetry detection. Examples are chosen to reflect $\chi$'s \cite{wang2025symmetry} failure modes.}
    \label{fig: supple_intrinsic_symmetry}
\end{figure*}

In Fig.~\ref{fig: supple_left/right_classification}, we include more examples to qualitatively show the left/right classification results. %
Results on additional shapes in various poses confirm the effectiveness of our symmetry-informative feature.

\newcommand{\imageheightai}{0.06\textheight}

\begin{figure*}
    \centering
    \begin{tabular}{cccccccccc}
          Input & Ground truth & \citet{liu2012finding} & $\chi$ \cite{wang2025symmetry} & $\chi$ \cite{wang2025symmetry} + Refine & Ours & Ours + Refine \\
          
          \adjustbox{valign=m}{\includegraphics[height=\imageheightai]{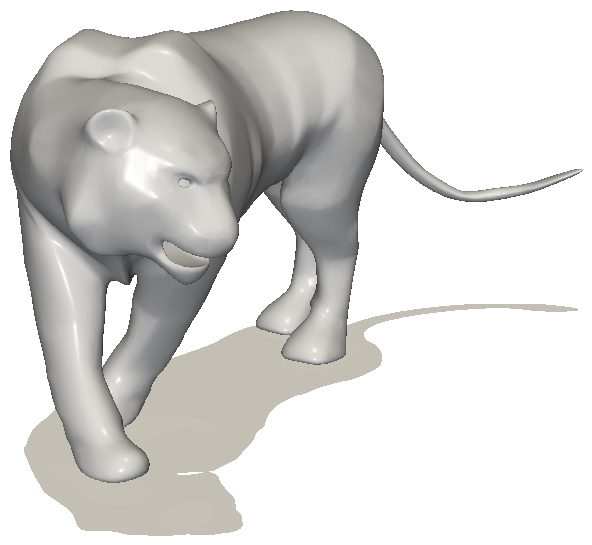}} &
          \adjustbox{valign=m}{\includegraphics[height=\imageheightai]{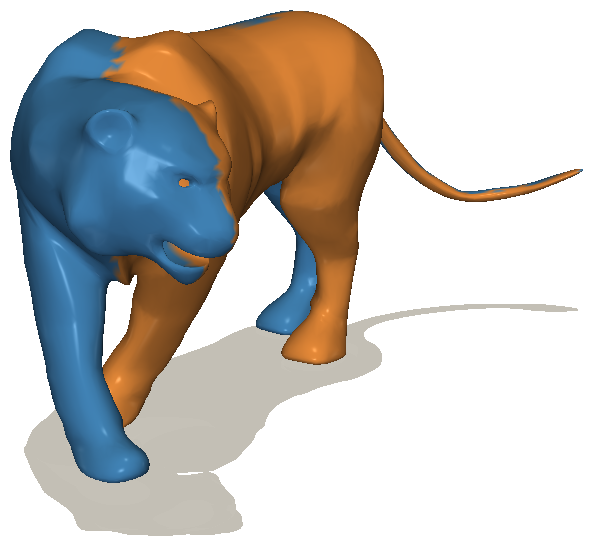}} &
          \adjustbox{valign=m}{\includegraphics[height=\imageheightai]{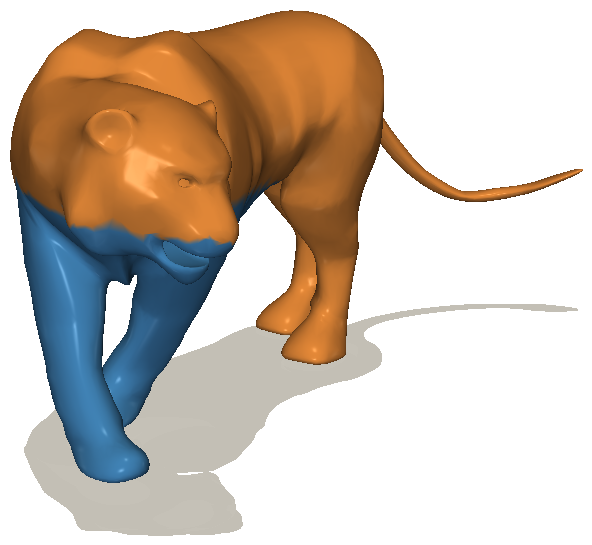}} &
          \adjustbox{valign=m}{\includegraphics[height=\imageheightai]{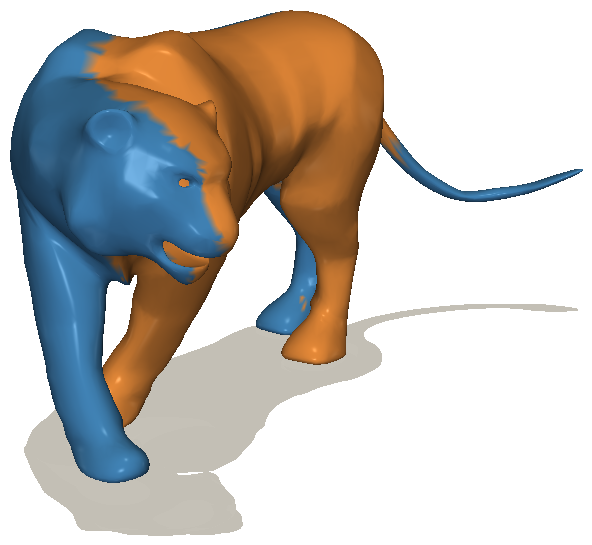}} &
          \adjustbox{valign=m}{\includegraphics[height=\imageheightai]{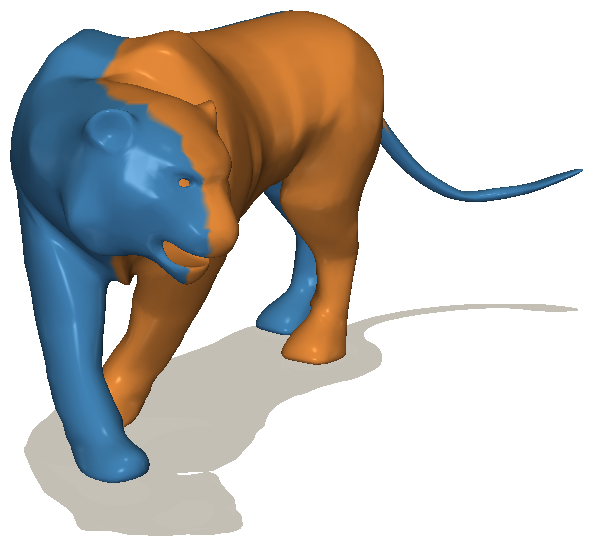}} &
          \adjustbox{valign=m}{\includegraphics[height=\imageheightai]{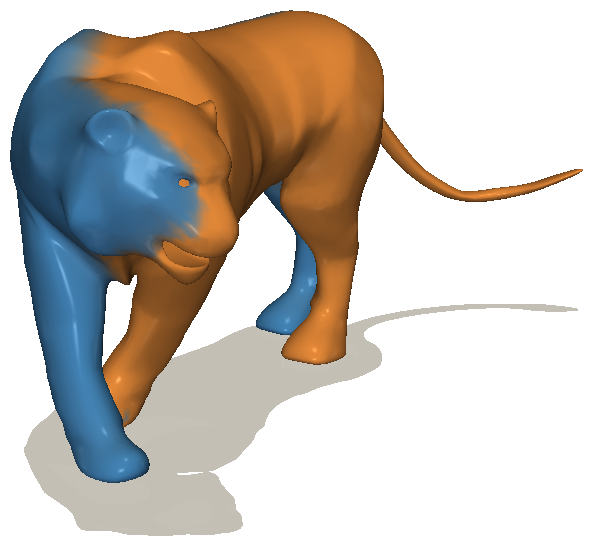}} &
          \adjustbox{valign=m}{\includegraphics[height=\imageheightai]{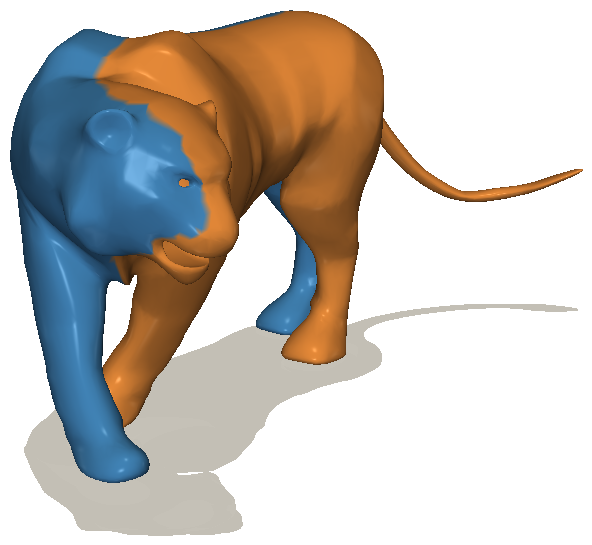}} \\

          \adjustbox{valign=m}{\includegraphics[height=\imageheightae]{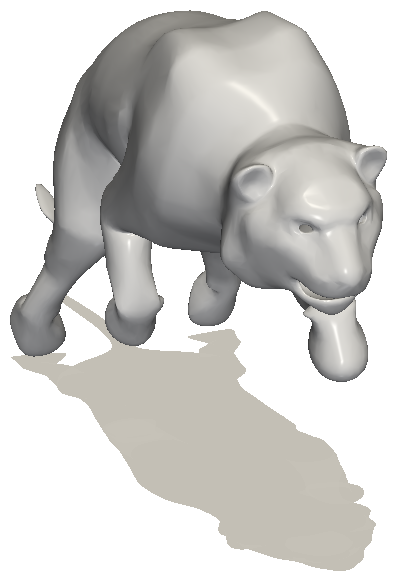}} &
          \adjustbox{valign=m}{\includegraphics[height=\imageheightae]{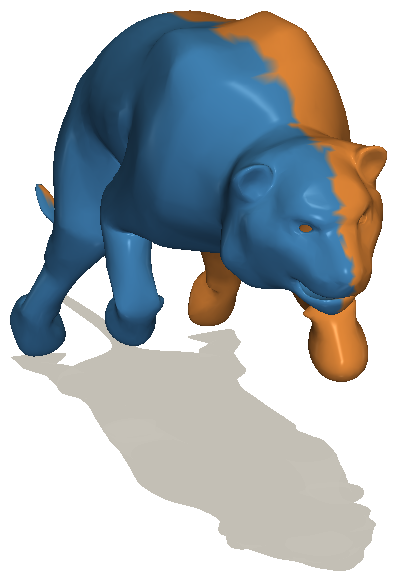}} &
          \adjustbox{valign=m}{\includegraphics[height=\imageheightae]{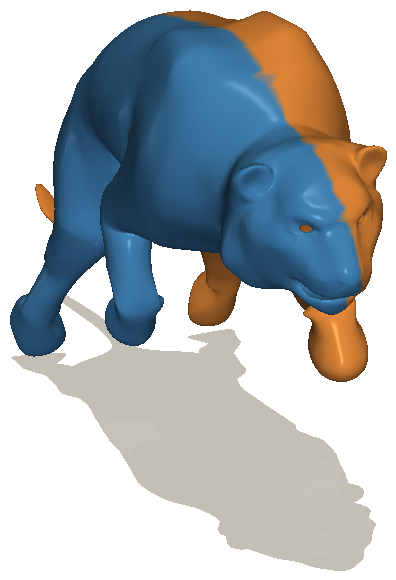}} &
          \adjustbox{valign=m}{\includegraphics[height=\imageheightae]{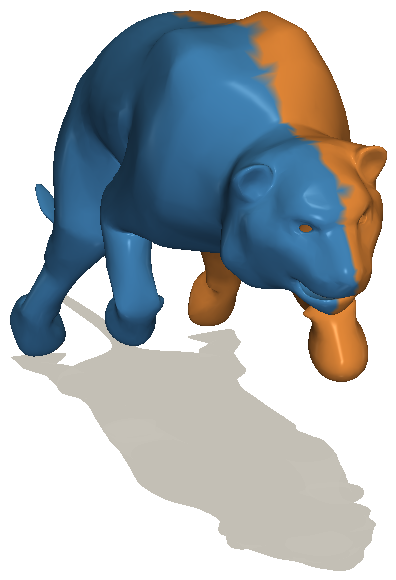}} &
          \adjustbox{valign=m}{\includegraphics[height=\imageheightae]{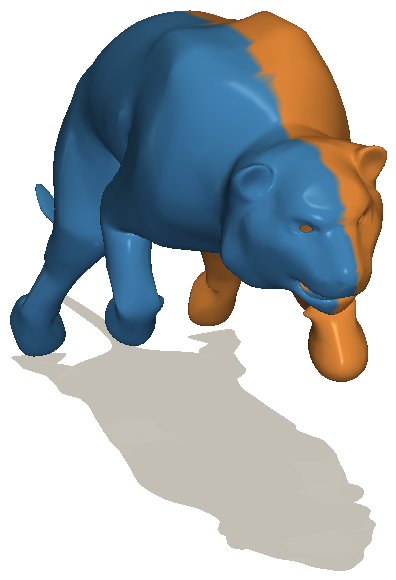}} &
          \adjustbox{valign=m}{\includegraphics[height=\imageheightae]{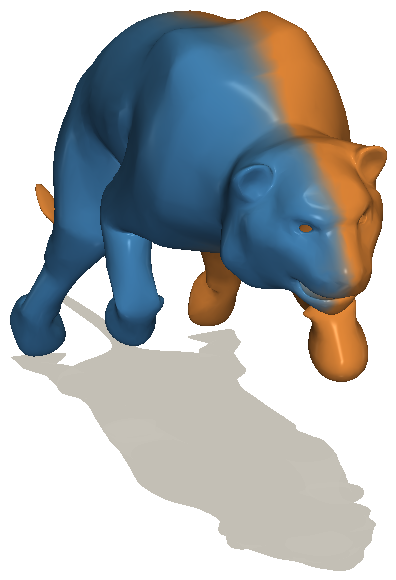}} &
          \adjustbox{valign=m}{\includegraphics[height=\imageheightae]{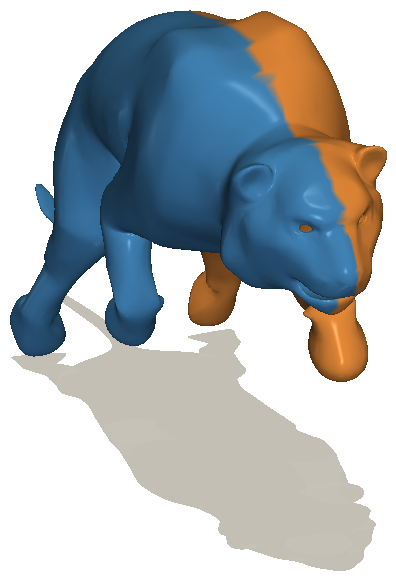}} \\

        \adjustbox{valign=m}{\includegraphics[height=\imageheightae]{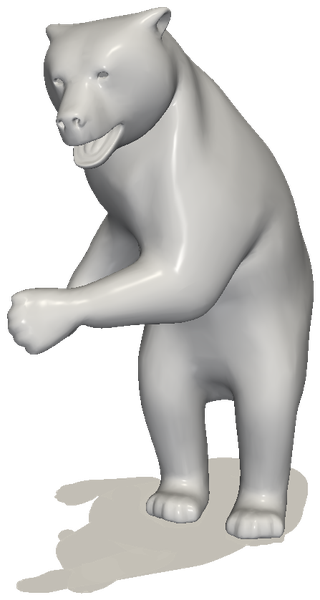}} &
        \adjustbox{valign=m}{\includegraphics[height=\imageheightae]{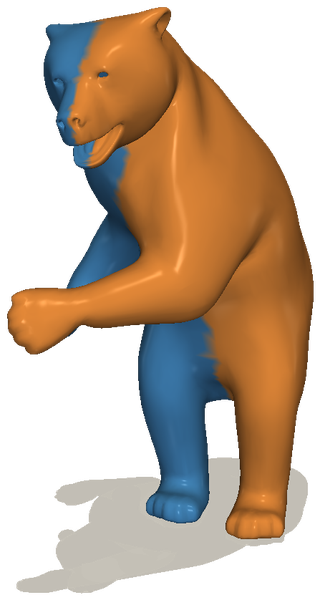}} &
        \adjustbox{valign=m}{\includegraphics[height=\imageheightae]{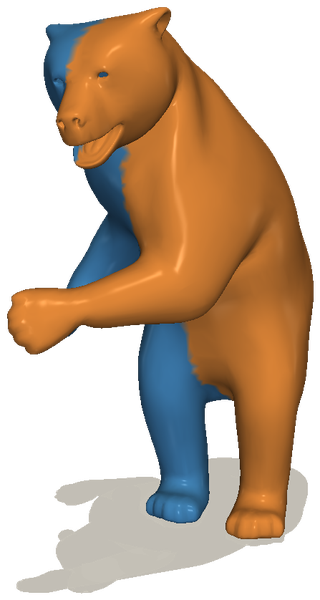}} &
          \adjustbox{valign=m}{\includegraphics[height=\imageheightae]{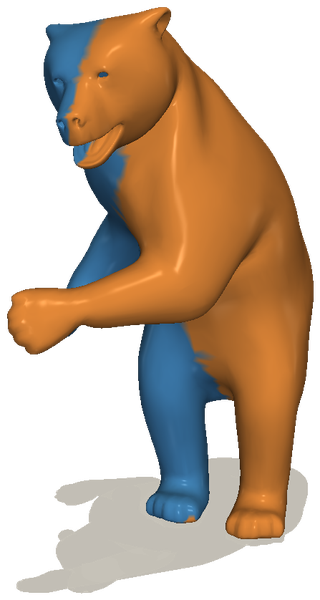}} &
          \adjustbox{valign=m}{\includegraphics[height=\imageheightae]{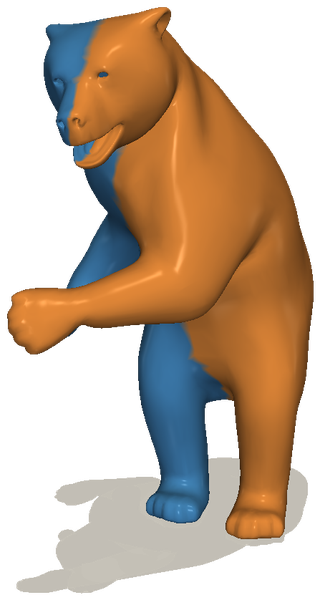}} &
          \adjustbox{valign=m}{\includegraphics[height=\imageheightae]{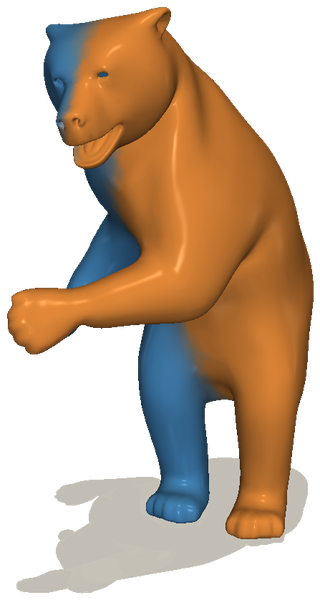}} &
          \adjustbox{valign=m}{\includegraphics[height=\imageheightae]{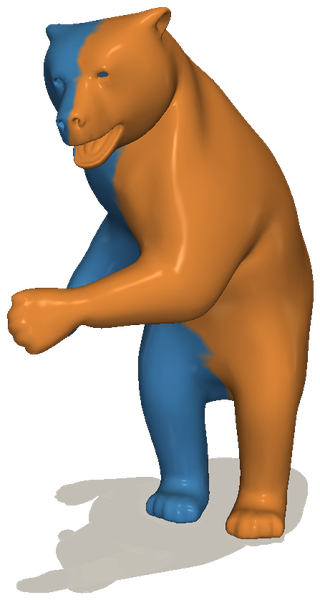}} \\

          \adjustbox{valign=m}{\includegraphics[height=\imageheightai]{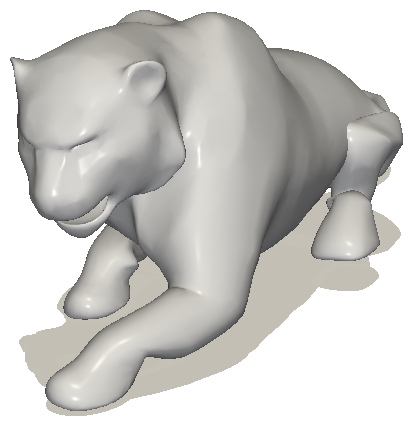}} &
          \adjustbox{valign=m}{\includegraphics[height=\imageheightai]{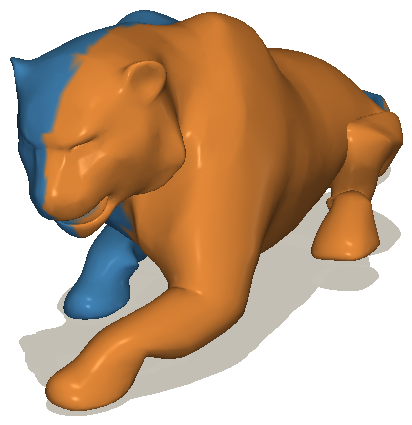}} &
          \adjustbox{valign=m}{\includegraphics[height=\imageheightai]{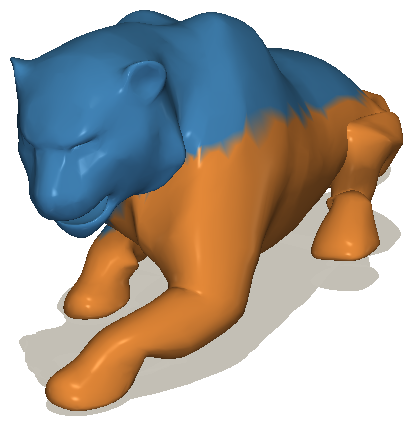}} &
          \adjustbox{valign=m}{\includegraphics[height=\imageheightai]{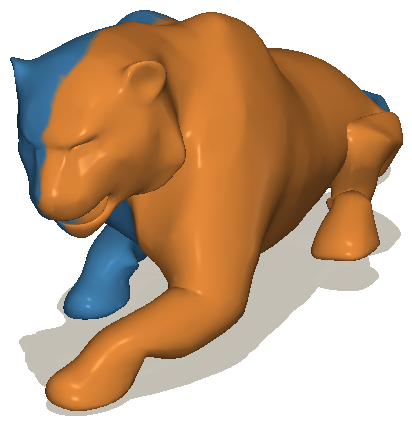}} &
          \adjustbox{valign=m}{\includegraphics[height=\imageheightai]{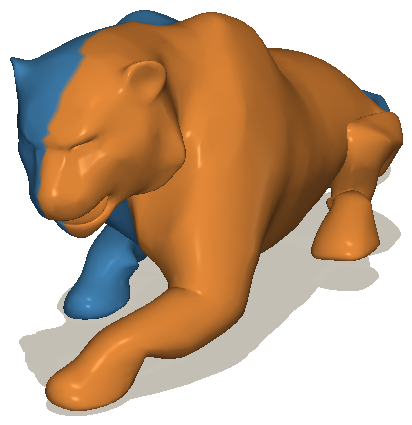}} &
          \adjustbox{valign=m}{\includegraphics[height=\imageheightai]{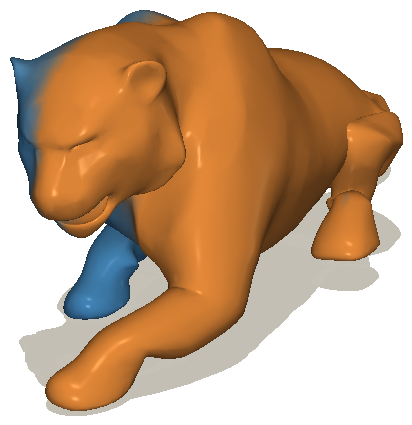}} &
          \adjustbox{valign=m}{\includegraphics[height=\imageheightai]{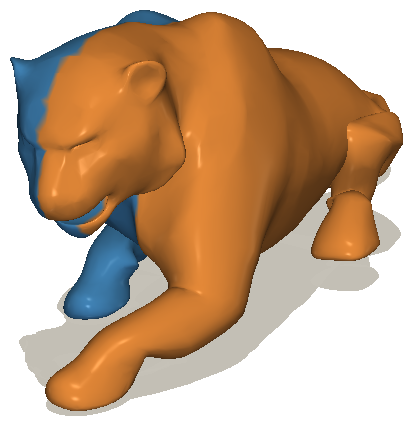}} \\

          \adjustbox{valign=m}{\includegraphics[height=\imageheightae]{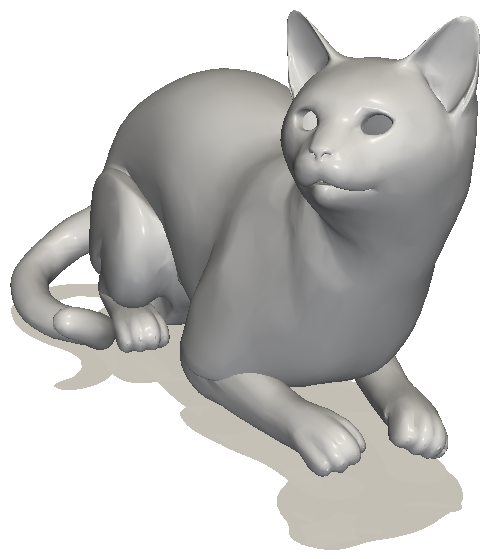}} &
          \adjustbox{valign=m}{\includegraphics[height=\imageheightae]{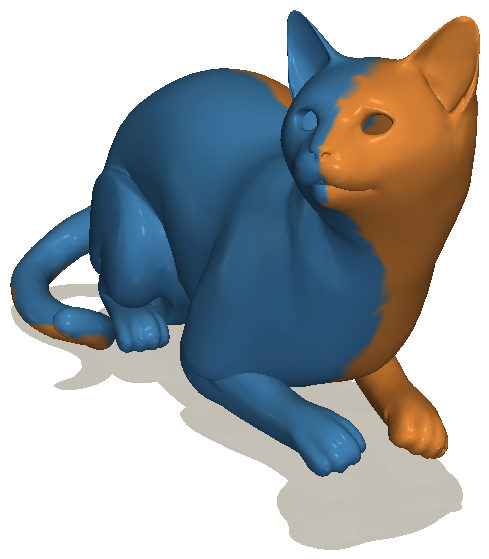}} &
          \adjustbox{valign=m}{\includegraphics[height=\imageheightae]{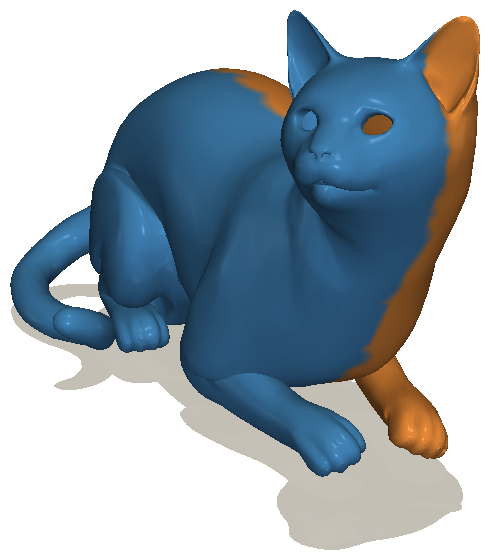}} &
          \adjustbox{valign=m}{\includegraphics[height=\imageheightae]{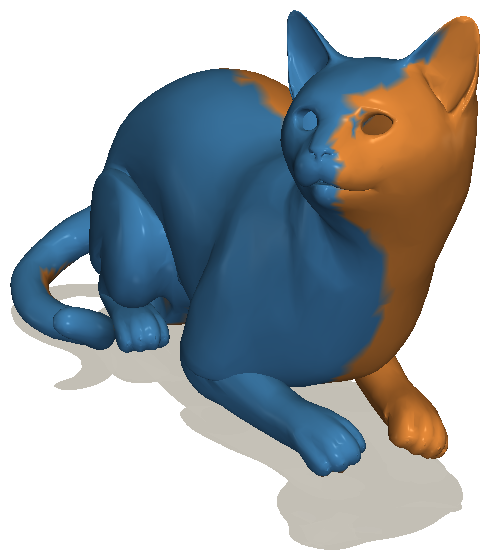}} &
          \adjustbox{valign=m}{\includegraphics[height=\imageheightae]{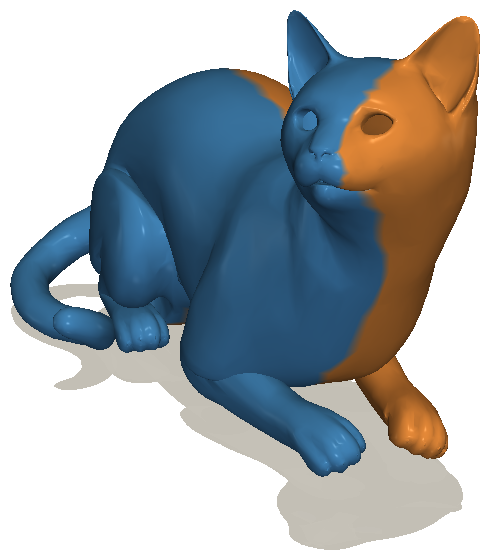}} &
          \adjustbox{valign=m}{\includegraphics[height=\imageheightae]{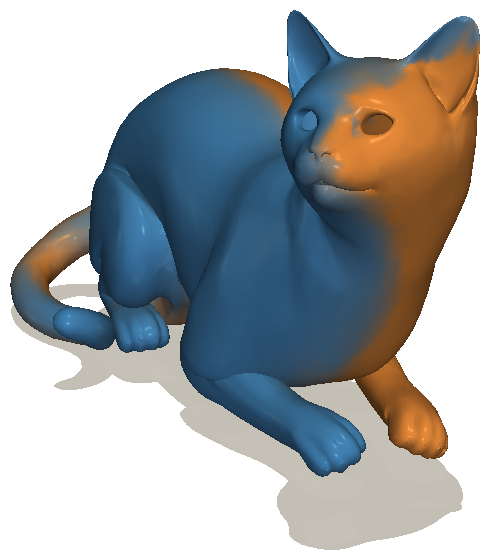}} &
          \adjustbox{valign=m}{\includegraphics[height=\imageheightae]{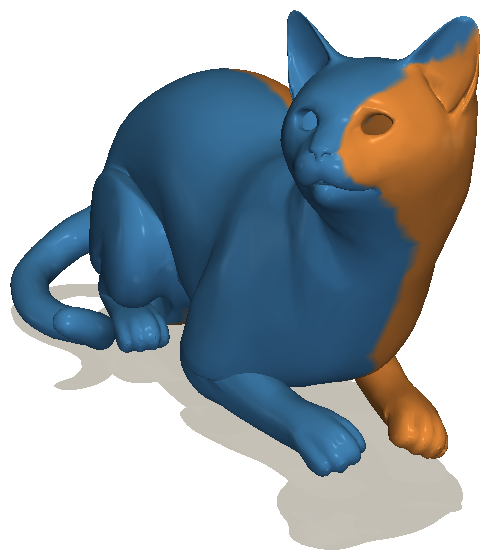}} \\

          \adjustbox{valign=m}{\includegraphics[height=\imageheightae]{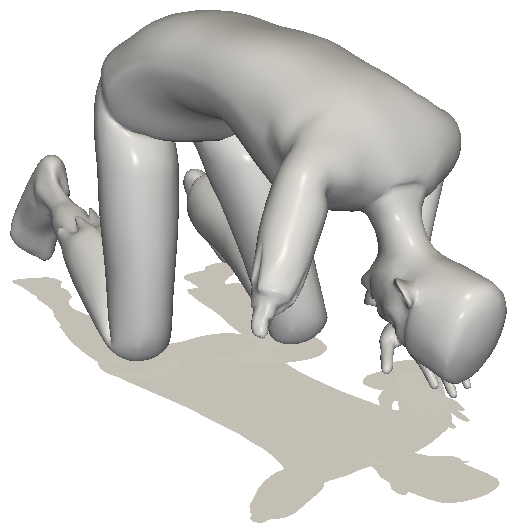}} &
          \adjustbox{valign=m}{\includegraphics[height=\imageheightae]{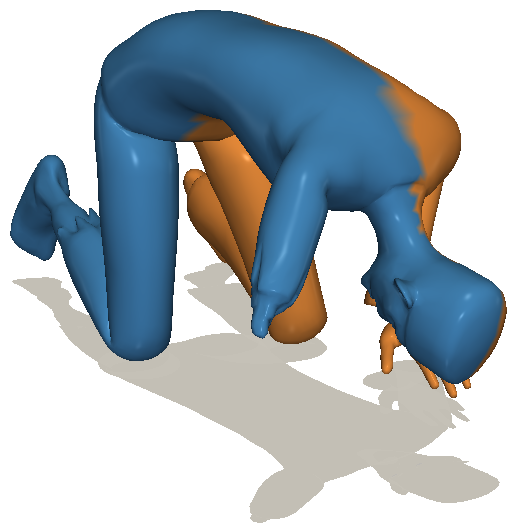}} &
          \adjustbox{valign=m}{\includegraphics[height=\imageheightae]{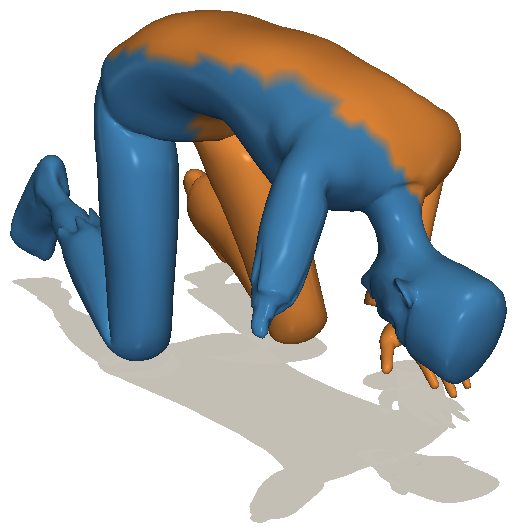}} &
          \adjustbox{valign=m}{\includegraphics[height=\imageheightae]{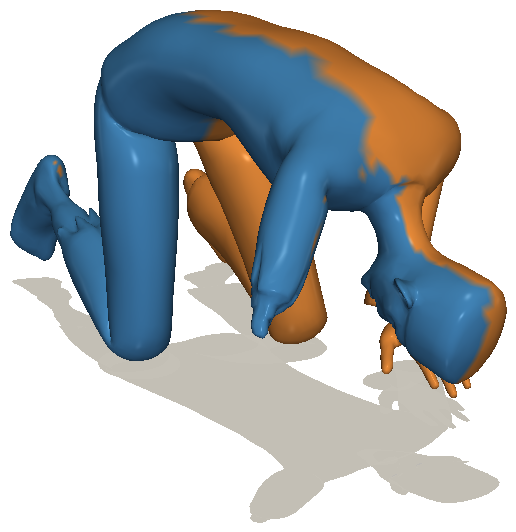}} &
          \adjustbox{valign=m}{\includegraphics[height=\imageheightae]{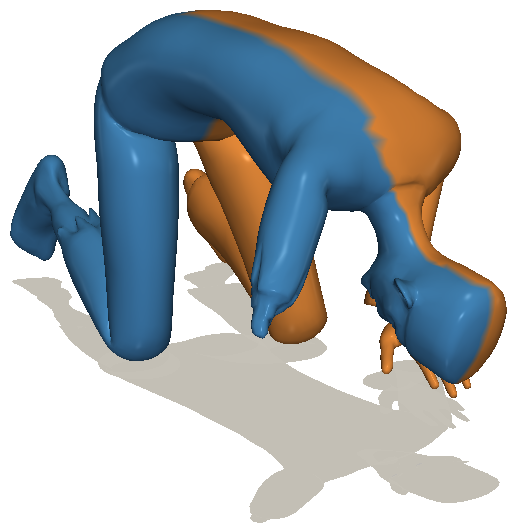}} &
          \adjustbox{valign=m}{\includegraphics[height=\imageheightae]{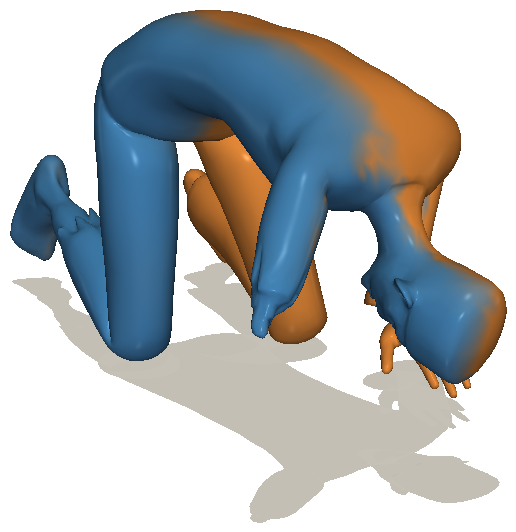}} &
          \adjustbox{valign=m}{\includegraphics[height=\imageheightae]{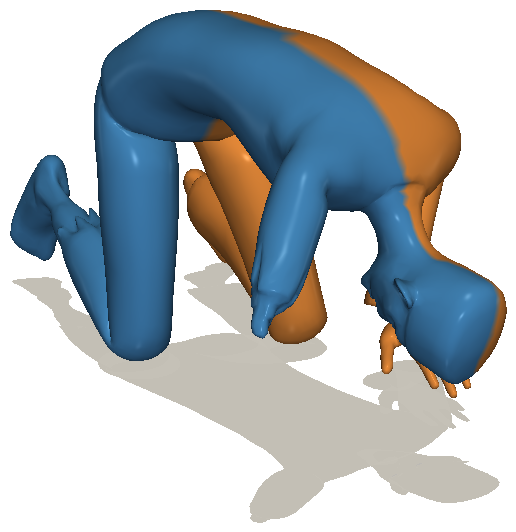}} \\

          \adjustbox{valign=m}{\includegraphics[height=\imageheightai]{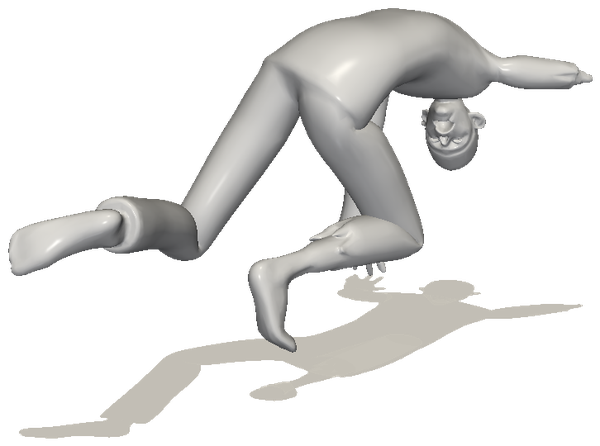}} &
          \adjustbox{valign=m}{\includegraphics[height=\imageheightai]{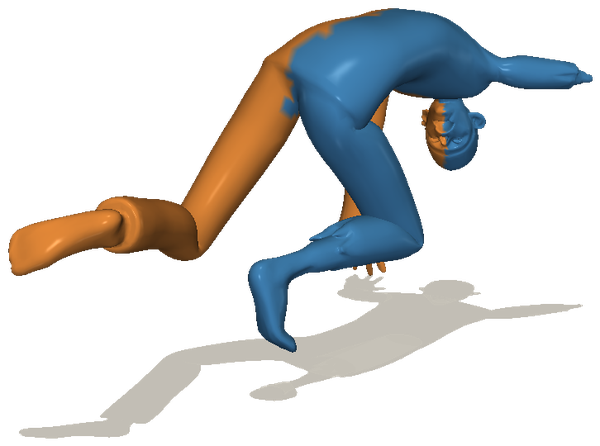}} &
          \adjustbox{valign=m}{\includegraphics[height=\imageheightai]{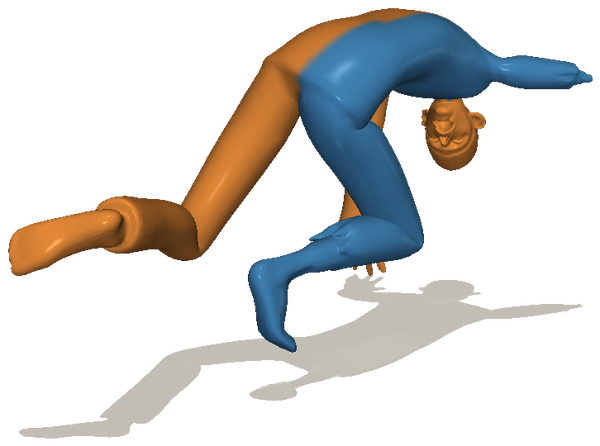}} &
          \adjustbox{valign=m}{\includegraphics[height=\imageheightai]{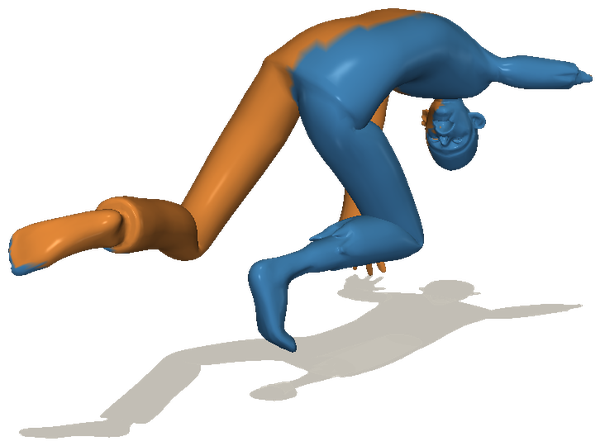}} &
          \adjustbox{valign=m}{\includegraphics[height=\imageheightai]{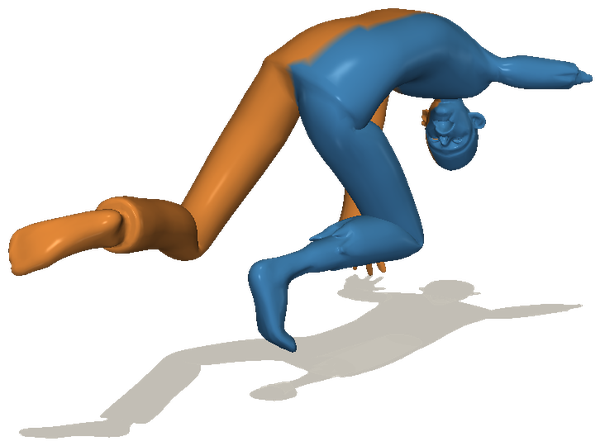}} &
          \adjustbox{valign=m}{\includegraphics[height=\imageheightai]{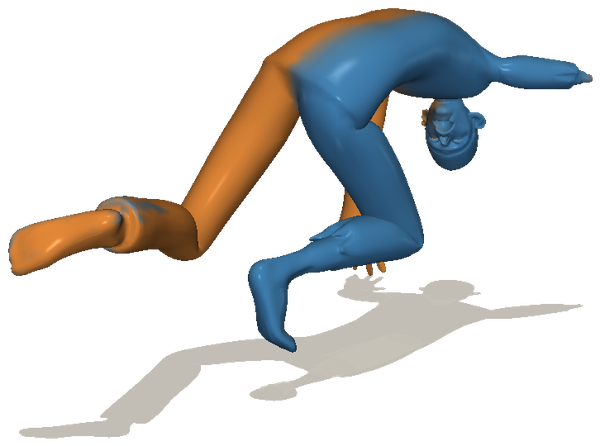}} &
          \adjustbox{valign=m}{\includegraphics[height=\imageheightai]{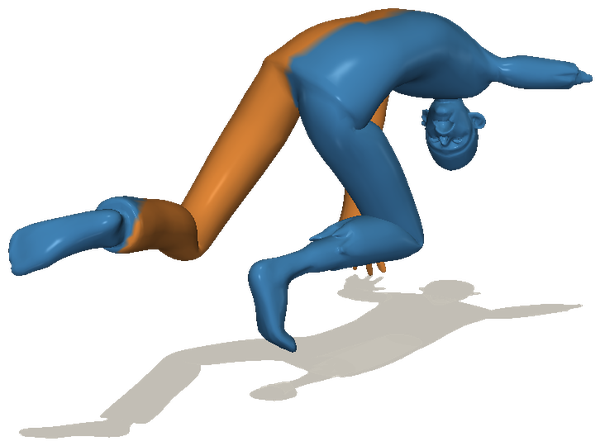}} \\

          \adjustbox{valign=m}{\includegraphics[height=\imageheightae]{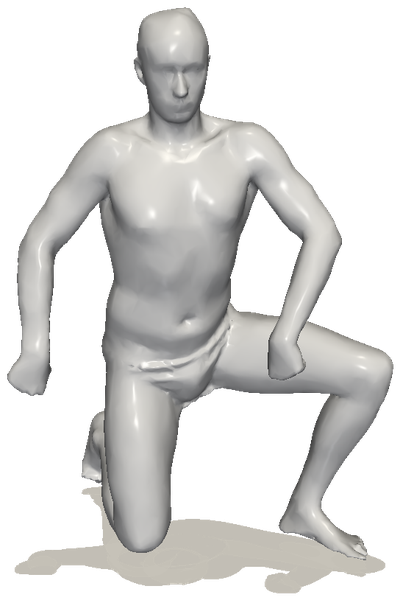}} &
          \adjustbox{valign=m}{\includegraphics[height=\imageheightae]{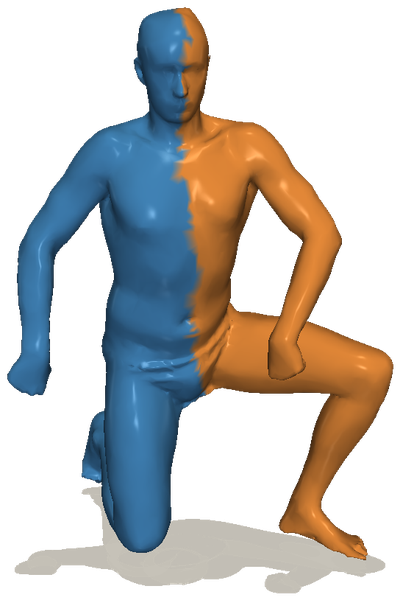}} &
          \adjustbox{valign=m}{\includegraphics[height=\imageheightae]{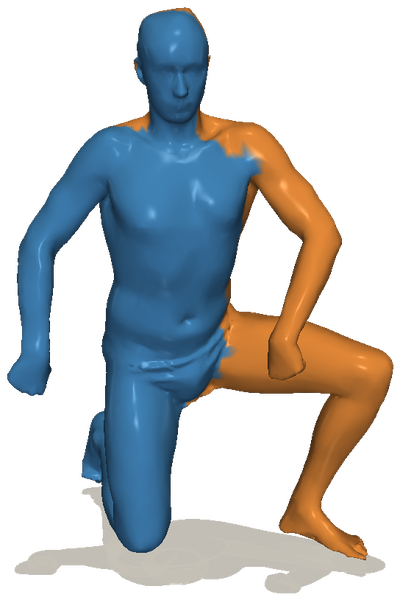}} &
          \adjustbox{valign=m}{\includegraphics[height=\imageheightae]{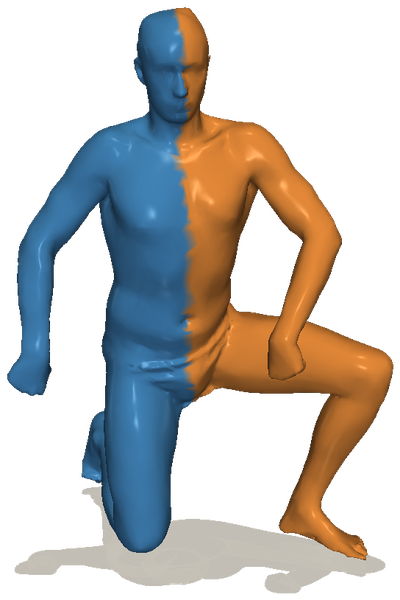}} &
          \adjustbox{valign=m}{\includegraphics[height=\imageheightae]{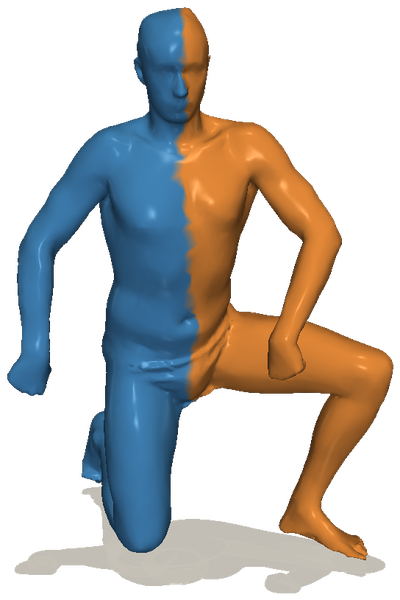}} &
          \adjustbox{valign=m}{\includegraphics[height=\imageheightae]{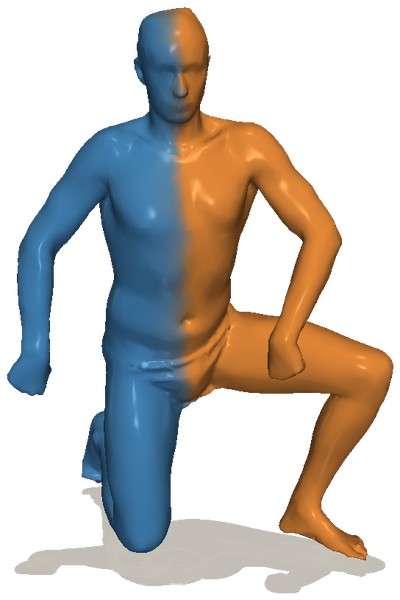}} &
          \adjustbox{valign=m}{\includegraphics[height=\imageheightae]{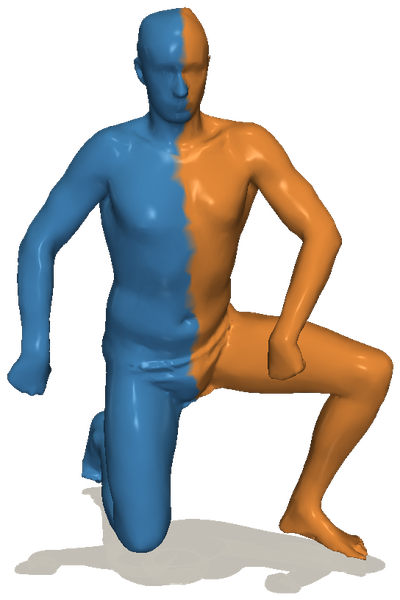}} \\

          \adjustbox{valign=m}{\includegraphics[height=\imageheightae]{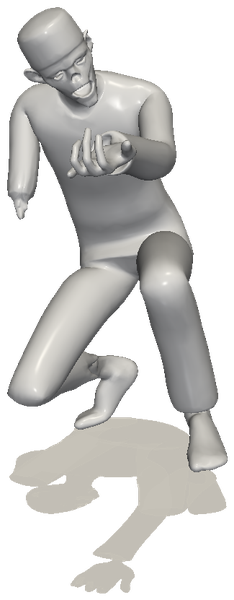}} &
          \adjustbox{valign=m}{\includegraphics[height=\imageheightae]{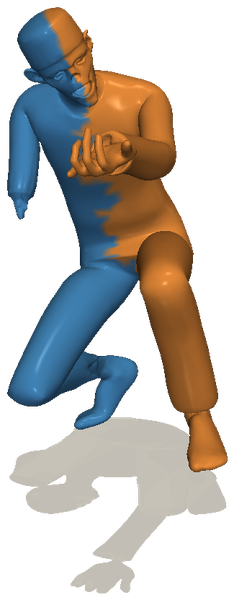}} &
          \adjustbox{valign=m}{\includegraphics[height=\imageheightae]{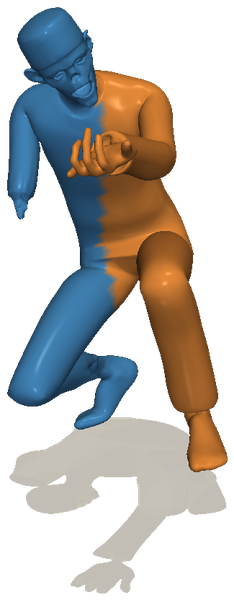}} &
          \adjustbox{valign=m}{\includegraphics[height=\imageheightae]{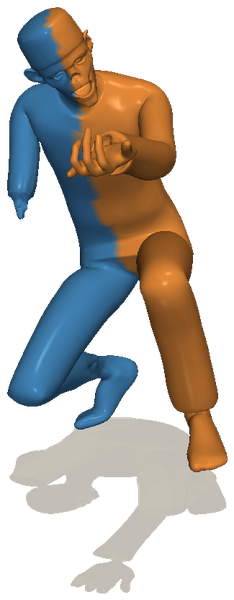}} &
          \adjustbox{valign=m}{\includegraphics[height=\imageheightae]{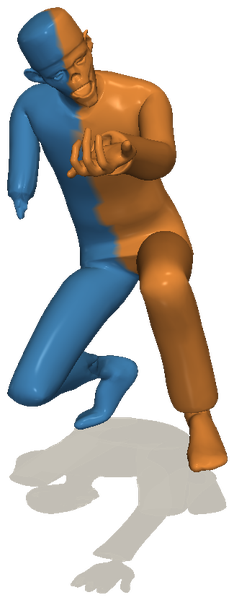}} &
          \adjustbox{valign=m}{\includegraphics[height=\imageheightae]{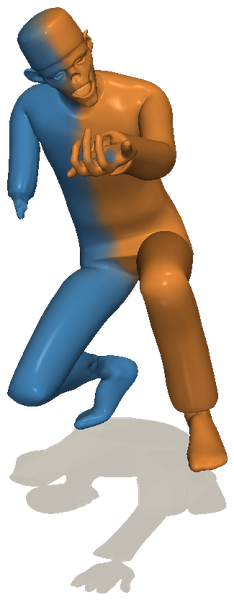}} &
          \adjustbox{valign=m}{\includegraphics[height=\imageheightae]{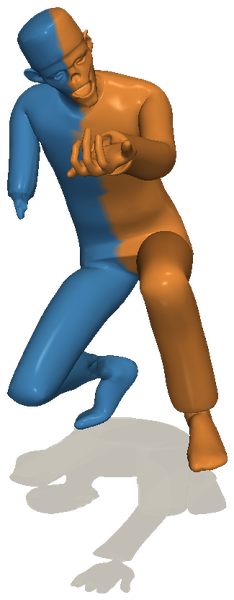}} \\

          \adjustbox{valign=m}{\includegraphics[height=\imageheightae]{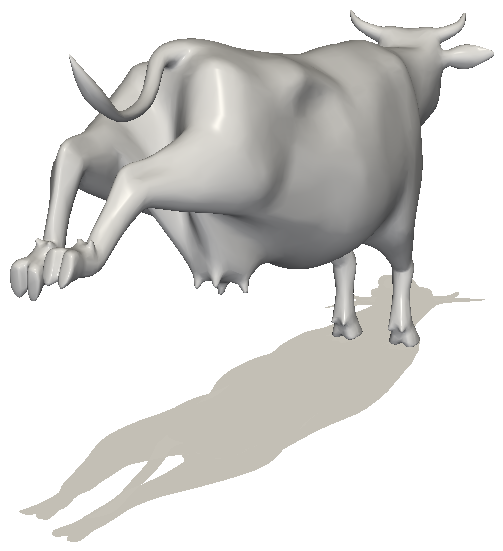}} &
          \adjustbox{valign=m}{\includegraphics[height=\imageheightae]{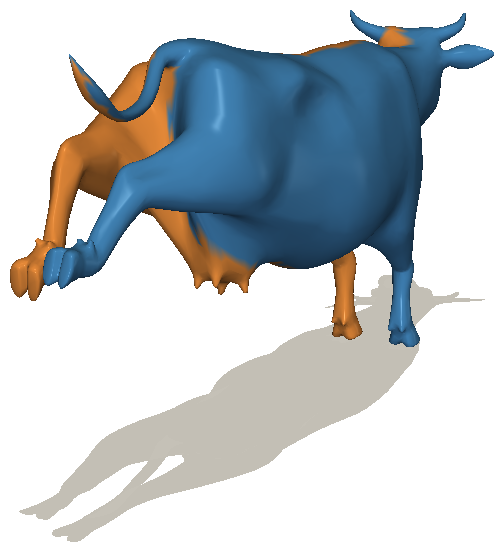}} &
          \adjustbox{valign=m}{\includegraphics[height=\imageheightae]{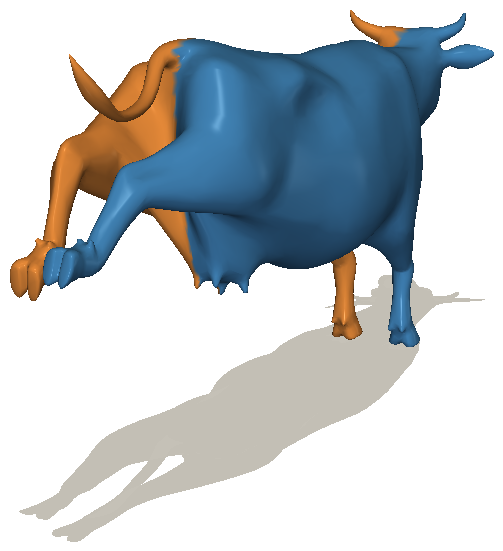}} &
          \adjustbox{valign=m}{\includegraphics[height=\imageheightae]{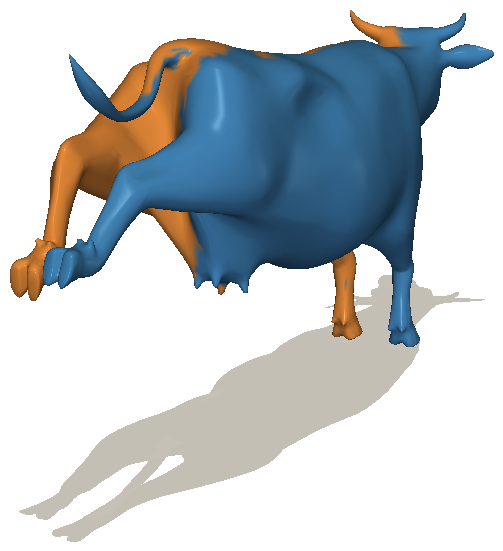}} &
          \adjustbox{valign=m}{\includegraphics[height=\imageheightae]{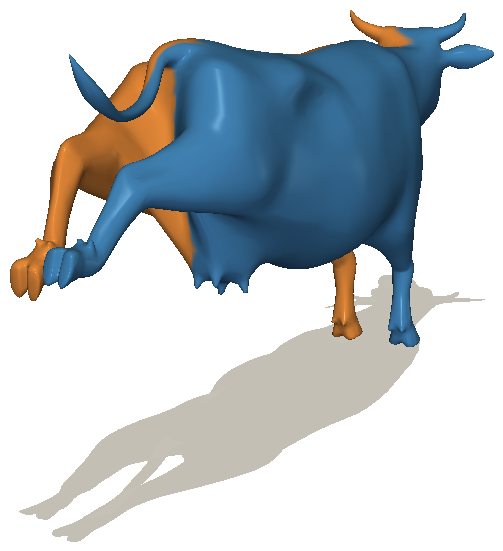}} &
          \adjustbox{valign=m}{\includegraphics[height=\imageheightae]{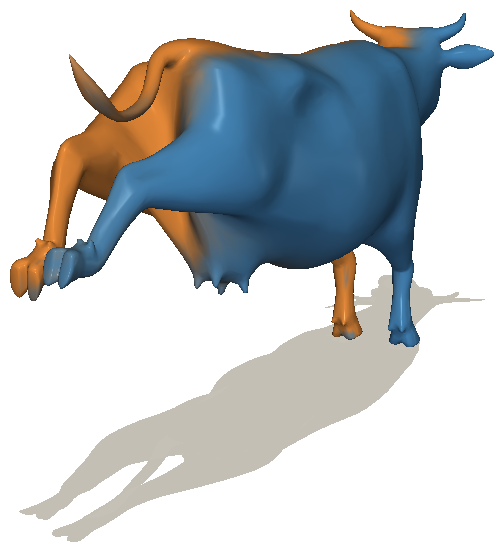}} &
          \adjustbox{valign=m}{\includegraphics[height=\imageheightae]{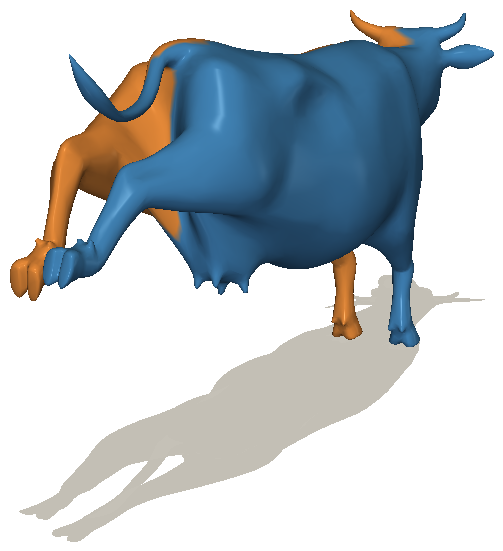}} \\

          \adjustbox{valign=m}{\includegraphics[height=\imageheightae]{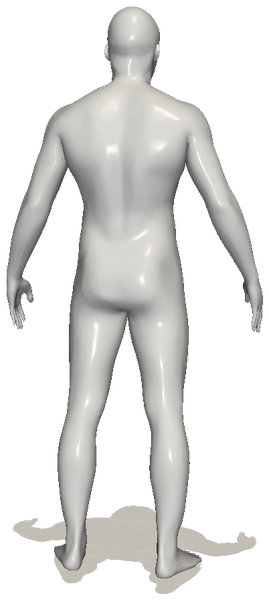}} &
          \adjustbox{valign=m}{\includegraphics[height=\imageheightae]{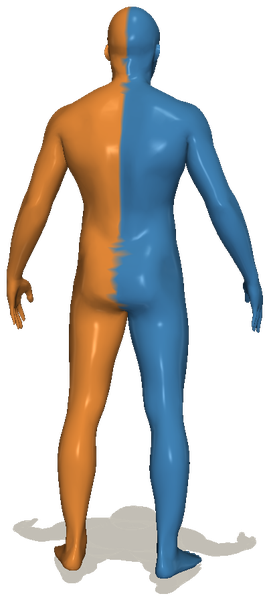}} &
          \adjustbox{valign=m}{\includegraphics[height=\imageheightae]{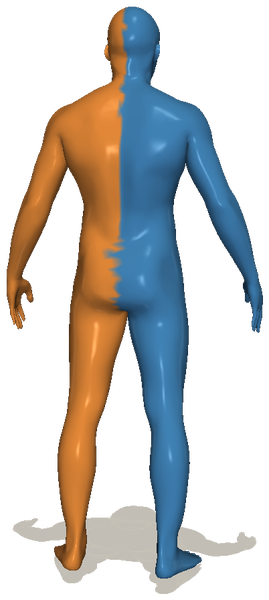}} &
          \adjustbox{valign=m}{\includegraphics[height=\imageheightae]{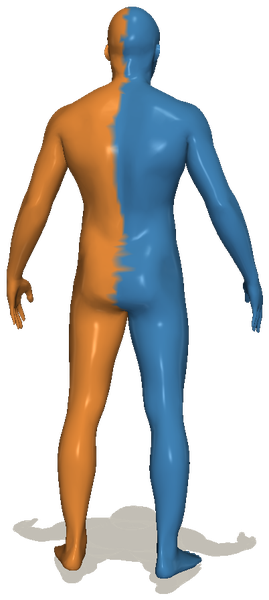}} &
          \adjustbox{valign=m}{\includegraphics[height=\imageheightae]{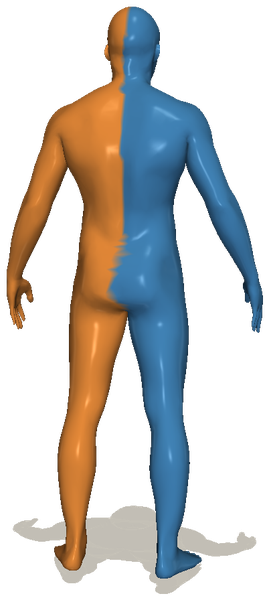}} &
          \adjustbox{valign=m}{\includegraphics[height=\imageheightae]{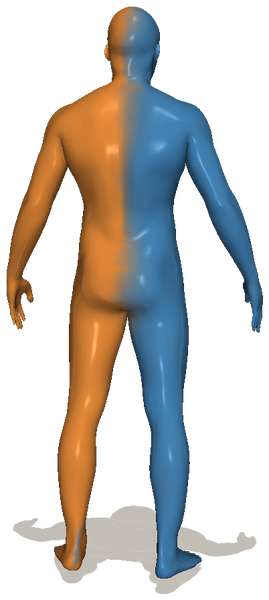}} &
          \adjustbox{valign=m}{\includegraphics[height=\imageheightae]{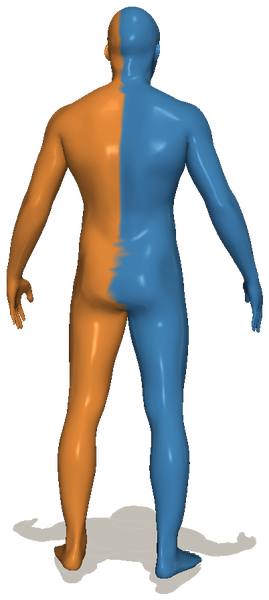}} \\

    \end{tabular}
    \caption{Qualitative results of left/right classification.}
    \label{fig: supple_left/right_classification}
\end{figure*}

In Fig.~\ref{fig: supple_matching}, we add qualitative results of more baselines (including $\chi$~\cite{wang2025symmetry} with refinement and ours without refinement) for a more comprehensive comparison. %
By comparing methods without refinement in the first four columns and by comparing methods with refinement in the last four columns, we can conclude that our method achieves better matching results compared to $\chi$~\cite{wang2025symmetry} (which is more prominent in the error visualizations).
Further, by comparing 
$\chi$~\cite{wang2025symmetry} with and without refinement as well as by comparing ours with and without refinement, we can confirm the effectiveness of our proposed feature refinement technique.

\newpage

\end{document}